\documentclass{article} 
\usepackage[preprint]{colm2026_conference}
\usepackage{microtype}
\usepackage{hyperref}
\usepackage{url}
\usepackage{booktabs}
\usepackage{multirow}
\usepackage{graphicx}
\usepackage[table,xcdraw]{xcolor}
\usepackage{wrapfig}
\usepackage{tabularx}
\usepackage[most]{tcolorbox}
\usepackage{array}
\usepackage{makecell}
\usepackage[inkscapelatex=false]{svg}
\usepackage{subcaption}
\usepackage{courier}
\usepackage{amssymb}
\usepackage[most]{tcolorbox}
\usepackage{longtable}

\usepackage{lineno}

\definecolor{darkblue}{rgb}{0, 0, 0.5}
\hypersetup{colorlinks=true, citecolor=darkblue, linkcolor=darkblue, urlcolor=darkblue}

\title{Seeing Isn’t Believing: Uncovering Blind Spots in Evaluator Vision-Language Models}

\author{
    Mohammed Safi Ur Rahman Khan$^{\spadesuit\diamondsuit}$, Sanjay Suryanarayanan$^\spadesuit$,
    Tushar Anand$^\clubsuit$,\\ ~\textbf{Mitesh M. Khapra}$^{\spadesuit\diamondsuit}$ \\ \\
    $^\spadesuit$Nilekani Centre at AI4Bharat 
    $^\diamondsuit$ Indian Institute of Technology Madras\\
    $^\clubsuit$BITS Pilani, Hyderabad\\
    \textbf{Correspondence:} \texttt{\{mohammed.safi, miteshk\}@dsai.iitm.ac.in}
    \\
    \\
 \includegraphics[height=1.1em]{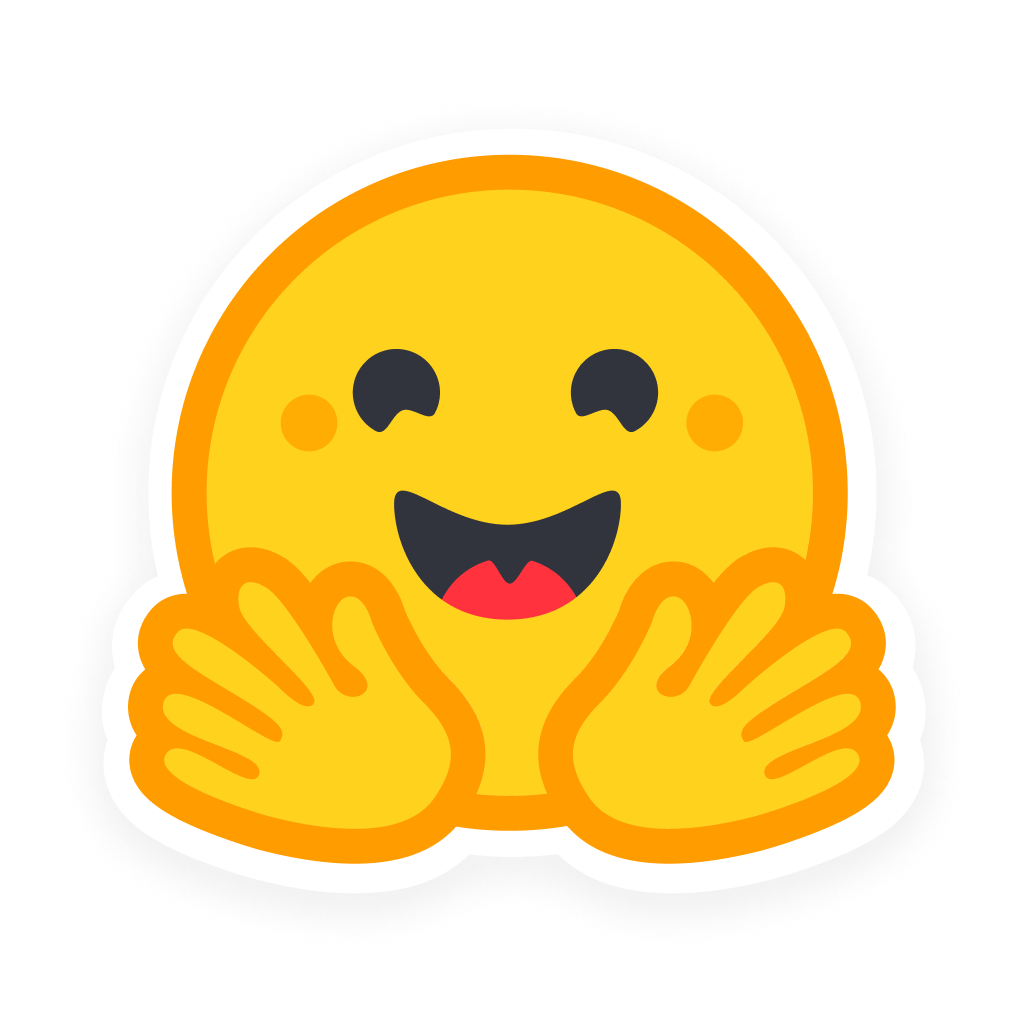}~\small{\url{https://huggingface.co/datasets/ai4bharat/Focus}}
 \\
 \includegraphics[height=1em]{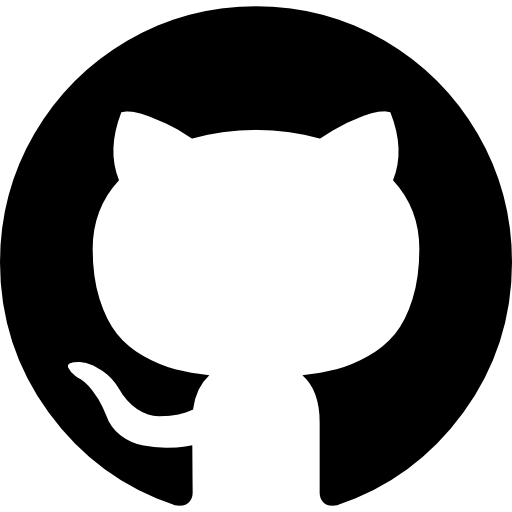}~\small{\url{https://github.com/AI4Bharat/focus}}
}

\newcommand{\bench}{\textsc{Focus}}
\newcommand{\itot}{\textsc{\textbf{i2t}}}
\newcommand{\ttoi}{\textsc{\textbf{t2i}}}

\newcommand{\gemini}{\textsc{gemini-3.1-pro}}
\newcommand{\gpt}{\textsc{gpt-5.4}}
\newcommand{\claude}{\textsc{claude-Opus-4.6}}
\newcommand{\qwen}{\textsc{Qwen3.5-397B-A17B}}
\newcommand{\vgemini}{\textsc{gemini-3-pro-image}}

\definecolor{i2t_si}{HTML}{d9ead3}
\definecolor{i2t_vr}{HTML}{fce5cd}
\definecolor{myred}{HTML}{f4cccc}
\definecolor{i2t_lg}{HTML}{ead1dc}
\definecolor{i2t_vg}{HTML}{cfe2f3}
\definecolor{mygray}{HTML}{efefef}

\definecolor{t2i_vf}{HTML}{DAF0F2}
\definecolor{t2i_sc}{HTML}{fff2cc}
\definecolor{t2i_pp}{HTML}{d9d2e9}
\definecolor{t2i_tr}{HTML}{FBDEDE}

\newif\ifcomments
\commentstrue
\ifcomments
    \newcommand{\todo}[1]{{\color{red} #1}}
    \newcommand{\safi}[1]{{\color{teal} #1}}
    
    \newcommand{\mk}[1]{{\color{green} #1}}
\else
    \newcommand{\todo}[1]{}
    \newcommand{\safi}[1]{}
    \newcommand{\sd}[1]{}
    \newcommand{\mk}[1]{}
\fi

\begin{document}

\ifcolmsubmission
\linenumbers
\fi

\maketitle

\begin{abstract}

Large Vision-Language Models (VLMs) are increasingly used to evaluate outputs of other models, for image-to-text (\itot) tasks such as visual question answering, and text-to-image (\ttoi) generation tasks. Despite this growing reliance, the reliability of these Evaluator VLMs remains underexplored. In this work, we systematically evaluate the reliability of Evaluator VLMs across both \itot~and \ttoi~tasks. We introduce targeted perturbations that degrade output quality along key error dimensions, including object hallucinations, spatial reasoning, factual grounding, and visual fidelity. These perturbations test whether Evaluator VLMs can reliably account for these quality degrading errors in their evaluations. Using a comprehensive benchmark of over 4000 perturbed instances spanning 40 perturbation dimensions, we evaluate 4 prominent VLMs using single-answer scoring, pairwise comparison, and reference-guided paradigms. Our findings reveal that current VLM evaluators exhibit substantial blind spots: they often fail to detect perturbed outputs - in some cases exceeding 50\%, struggle particularly with fine-grained compositional and spatial errors, and are often insensitive to hallucinated content that contradicts the input image. Pairwise comparison proves more reliable, though failure rates persist. These results highlight the unreliable nature of current Evaluator VLMs and urge caution in their deployment for benchmarking and development decisions. Code and data have been made publicly available.
\end{abstract}

\section{Introduction}


Large Vision-Language Models (VLMs) are increasingly used to evaluate the outputs of other VLMs~\citep{zhang2023gpt4vision, yu2023mmvet, pu2025judge} and image generation models~\citep{Wen2023ImprovingCT, zhou2025multimodalllmscustomizedreward, Yang2025SelfRewardingLV} as they are scalable and cost-effective than human evaluation. Beyond benchmarking, these models are also used as reward models during training, where their feedback directly shapes model behavior~\citep{Li2025UniworldV2RI, yasunaga2025multimodal}. As a result, unreliablity of evaluator VLMs can have broad impact: they can produce misleading rankings and also reinforce undesirable behaviors during training. It is therefore important to rigorously assess their reliability.


Recent work has studied evaluator VLMs through their correlation with human judgments~\citep{kasaei2025evaluating, hu2025multimodal}. However, establishing the dependability of evaluator VLMs requires rigorous scrutiny, to identify potential blind spots in their capabilities. For example, in image-to-text tasks such as Visual Question Answering, an evaluator must verify whether generated text is grounded in the image by detecting hallucinated objects, incorrect attributes, and fabricated facts~\citep{Jing2023FaithScoreFE, Bai2024HallucinationOM}. In text-to-image tasks, it should be able to judge whether a generated image faithfully reflects the prompt, including objects, attributes, physical plausibility, and rendered text~\citep{huang2023t2i0compbench000, meng2024phybench}. \textit{Can current VLMs reliably perform such fine-grained, multi-dimensional assessments, or do they exhibit systematic blind spots that render their judgments unreliable?} 

In this work, we introduce \bench, a comprehensive meta-evaluation benchmark designed to uncover blind spots in Evaluator VLMs across both \itot~and \ttoi~tasks. Our approach is inspired by prior meta-evaluation frameworks such as Checklist~\citep{ribeiro2020beyond} and FBI~\citep{doddapaneni2024finding} and is grounded in a simple principle: if a perturbation introduces a clear error into a model output, a reliable evaluator should detect this degradation and adjust its judgment accordingly. We design targeted perturbations spanning diverse failure modes, organized into fine-grained dimensions based on commonly reported errors in the literature (sample descriptions are shown in Table \ref{tab: pert_cat}). Starting with 600 and 750 prompts for \itot~and \ttoi~tasks respectively, sampled from real-world benchmarks, we generate gold responses using \gemini~and \vgemini. We then introduce targeted perturbations through a rigorous human-in-the-loop process, resulting in a dataset of over 4000 instances. Each instance contains a prompt, a gold response, and a perturbed response, and is validated by expert annotators.

\begin{figure}[!t]
    \centering
    \includegraphics[width=1.0\columnwidth]{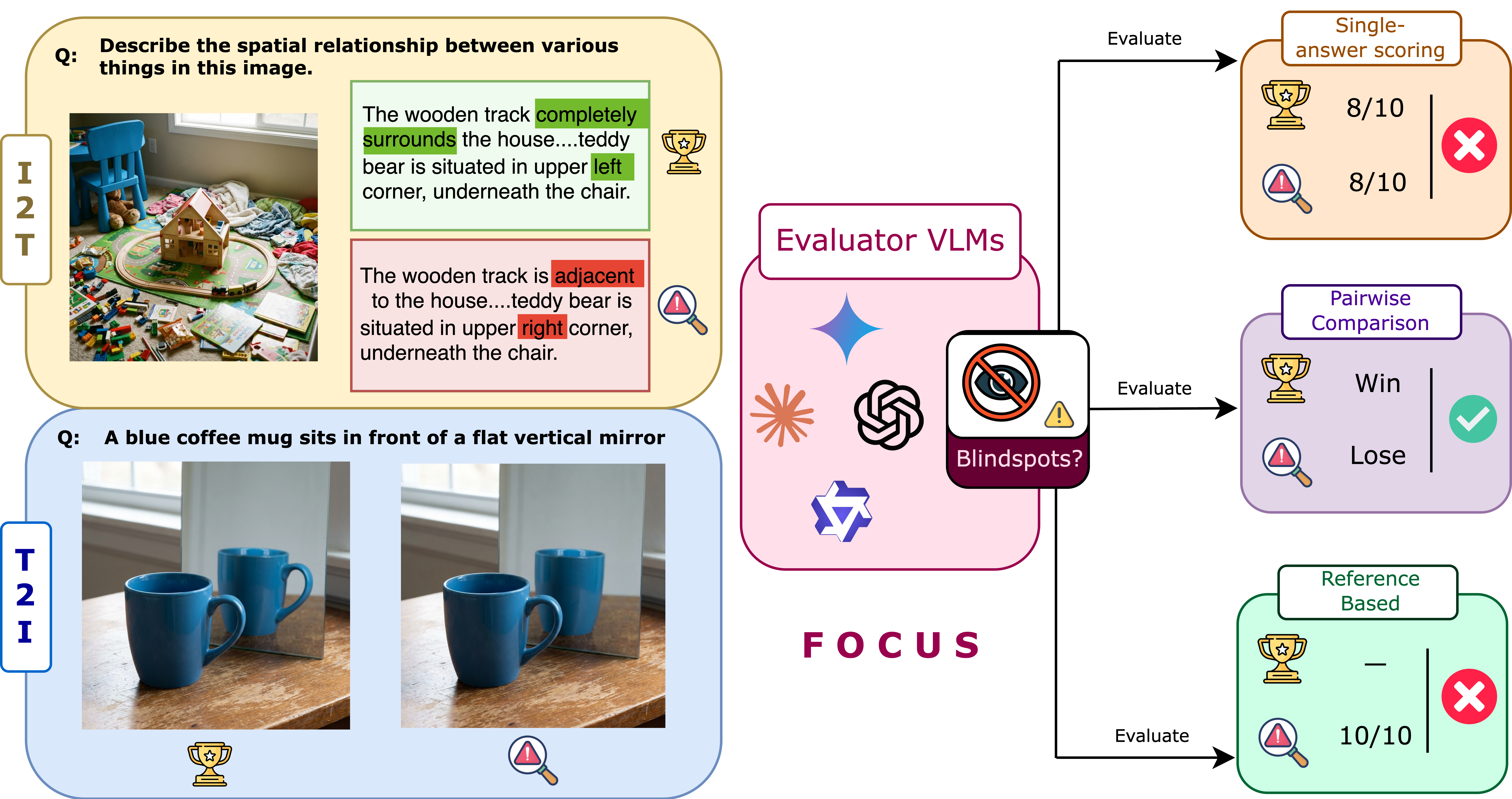}
    \caption{\bench~is a meta-evaluation benchmark to evaluate robustness of Evaluator VLMs.}
    \label{fig:category_by_paradigm}
\end{figure}

Using this benchmark, we evaluate four prominent VLMs under three widely adopted evaluation paradigms: single-answer scoring, pairwise comparison, and reference-guided evaluation. Within each paradigm, we explore multiple prompting strategies actually proposed in existing literature and used in practice~\citep{ge2023mllmbench, chen2024mjbench, pu2025judge}. \emph{Our findings paint a slightly sobering picture of current VLM-based evaluation}. Evaluators frequently fail to detect quality-degrading perturbations, in some cases exceeding 50\%, with performance notably worse for \ttoi~than \itot~, suggesting that current evaluators struggle with fine-grained visual understanding. Pairwise comparison emerges as the most reliable paradigm, contrasting with text-only settings where reference-based evaluation performed best \citep{doddapaneni2024finding}. Failures are concentrated in categories requiring fine-grained visual grounding, compositional reasoning, and physical plausibility, and increasing reasoning budgets does not consistently help. We also notice that evaluators sometimes identify errors in justifications but fail to reflect them in final scores.

These findings have broader implications beyond evaluation. As VLMs are increasingly used as reward models during training, the blind spots we uncover suggest that reward signals from these evaluators may fail to penalize critical errors. This could inadvertently reinforce the very behaviors they should correct. Our results highlight significant \textit{blindspots} in current Evaluator VLMs and caution against \textit{blind} reliance on them as standalone judges.

\section{\bench~benchmark}
\label{sec: bench}
We introduce \bench, a meta-evaluation benchmark for assessing the reliability of Evaluator Vision Language Models (VLMs), or \textit{evaluators}. Specifically, \bench~evaluates how well evaluators assess the outputs of other VLMs and image generation models, hereby called as \textit{evaluatees}. \bench~consists of two splits: \itot~(Image-to-Text) and \ttoi~(Text-to-Image).

The \itot~split covers tasks in which the \textit{evaluatee} takes an image and a text prompt as input and produces a text response, such as Visual Question Answering (VQA) and image captioning. Each instance in this split is represented as a tuple ($I$, $T$, $A_{gold}$, $A_{perturb}$), where $I$ is the input image, $T$ is the input text prompt, $A_{gold}$ is the correct (or \textit{gold}) answer, and $A_{perturb}$ is a \textit{perturbed} version of the gold answer. Similarly, the \ttoi~split covers tasks in which the \textit{evaluatee} takes a text prompt as input and produces an image, such as text-to-image generation. Each instance in this split is represented as a tuple ($T$, $I_{gold}$, $I_{perturb}$), where $T$ is the input prompt, $I_{gold}$ is the \textit{gold} image, and $I_{perturb}$ is a \textit{perturbed} version of the gold image.

In both splits, the perturbed output is created by introducing controlled errors along different perturbation dimensions (Table \ref{tab: pert_cat}). This setup allows us to evaluate whether evaluator VLMs can reliably account for these errors during evaluation. Each instance in \bench~is created with human oversight \textit{throughout} the creation process, including prompt selection ($\S$\ref{sec: prompt_select}), definition of perturbation dimensions ($\S$\ref{sec: pert_categories}), and perturbation creation ($\S$\ref{sec: pert_Generation}). Detailed statistics of \bench~are presented in Table \ref{tab:perturbation-distribution} and we describe the benchmark creation process below.

\subsection{Prompt selection}
\label{sec: prompt_select}
Each instance of \bench~is sourced from recent popular evaluation benchmarks. For \itot, we manually sampled 600 instances (text prompts paired with images) from seven recent benchmarks including MMBench~\citep{liu2023mmbench}, MMDocBench~\citep{Zhu2024MMDocBenchBL} and RealWorldQA\footnote{\url{https://huggingface.co/datasets/xai-org/RealworldQA}}. We selected these benchmarks for their focus on open-ended generation tasks, an important use case for Evaluator VLMs. Similarly, for \ttoi, we manually sampled 750 instances (text prompts) from seven widely used image generation benchmarks including MJ-Bench~\citep{chen2024mjbench}, T2I-CompBench++~\citep{huang2023t2i0compbench000} and T2I-ReasonBench~\citep{sun2025t2i0reasonbench0}. Details of benchmarks is provided in Appendix \ref{app: bench_details}.

The gold answers ($A_{gold}$) and gold images ($I_{gold}$) were generated using \gemini~and \vgemini~models, respectively and manually reviewed for quality and correctness. \textit{Importantly, we note that perfectly accurate gold answers and images are not essential for our study, as our primary focus is on directional score changes}. Specifically, we test whether perturbed responses with clear errors receive lower scores than their corresponding original responses. Thus, it is sufficient for the gold outputs to be reasonably accurate and relevant.

\subsection{Perturbation categories}
\label{sec: pert_categories}
Both VLMs and image generation models exhibit diverse failure modes documented in prior work~\citep{huang2023t2i0compbench000, meng2024phybench, Jing2023FaithScoreFE, Bai2024HallucinationOM}. A robust evaluator should reliably detect these errors and account for them during evaluation. We therefore design perturbations grounded in these failure modes and error dimensions. Detailed descriptions of all perturbation dimensions and examples are provided in Appendix \ref{app: pert_details}. For the \itot~split, we group perturbation dimensions into four broad categories:

\setlength\fboxsep{1.0pt}
\noindent \colorbox{i2t_vg}{\textbf{Visual Grounding (VG)}:}~Perturbations that modify directly observable visual elements such as entities, attributes, spatial relations, or object presence. For example, replacing \textit{"a dalmatian sitting on the grass"} with \textit{"a labrador sitting on the grass"}.

\noindent \colorbox{i2t_si}{\textbf{Semantic Interpretation (SI)}:}~Perturbations that degrade contextual or semantic understanding by altering cultural cues, contextual meaning, or introducing subtle inconsistencies. For example, replacing \textit{"celebrating Diwali with diyas"} with \textit{"celebrating Diwali with candles"}.

\noindent \colorbox{i2t_vr}{\textbf{Visual Reasoning (VR)}:}~Perturbations that introduce logical, numerical, or causal errors. For example, replacing \textit{"population increased by 15\%"} with \textit{"population increased by 12\%"}.

\noindent \colorbox{i2t_lg}{\textbf{Long-form Generation (LG)}:}~Perturbations that introduce inconsistencies between fluent long-form text and the underlying visual content. For example, describing \textit{"a knight riding under the bright moon"} when the image depicts a daytime scene.

Similarly, for the \ttoi~split, we define four complementary perturbation categories:

\setlength\fboxsep{1.0pt}
\noindent \colorbox{t2i_vf}{\textbf{Visual Fidelity (VF)}:}~Perturbations that alter key visual elements of the generated image such as objects, attributes, or spatial layouts. For example, rendering \textit{a red car} as \textit{a blue car}.

\noindent \colorbox{t2i_sc}{\textbf{Scene Coherence (SC)}:}~Perturbations that degrade scene-level consistency through stylistic mismatches, incomplete rendering, or contextual inconsistencies in the generated image. For example, generating \textit{photorealistic humans in a flat 2D cartoon environment}.

\noindent \colorbox{t2i_pp}{\textbf{Physical Plausibility (PP)}:}~Perturbations introducing violations of physical laws, causal logic, or common sense. For example, \textit{shadows pointing towards the light source}.

\noindent \colorbox{t2i_tr}{\textbf{Text Rendering (TR)}:}~Perturbations that corrupt textual or symbolic elements in a generated image. For example, \textit{``COEFEE''} instead of \textit{``COFFEE''} on a shop sign.

\definecolor{corrgreen}{HTML}{008000} 
\definecolor{pertred}{HTML}{D2042D}   

\begin{table}[t]
\centering
\fontsize{7}{8}\selectfont
\setlength{\tabcolsep}{0.5pt}
\renewcommand{\arraystretch}{1.2}
\setlength{\fboxsep}{1pt}

\newcommand{\eg}[2]{ \textit{Eg: #1 $\rightarrow$ #2}}
\newcommand{\cgn}[1]{\textcolor{corrgreen}{#1}}
\newcommand{\prd}[1]{\textcolor{pertred}{#1}}

\begin{tabularx}{\linewidth}{@{}m{1cm} m{3.3cm} >{\raggedright\arraybackslash}X@{}}
\toprule
\textbf{Catg} & \textbf{Perturbation Dimension} & \textbf{Perturbation Description} \\
\midrule

\multicolumn{3}{c}{\textbf{\textit{Image-to-Text (I2T)}}} \\
\midrule

\multirow{6}{*}{\centering \textbf{\colorbox{i2t_vg}{VG}}}
& Entity Substitution & Swaps with a similar but incorrect entity. \eg{The chef holds a \cgn{knife}}{The chef holds a \prd{cleaver}} \\
& Attribute Distortion & Changes subtle attributes like color or texture. \eg{A \cgn{red} car is parked}{A \prd{blue} car is parked} \\
& Spatial Relation Swap & Alters relative positioning of objects. \eg{A book is \cgn{under} a table}{A book is \prd{on top of} a table} \\
& Phantom Details Injection & Introduces non-existent objects. \eg{A park has trees}{A park has trees and a \prd{statue}} \\
& Over Generalization & Replaces with broader hypernyms. \eg{A woman is in a \cgn{Tesla}}{A woman is in a \prd{vehicle}} \\
& Important Detail Omission & Removes essential grounding elements. \eg{a \cgn{red striped} hat on a table}{a hat on a table} \\
\addlinespace[3pt]
\midrule
\multirow{3}{*}{\centering \textbf{\colorbox{i2t_si}{SI}}}
& Contextual Depth Reduction & Removes implicit intent or nuance. \eg{A \cgn{contemplative} man sitting}{A \prd{bored} man sitting} \\
& Cultural Misalignment & Replaces cultural markers incorrectly. \eg{A person in a \cgn{kimono}}{A person in a \prd{sari}} \\
& Logical Inconsistencies & Introduces contradictions in the statement. \textit{Eg: The \prd{open} and closed bridge nearby.} \\
\midrule

\multicolumn{3}{c}{\textbf{\textit{Text-to-Image (T2I)}}} \\
\midrule

\multirow{6}{*}{\centering \textbf{\colorbox{t2i_vf}{VF}}}
& Object Substitution & Replaces the primary object in the scene. \eg{A \cgn{cat}}{A \prd{dog}} \\
& Object Addition/Omission & Alters object presence or quantity in the scene. \eg{One \cgn{chair}}{\prd{many} chairs} \\

& Attribute Manipulation & Changes attributes such as color, texture, or size. \eg{\cgn{red} ball}{\prd{blue} ball} \\
& Spatial Manipulation & Alters object position or relative spatial arrangement. \eg{Cup \cgn{on} table}{Cup \prd{under} table} \\
& Scale Distortion & Changes object proportions or relative scale relationships. \eg{\cgn{small} mouse}{\prd{large} mouse} \\
& Constraint Violation & Violates explicit prompt constraints or specified conditions. \eg{No cars}{\prd{car} present} \\
\addlinespace[3pt]
\midrule
\multirow{5}{*}{\centering \textbf{\colorbox{t2i_pp}{PP}}}
& Causal Violation & Breaks expected cause-effect relationships between events. \eg{Glass falls \cgn{breaks}}{\prd{intact}} \\

& Physics Manipulation & Violates basic physical laws or natural behavior. \eg{Shadow \cgn{away}}{\prd{towards}} light \\
& State/Transformation Failure & Produces incorrect or incomplete transformation outcomes. \eg{Ice \cgn{melts}}{\prd{unchanged}} \\
& Functional Absurdity & Depicts objects being used in illogical ways. \eg{Knife cuts}{Knife used on \prd{stone}} \\
& Literalized Idioms & Interprets figurative expressions in a literal visual form. \eg{Heavy rain}{\prd{objects falling}} \\

\bottomrule
\end{tabularx}

\caption{Select perturbation categories, dimensions and examples. Original elements are shown in \cgn{green} and perturbed elements in \prd{red}. See Table \ref{tab:pert_cat} for full list.}
\label{tab: pert_cat}
\end{table}

\subsection{Perturbation generation}
\label{sec: pert_Generation}
To generate perturbed responses across categories, we adopt a two-step process: automatic perturbation generation followed by thorough human verification. This approach allows us to efficiently create diverse perturbations while ensuring high quality and correctness. For \itot~tasks, we prompt \gemini~~$f(\cdot)$ with perturbation-specific instructions ($P_{perturb}$), together with the input image ($I$), text prompt ($T$), and gold answer ($A_{gold}$). The model generates a perturbed answer ($A_{perturb}$) along with a short description ($descr$) of the introduced error. Formally, this process can be represented as: $f(P_{perturb}, I, T, A_{gold}) \rightarrow (A_{perturb}, descr)$. For \ttoi~tasks, we first prompt \gemini~~$f(\cdot)$ with perturbation-specific instructions ($P_{perturb}$), along with the text prompt ($T$) and the gold image ($I_{gold}$) to generate an edit instruction. Using the generated edit instruction, we prompt \vgemini~~$g(\cdot)$ to edit the gold image, and produce the perturbed image ($I_{perturb}$). Formally, this process is represented as: $g(I_{gold}, f(P_{perturb}, T, I_{gold})) \rightarrow (I_{perturb})$.


Although this automatic pipeline produces strong perturbations, all perturbed instances ($A_{perturb}$ and $I_{perturb}$) are further reviewed by human annotators, including the authors. Each perturbed response is compared with the corresponding gold response and labeled as \textit{valid}, \textit{invalid}, or \textit{score-invariant}. A perturbation is marked as \textit{valid} if it introduces a subtle but meaningful error that \emph{should} receive a lower score than the gold response. It is marked as \textit{invalid} if it is overly obvious or nonsensical in context. Following \citet{doddapaneni2024finding}, perturbations that should not receive a score penalty, such as paraphrases of the original response or image edits that do not contradict the input prompt, are categorized as \textit{score-invariant}. To support this process, we developed a custom annotation tool. Details of the tool and annotation workflow are provided in Appendix \ref{app: app}.

\section{Experimental Setup}



In this section, we first describe the prompting strategies used to benchmark Evaluator VLMs on \bench, and then discuss the evaluation metrics. An Evaluator VLM, $f(\cdot)$, takes as input an evaluation instruction ($P_{eval}$), the task input ($T$ and/or $I$), and the response(s) to be evaluated ($I$ or $A$), and produces a judgment along with a supporting justification. Based on existing literature, we focus on the three most commonly used evaluation paradigms: (i) Single-answer scoring, (ii) Pairwise comparison, and (iii) Reference-guided evaluation. For each paradigm, we explore commonly used prompting strategies proposed in prior works~\citep{chen2024mllmasajudge, pu2025judge, lin2025self0improving, li2024vlrewardbench0, yang2025probench0, chen2024mjbench, cui2024exploring}. Our setups are directly inspired by existing literature, since our goal is to evaluate VLM Evaluators under commonly used settings rather than propose new evaluation strategies. The exact prompts and details for each evaluator are listed in the Appendix~\ref{app: eval_details}.

Within each paradigm, we explore four prompting strategies that progressively add structure: \textbf{Vanilla [V]} (input + output only), \textbf{Rubric/Rules [R]} (adds a grading rubric or rule set), \textbf{Axes [Ax]} (evaluation along predefined axes), and \textbf{Axes+Rubric/Rules [Ax+R]} (axes with per-axis rubrics or rules). We use $O_{model}$ to denote a single evaluatee output and $O_1, O_2$ for candidate pairs. Table~\ref{tab:paradigms} summarizes the input--output signatures for all strategies.

\begin{table}[t]
\centering
\small
\renewcommand{\arraystretch}{1.15}
\begin{tabular}{@{}llll@{}}
\toprule
\textbf{Paradigm} & \textbf{Strategy} & \textbf{Input to $f(\cdot)$} & \textbf{Output} \\
\midrule
\multirow{4}{*}{\shortstack[l]{Single-answer\\scoring }}
& V & $P_{eval},\; T,\; I_{in},\; O_{model}$ & $(score,\; just)$ \\
& R & $P_{eval},\; R,\; T,\; I_{in},\; O_{model}$ & $(score,\; just)$ \\
& Ax & $P_{eval},\; [Ax],\; T,\; I_{in},\; O_{model}$ & $([score],\; just)$ \\
& Ax+R & $P_{eval},\; [\{Ax, R\}],\; T,\; I_{in},\; O_{model}$ & $([score],\; just)$ \\
\midrule
\multirow{4}{*}{\shortstack[l]{Pairwise\\comparison }}
& V & $P_{eval},\; T,\; I_{in},\; O_1,\; O_2$ & $(verdict,\; exp)$ \\
& R & $P_{eval},\; R,\; T,\; I_{in},\; O_1,\; O_2$ & $(verdict,\; exp)$ \\
& Ax & $P_{eval},\; [Ax],\; T,\; I_{in},\; O_1,\; O_2$ & $([verdict],\; exp)$ \\
& Ax+R & $P_{eval},\; [\{Ax, R\}],\; T,\; I_{in},\; O_1,\; O_2$ & $([verdict],\; exp)$ \\
\midrule
\shortstack[l]{Reference-guided\\scoring}
& Ref & $P_{eval},\; T,\; I_{in},\; O_{gold},\; O_{model}$ & $(score,\; just)$ \\
\bottomrule
\end{tabular}
\caption{Evaluation paradigms and prompting strategies. $I_{in}$ is included only for \itot. $R$ denotes a rubric (single-answer) or rules (pairwise); $[Ax]$ denotes predefined evaluation axes. Brackets around output (e.g., $[score]$) indicate per-axis judgments.}
\label{tab:paradigms}
\end{table}

\label{sec: single_answer}
\label{sec: pairwise}
\label{sec: reference}

\noindent\textbf{Single-answer scoring.}\label{sec: single_answer}~~In this paradigm, the evaluator scores a single response independently based on the provided input and its parametric knowledge. For \itot~tasks, the evaluator receives the text prompt $T$, input image $I_{in}$, and generated response $A_{model}$; for \ttoi~tasks, it receives $T$ and the generated image $I_{model}$. This is the most commonly used paradigm in practice~\citep{cui2024exploring, chen2024mjbench, pu2025judge}.

\noindent\textbf{Pairwise comparison.}\label{sec: pairwise}~~Here, the evaluator is tasked with selecting the better response between two candidates. For \itot, it receives $T$, $I_{in}$, and two candidate responses $A_1$ and $A_2$; for \ttoi, it receives $T$ and two generated images $I_1$ and $I_2$. This paradigm is widely used in preference-based evaluation and reward modeling~\citep{lin2025self0improving, chen2024mllmasajudge}.

\noindent\textbf{Reference-guided scoring.}\label{sec: reference}~~In this paradigm, the evaluator scores a model output by comparing it against a reference (gold) output $O_{gold}$. This provides the evaluator with an explicit quality anchor, potentially simplifying judgment. However, \textit{this approach may not be feasible for many open-ended tasks} where good references are not easily available. For \itot, the reference is a gold answer $A_{gold}$; for \ttoi, it is a gold image $I_{gold}$.

\subsection{Metrics}
\label{sec: metrics}
In the single-answer scoring paradigm, we measure the percentage of instances in which the score remains unchanged after perturbation. For evaluators that score along multiple axes, we consider only the axes relevant to the perturbation category. A perturbation is counted as undetected only if the scores on all relevant axes remain unchanged. Details about the axes are in Appendix \ref{app: eval_details}. Ideally, the evaluator should assign a lower score to the perturbed output. In the pairwise comparison paradigm, we present the \textit{gold} output alongside the \textit{perturbed} output and ask the evaluator to choose the better one. Our metric is the percentage of instances in which the evaluator fails to select the gold output alone. For axis-based pairwise evaluators, we again restrict the analysis to the axes relevant to the perturbation category. To mitigate position bias~\citep{llm-judge}, we run each evaluation twice, swapping the order of the gold and perturbed outputs. In reference-guided scoring, the gold output is used as the reference, and we measure the percentage of instances in which the evaluator assigns a perfect score to the perturbed output. We use these metrics for both \itot\ and \ttoi\ tasks, since the evaluation paradigms are structurally identical.
\section{Results and discussion}

We evaluate several frontier VLMs, including \gemini{ \raisebox{-0.15em}{\includegraphics[height=1em]{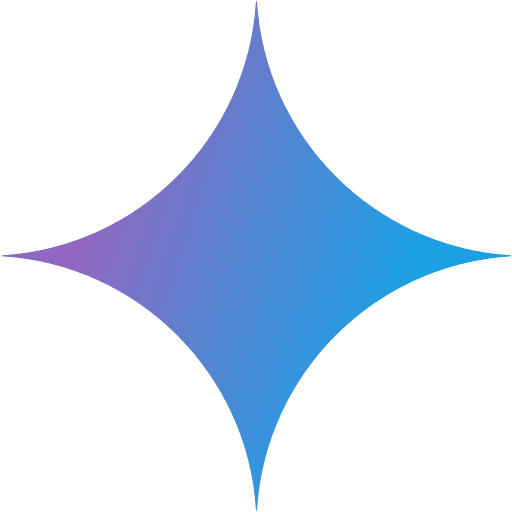}}}, \gpt{ \raisebox{-0.15em}{\includegraphics[height=1em]{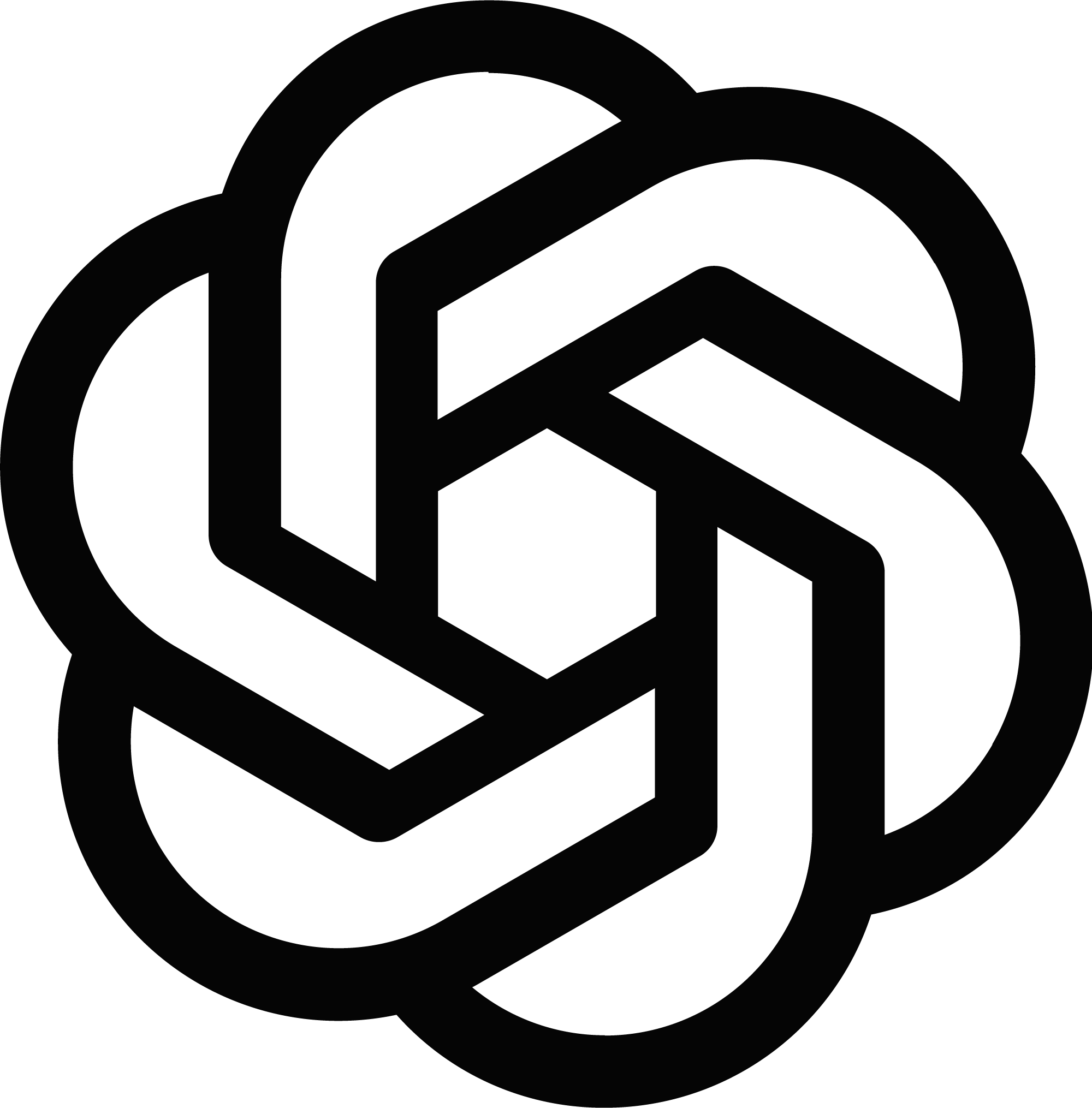}}}, \claude{ \raisebox{-0.15em}{\includegraphics[height=0.9em]{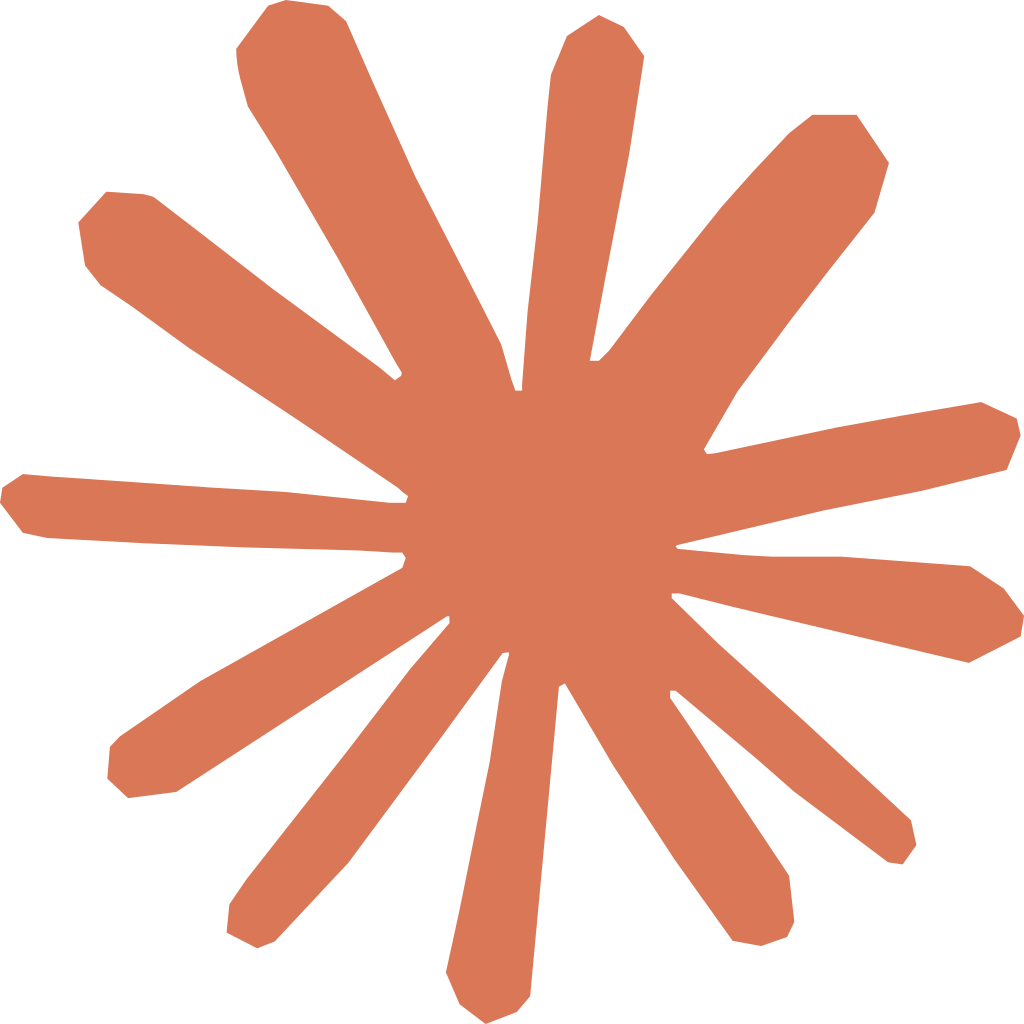}}}, and \qwen{ \raisebox{-0.15em}{\includegraphics[height=1.0em]{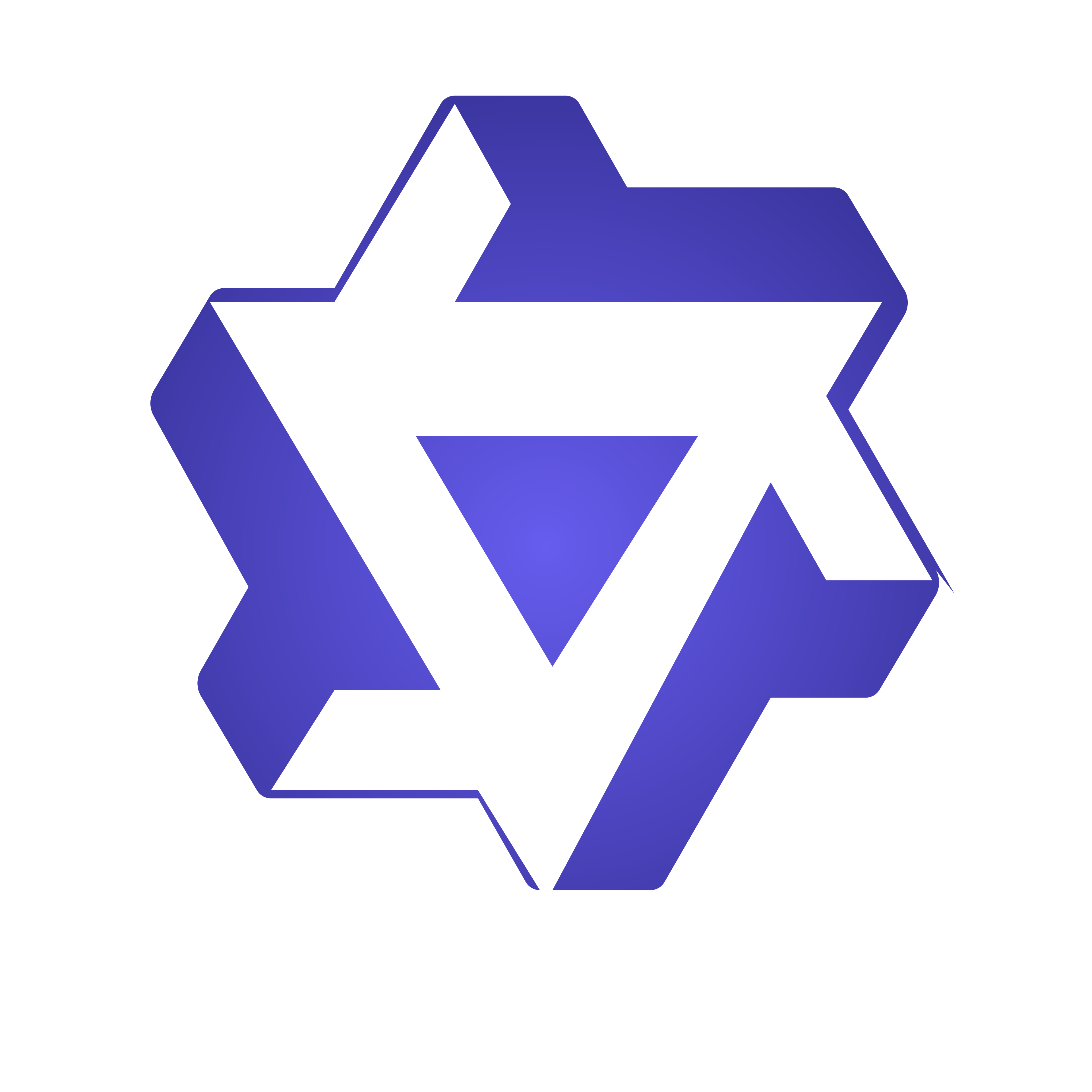}}}. To ensure a fair comparison, we use identical evaluation prompts across all models and settings. Since these models support reasoning, we use the highest available reasoning setting. Wherever supported, we also set the sampling temperature to 0 to improve reproducibility and reduce variance across runs.

\subsection{Which evaluator paradigms and strategies are most reliable?}
\begin{table}[t]
\centering
\small
\resizebox{0.99\columnwidth}{!}{%
\begin{tabular}{@{}lcccccccc|cccccccc@{}}
\toprule
\multirow{3}{*}{{$\mathbf{f(\cdot)}$}} & \multicolumn{8}{c|}{\textbf{I2T}} & \multicolumn{8}{c}{\textbf{T2I}} \\ \cmidrule(l){2-17} 

 & \multicolumn{2}{c}{\includegraphics[width=0.4cm]{figures/logos/gemini.png}} & \multicolumn{2}{c}{\includegraphics[width=0.4cm]{figures/logos/openai.png}} & \multicolumn{2}{c}{\includegraphics[width=0.4cm]{figures/logos/claude.png}} & \multicolumn{2}{c|}{\includegraphics[width=0.4cm]{figures/logos/qwen.png}} & \multicolumn{2}{c}{\includegraphics[width=0.4cm]{figures/logos/gemini.png}} & \multicolumn{2}{c}{\includegraphics[width=0.4cm]{figures/logos/openai.png}} & \multicolumn{2}{c}{\includegraphics[width=0.4cm]{figures/logos/claude.png}} & \multicolumn{2}{c}{\includegraphics[width=0.4cm]{figures/logos/qwen.png}} \\

 & \textbf{VP$\downarrow$} & \textbf{SI$\uparrow$} & \textbf{VP$\downarrow$} & \textbf{SI$\uparrow$} & \textbf{VP$\downarrow$} & \textbf{SI$\uparrow$} & \textbf{VP$\downarrow$} & \textbf{SI$\uparrow$} & \textbf{VP$\downarrow$} & \textbf{SI$\uparrow$} & \textbf{VP$\downarrow$} & \textbf{SI$\uparrow$} & \textbf{VP$\downarrow$} & \textbf{SI$\uparrow$} & \textbf{VP$\downarrow$} & \textbf{SI$\uparrow$} \\ \midrule

 & \multicolumn{16}{c}{\textit{Single answer scoring}} \\
V & \cellcolor[HTML]{8FD1B1}32.3 & \cellcolor[HTML]{FFDE81}43.4 & \cellcolor[HTML]{E5F4ED}44.1 & \cellcolor[HTML]{FFD96F}47.8 & \cellcolor[HTML]{D3EDE0}41.6 & \cellcolor[HTML]{FFDE82}43.1 & \cellcolor[HTML]{CBEADB}40.6 & \cellcolor[HTML]{FFD666}50 & \cellcolor[HTML]{B8E2CD}46.2 & \cellcolor[HTML]{FFEAAF}54.7 & \cellcolor[HTML]{EBF7F1}53.6 & \cellcolor[HTML]{FFDE84}62.7 & \cellcolor[HTML]{E8F5EF}53.1 & \cellcolor[HTML]{FFDD7D}64.0 & \cellcolor[HTML]{F4FAF7}54.8 & \cellcolor[HTML]{FFD666}68.3 \\
Ax & \cellcolor[HTML]{6EC49A}27.8 & \cellcolor[HTML]{FFE69F}36.0 & \cellcolor[HTML]{71C59C}28.2 & \cellcolor[HTML]{FFEEBD}28.4 & \cellcolor[HTML]{57BB8A}24.6 & \cellcolor[HTML]{FFF0C6}26.3 & \cellcolor[HTML]{96D4B6}33.3 & \cellcolor[HTML]{FFDF85}42.5 & \cellcolor[HTML]{5DBD8E}33.0 & \cellcolor[HTML]{FFFAEC}43.3 & \cellcolor[HTML]{63C092}34.0 & \cellcolor[HTML]{FFFBED}43.0 & \cellcolor[HTML]{9CD7BA}42.2 & \cellcolor[HTML]{FFEFC2}51.2 & \cellcolor[HTML]{BAE3CF}46.5 & \cellcolor[HTML]{FFE292}60.2 \\
R & \cellcolor[HTML]{AEDEC7}36.6 & \cellcolor[HTML]{FFDF87}42.0 & \cellcolor[HTML]{EAF6F0}44.8 & \cellcolor[HTML]{FFDA75}46.3 & \cellcolor[HTML]{FFFFFF}47.6 & \cellcolor[HTML]{FFD96F}48 & \cellcolor[HTML]{CDEADC}40.8 & \cellcolor[HTML]{FFDB78}45.7 & \cellcolor[HTML]{B8E2CE}46.2 & \cellcolor[HTML]{FFE7A5}56.6 & \cellcolor[HTML]{E3F3EB}52.3 & \cellcolor[HTML]{FFDC7C}64.3 & \cellcolor[HTML]{FFFFFF}56.3 & \cellcolor[HTML]{FFDD7D}64.0 & \cellcolor[HTML]{F0F9F4}54.2 & \cellcolor[HTML]{FFD86D}67 \\
Ax+R & \cellcolor[HTML]{6FC49A}27.9 & \cellcolor[HTML]{FFE59D}36.5 & \cellcolor[HTML]{64C093}26.4 & \cellcolor[HTML]{FFF1C8}25.8 & \cellcolor[HTML]{57BB8A}24.6 & \cellcolor[HTML]{FFEFC2}27.2 & \cellcolor[HTML]{B6E1CC}37.6 & \cellcolor[HTML]{FFDB76}46.1 & \cellcolor[HTML]{57BB8A}32.1 & \cellcolor[HTML]{FFFFFF}39.6 & \cellcolor[HTML]{66C194}34.3 & \cellcolor[HTML]{FFFBF0}42.5 & \cellcolor[HTML]{7ECBA5}37.9 & \cellcolor[HTML]{FFFAEC}43.3 & \cellcolor[HTML]{C3E6D5}47.7 & \cellcolor[HTML]{FFE7A3}56.9 \\ \midrule
 & \multicolumn{16}{c}{\textit{Pairwise comparison}} \\
V & \cellcolor[HTML]{6DC399}13.1 & \cellcolor[HTML]{FFFDF5}14.6 & \cellcolor[HTML]{94D4B5}15.7 & \cellcolor[HTML]{FFF9E6}18.3 & \cellcolor[HTML]{FFFFFF}22.8 & \cellcolor[HTML]{FFFFFC}12.8 & \cellcolor[HTML]{7ECBA5}14.3 & \cellcolor[HTML]{FFFDF7}14 & \cellcolor[HTML]{D5EEE1}31.5 & \cellcolor[HTML]{FFE49A}27.1 & \cellcolor[HTML]{D8EFE3}31.8 & \cellcolor[HTML]{FFE18F}28.8 & \cellcolor[HTML]{FFFFFF}36.1 & \cellcolor[HTML]{FFD666}35.1 & \cellcolor[HTML]{E0F2E9}32.8 & \cellcolor[HTML]{FFF0C6}20.3 \\
Ax & \cellcolor[HTML]{57BB8A}11.6 & \cellcolor[HTML]{FFFFFD}12.7 & \cellcolor[HTML]{94D3B4}15.7 & \cellcolor[HTML]{FFF4D6}22.2 & \cellcolor[HTML]{82CCA8}14.5 & \cellcolor[HTML]{FFFFFE}12.3 & \cellcolor[HTML]{76C79F}13.7 & \cellcolor[HTML]{FFFEF9}13.7 & \cellcolor[HTML]{57BB8A}17.7 & \cellcolor[HTML]{FFFBED}14.2 & \cellcolor[HTML]{94D3B4}24.4 & \cellcolor[HTML]{FFFFFE}11.7 & \cellcolor[HTML]{8DD0AF}23.6 & \cellcolor[HTML]{FFFFFF}11.4 & \cellcolor[HTML]{A1D9BE}25.9 & \cellcolor[HTML]{FFF2CE}19 \\
R & \cellcolor[HTML]{6FC59B}13.3 & \cellcolor[HTML]{FFFBEF}16.0 & \cellcolor[HTML]{8ED1B0}15.3 & \cellcolor[HTML]{FFF9E8}17.9 & \cellcolor[HTML]{DCF0E6}20.5 & \cellcolor[HTML]{FFF6DB}21 & \cellcolor[HTML]{74C69E}13.6 & \cellcolor[HTML]{FFFDF5}14.6 & \cellcolor[HTML]{E9F6F0}33.8 & \cellcolor[HTML]{FFDD80}31.1 & \cellcolor[HTML]{D9EFE4}32.0 & \cellcolor[HTML]{FFDF86}30.2 & \cellcolor[HTML]{F0F9F4}34.5 & \cellcolor[HTML]{FFDE84}30.5 & \cellcolor[HTML]{DDF1E7}32.4 & \cellcolor[HTML]{FFF5D7}17.7 \\
Ax+R & \cellcolor[HTML]{5BBC8D}11.9 & \cellcolor[HTML]{FFFFFF}12.0 & \cellcolor[HTML]{AADCC3}17.2 & \cellcolor[HTML]{FFF4D5}22.5 & \cellcolor[HTML]{9DD7BB}16.3 & \cellcolor[HTML]{FFFCF4}14.9 & \cellcolor[HTML]{85CDAA}14.7 & \cellcolor[HTML]{FFFCF4}14.9 & \cellcolor[HTML]{59BC8C}18.0 & \cellcolor[HTML]{FFFDF5}13 & \cellcolor[HTML]{98D5B7}24.8 & \cellcolor[HTML]{FFFFFF}11.4 & \cellcolor[HTML]{97D5B6}24.7 & \cellcolor[HTML]{FFF9E8}15.0 & \cellcolor[HTML]{9CD6BA}25.2 & \cellcolor[HTML]{FFF6DB}17 \\ \midrule
 & \multicolumn{16}{c}{\textit{Reference guided scoring}} \\
Ref & \cellcolor[HTML]{57BB8A}15.1 & \cellcolor[HTML]{FFFAEA}17.3 & \cellcolor[HTML]{B2DFC9}18.7 & \cellcolor[HTML]{FFF3CF}24 & \cellcolor[HTML]{CFEBDD}19.9 & \cellcolor[HTML]{FFF4D4}22.7 & \cellcolor[HTML]{FFFFFF}21.8 & \cellcolor[HTML]{FFF0C7}26.1 & \cellcolor[HTML]{57BB8A}21.7 & \cellcolor[HTML]{FFE089}30.4 & \cellcolor[HTML]{FFFFFF}30.0 & \cellcolor[HTML]{FFD768}35.6 & \cellcolor[HTML]{D4EDE1}27.9 & \cellcolor[HTML]{FFD666}35.9 & \cellcolor[HTML]{D2ECDF}27.8 & \cellcolor[HTML]{FFDB79}33 \\
\bottomrule
\end{tabular}%
}
\caption{Comparison of evaluator paradigms and strategies on \itot~and \ttoi. The numbers indicate the percentage of instances where the score/verdict generated by the evaluator is not affected by the perturbation. \textbf{VP} denotes valid perturbations, and \textbf{SI} denotes score-invariant perturbations. For columns marked with $\downarrow$, lower values and darker green are better; for columns marked with $\uparrow$, higher values and darker yellow are better.}
\label{tab: results_main}
\end{table}
Referring to the \textbf{VP} columns (green cells) in Table \ref{tab: results_main}, we observe a clear and consistent pattern across both \itot~ and \ttoi~: pairwise comparison emerges as the \textit{most reliable} evaluation paradigm, while single-answer scoring is the weakest. In \itot~ (left side of Table \ref{tab: results_main}), all pairwise strategies substantially outperform their single-answer counterparts, with the best performance achieved by the Axes and Axes+Rules strategies. This trend is even more pronounced in \ttoi~ (right side of Table \ref{tab: results_main}), where single-answer scoring shows high failure rates, exceeding 50\% in some cases. These results suggest that relative judgments between two candidates are more robust than scoring a single response in isolation. Interestingly, reference-guided evaluation improves over single-answer scoring but generally remains behind the best pairwise strategies. This contrasts previous findings in text-only evaluator LLMs, where reference-based evaluation performed best~\citep{doddapaneni2024finding}.

A second observation concerns the role of \textit{strategies} within each paradigm. In single-answer scoring, Axes-based strategies consistently perform better, indicating that explicitly defining evaluation dimensions improves reliability. In contrast, providing generic rubrics alone often degrades performance. Similarly for pairwise comparison too, structured strategies like Axes and Axes+Rules consistently achieve the best results, particularly in \ttoi~ where the gap over single answer scoring is substantial. Overall, these findings suggest a practical hierarchy: structured pairwise evaluation is the most reliable paradigm, reference-guided evaluation serves as a useful but weaker alternative, and single-answer scoring remains the least reliable even with additional strategies.

\subsection{How does evaluator performance vary across VLMs?}
Evaluator performance varies noticeably across VLMs (Table \ref{tab: results_main}), with trends depending on both the task and evaluation paradigm. The clearest separation appears in the pairwise comparison paradigm, where \gemini~consistently achieves the lowest failure rates across both \itot~and \ttoi. In contrast, \claude, despite being a strong model and often ranking highly on general leaderboards\footnote{\url{https://lmarena.ai/}}
, shows relatively higher failure rates across strategies. A similar ordering is observed in \ttoi, where \gemini~again performs best across most pairwise strategies, suggesting stronger ability in identifying relative quality differences between outputs.

The pattern is less uniform in the single-answer scoring paradigm. While \gemini~remains strong—particularly in \ttoi, where it consistently achieves the best performance across strategies, the \itot~setting is more mixed. Under more structured strategies such as Axes+Rubric, \claude~becomes competitive and in some cases performs comparably to \gemini. Across paradigms, \gpt~generally performs competitively but remains slightly behind \gemini, while \qwen~tends to show higher failure rates, especially in the more challenging \ttoi~setting. Overall, evaluator reliability varies significantly across tasks, paradigms, and prompting strategies, highlighting the importance of careful evaluator selection.
\begin{figure}[!t]
    \centering
    \includegraphics[width=1.0\columnwidth]{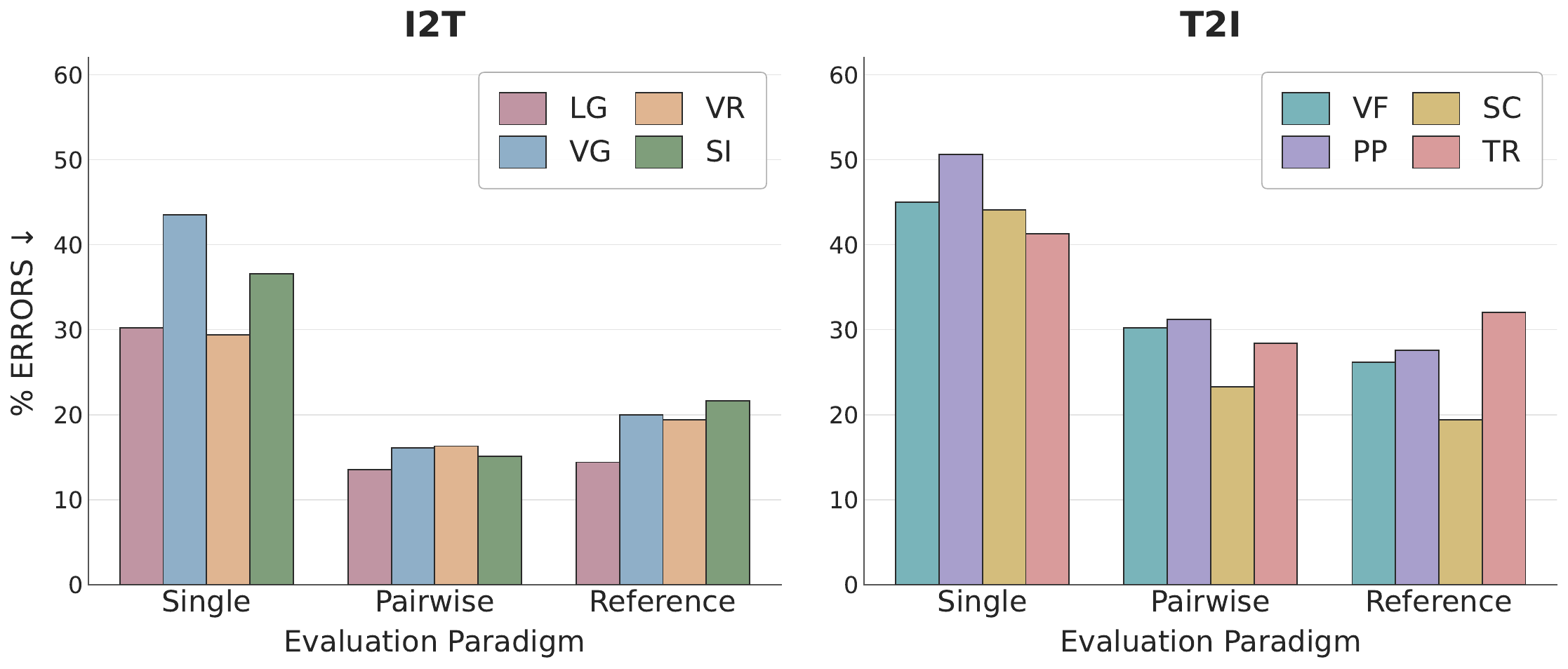}
    \caption{Comparing the performance of different evaluator paradigms across perturbation categories. Results are averaged across evaluator VLMs and strategies. Lower is better.}
    \label{fig:category_by_paradigm}
\end{figure}

\subsection{Which perturbation categories are hardest to detect?}


Figure~\ref{fig:category_by_paradigm} breaks down performance by perturbation category and evaluator paradigm, averaged across evaluator VLMs. In \itot, \textit{Visual Grounding} and \textit{Semantic Interpretation} are the most challenging categories under \textit{Single Scoring}, likely because they involve subtle mismatches in entities, attributes, or context despite otherwise fluent outputs. \textit{Pairwise Comparison} substantially reduces failures on these categories, suggesting that relative judgments make such errors easier to detect than independent scoring. In \ttoi, \textit{Physical Plausibility} is the hardest category under both \textit{Single Scoring} and \textit{Pairwise Comparison}, since these errors often require deeper reasoning about physics or common sense, whereas \textit{Scene Coherence} is the easiest because the inconsistencies are visually obvious. Surprisingly, under \textit{Reference}, \textit{Text Rendering} is particularly challenging. Overall, evaluator failures concentrate in categories that require fine-grained grounding, nuanced semantics, or deeper reasoning.

\subsection{Does the choice of reference matter?}

\begin{wraptable}[9]{r}{0.48\textwidth} 
    \centering
    \small
    \vspace{-33pt} 
    \begin{tabular}{@{}lcccc@{}}
        \toprule
        \textbf{Task} & \textbf{Model} & \textbf{Orig $\downarrow$} & \textbf{New $\downarrow$} & \textbf{$\Delta \downarrow$} \\ \midrule
        \multirow{2}{*}{I2T} & Gemini & 15.08 & 18.83 & 3.75 \\
                             & Qwen   & 21.75 & 23.1  & 1.35 \\ \midrule
        \multirow{2}{*}{T2I} & Gemini & 21.65 & 16.45 & -5.2 \\
                             & Qwen   & 27.8  & 17.28 & -10.53 \\ \bottomrule
    \end{tabular}
    \caption{Effect of reference variation. Orig. uses default gold reference, while New uses an alternative. Lower is better.}
    \label{tab:ref_brittleness}
\end{wraptable}

In our \textsc{Reference} setting, \textit{perturbed} responses differ from references (i.e., \textit{gold} responses) only through injected perturbations, making evaluation relatively easy. To truly test the robustness, we regenerate the references using a different sampling temperature, which produces paraphrased answers for \itot~and visually distinct but still correct images for \ttoi. As shown in Table \ref{tab:ref_brittleness}, reference variation affects the two tasks differently: performance drops slightly for \itot~but improves for \ttoi. The drop in \itot~suggests that text evaluators are sensitive to surface-level similarity between the candidate and the reference, whereas the gain in \ttoi~indicates that image evaluators benefit from semantically correct yet visually diverse references.

\begin{figure}[t]
    \centering
    \begin{minipage}{0.49\textwidth}
    
        \centering
        \includegraphics[trim=7pt 5pt 7pt 7pt, clip, width=\textwidth]{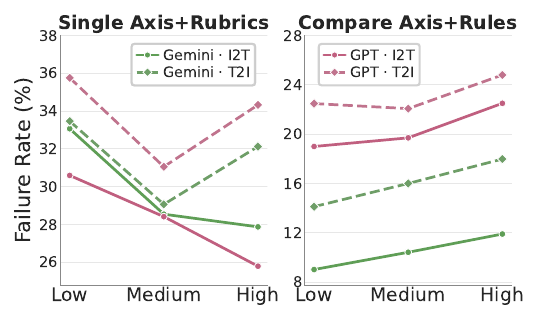}
        \caption{Effect of reasoning budget on evaluator performance. We plot the percentage of instances where the score/verdict of the evaluator is not affected by the perturbation - lower is better. }
        \label{fig:reasoning_budget}
    \end{minipage}
    \hfill 
    \begin{minipage}{0.49\textwidth}
    \vspace{-12pt}
        \centering
        \includegraphics[trim=7pt 5pt 7pt 7pt, clip, width=\textwidth]{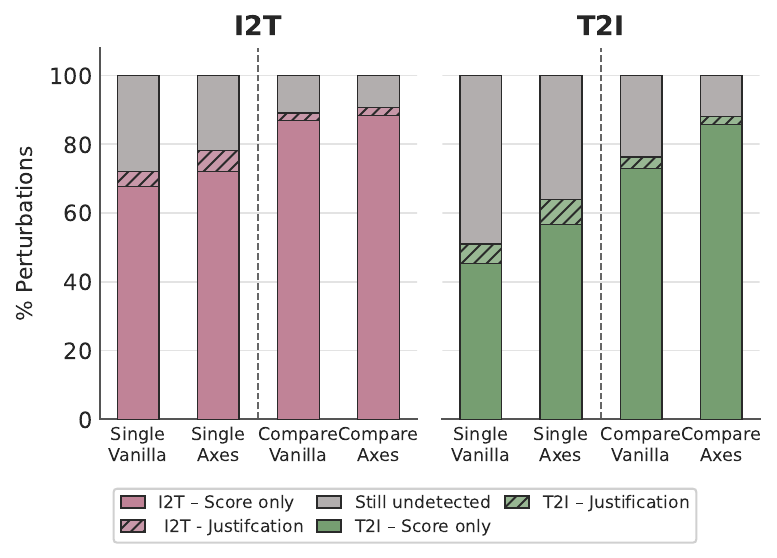}
        \caption{Comparison of perturbations detected solely by evaluator scores and those identified through justifications. The shaded region denotes perturbations detected in justifications but not reflected in scores.}
        \label{fig:justification}
    \end{minipage}
\end{figure}


\subsection{How does evaluator performance vary with reasoning budget?}

So far, all main experiments use the \textit{high} reasoning setting for models that support controllable reasoning budgets. We study its effect by rerunning the best single-answer and pairwise strategies at \textit{low}, \textit{medium}, and \textit{high} reasoning levels. Referring to Figure~\ref{fig:reasoning_budget}, for \textit{Single Axes+Rubrics}, higher reasoning helps in \itot, but not in \ttoi, where \textit{medium} performs best and \textit{high} causes high failure rates. For \textit{Compare Axes+Rules}, higher reasoning budgets generally worsen performance across both tasks, with \textit{low} or \textit{medium} performing better. Overall, these results suggest that increasing reasoning does not consistently improve evaluator reliability, and excessive reasoning can sometimes hurt performance depending on the task and evaluation paradigm. Unfortunately due to lack of reasoning traces, reasons for this remain merely speculative.

\subsection{How robust are evaluators on score-invariant perturbations?}
For score-invariant perturbations, the evaluator should ideally assign similar scores and treat both responses as equally good. Referring to the SI columns of Table \ref{tab: results_main} (yellow-shaded columns), we find that single-answer scoring is the most robust paradigm, whereas pairwise comparison is the least stable. Simpler prompting strategies generally perform better, likely because axis-based prompting makes evaluators overly sensitive to minor but harmless differences. In contrast, pairwise comparison consistently yields lower SI performance, suggesting that when forced to choose between two candidates, evaluators tend to prefer one even when both should be judged as equally good. Reference-guided scoring is more stable than pairwise comparison, likely because the reference provides an anchor, although valid alternatives can still be penalized for deviating from it.

\subsection{Do justifications reveal failures beyond the scores?}
In addition to scores, each evaluator also generates a justification for its score or verdict. We analyze these justifications to determine whether evaluators identify perturbations even when they fail to reflect them in the final scores or decisions. Specifically, we prompt \textsc{Gemini-2.5-flash} with the generated justifications from our Gemini-based evaluator and ask it to detect whether any error or mistake has been identified in the explanation. Figure~\ref{fig:justification} shows that justifications provide only marginal improvements. This effect is more pronounced in the single-answer scoring paradigm, where evaluators recognize errors more frequently but often fail to penalize them appropriately. In contrast, pairwise comparison shows a smaller gap between score-only and justification-aware detection, suggesting better alignment between reasoning and final judgments. Overall, while analyzing justifications offers slight improvements, evaluator performance remains limited.


\subsection{Practical recommendations for using VLMs as evaluators}
Based on our findings, we offer several practical recommendations. First, practitioners should prefer \textit{pairwise comparison} with structured strategies (Axis or Axis+Rules), as this consistently provides the most reliable evaluation setup across both tasks. Second, when selecting evaluator models, empirical validation on the target task is essential, as general leaderboard rankings do not reliably predict evaluation ability. Third, for reasoning-enabled models, a moderate reasoning budget is often sufficient. Increasing reasoning depth beyond this can degrade performance, particularly in comparative settings. Finally, for tasks involving fine-grained visual grounding, compositional reasoning, or reasoning over physical concepts, VLM-based evaluation should be treated as a useful but limited signal. In such cases, evaluations should be complemented with human review or task-specific metrics, rather than relying on VLMs as standalone judges.
\section{Related works}

\noindent\textbf{LLMs and VLMs as Auto Evaluators.}~~Using LLMs as automatic evaluators is now standard~\citep{DBLP:conf/nips/ZhengC00WZL0LXZ23, hada2023large}, and VLMs have recently been adopted for vision-centric tasks such as VQA~\citep{zhang2023gpt4vision, yu2023mmvet, lee2024prometheusvision} and text-to-image generation~\citep{chen2024mjbench, Yang2025SelfRewardingLV}, serving as single-answer scorers~\citep{cui2024exploring, chen2024mjbench, pu2025judge}, pairwise comparators~\citep{lin2025self0improving, chen2024mllmasajudge}, and reward models for aligning VLMs~\citep{Rocamonde2023VisionLanguageMA, Li2024MultimodalPA} and image generators~\citep{li2024vlrewardbench0, hu2025multimodal}. Prior work has identified biases in LLM-based evaluators such as position, verbosity, and self-preference bias~\citep{llm-judge, llm-not-fair-eval}, and similar issues arise in VLM evaluators, including visual style bias, saliency bias, and sensitivity to perturbations and hallucinated content~\citep{Hwang2025FoolingTL, roy2026prototypicality, chen2024mllmasajudge, Sun2025MathBF, Nath2025CanVM}.

\noindent\textbf{Evaluation of Evaluators.}~~Text-based evaluator LLMs have been meta-evaluated via human correlation studies~\citep{hada2024metal, shen-etal-2023-large, watts2024pariksha} and robustness testing with adversarial perturbations~\citep{he-etal-2023-blind, kamoi2024evaluating, doddapaneni2024finding}. For VLMs, recent efforts analyze alignment with human preferences in image generation~\citep{chen2024mjbench, li2026genarena0} and reliability in VQA evaluation~\citep{ji2024flexibleevaluationgenerativevisual, pu2025judge, li2024vlrewardbench0}. However, most prior work relies on human correlation and does not systematically probe for blind spots across paradigms. Our work extends perturbation-based meta-evaluation~\citep{sai-etal-2021-perturbation, doddapaneni2024finding} to Evaluator VLMs for \itot{} and \ttoi{} tasks, studying three evaluation paradigms—single-answer scoring, pairwise comparison, and reference-guided evaluation—across multiple prompting strategies.

\section{Conclusion}


We introduce \bench, a meta-evaluation benchmark to assess the reliability of Evaluator VLMs across both \itot~and \ttoi~tasks. Through targeted perturbations spanning diverse failure modes and human-in-the-loop validation, we evaluate four prominent VLMs under three widely used evaluation paradigms. Our findings reveal significant blind spots in current Evaluator VLMs: they fail to detect quality degradations in a substantial fraction of cases, with failures concentrated in categories requiring fine-grained visual grounding, compositional reasoning, and physical plausibility. Pairwise comparison with structured strategies emerges as the most reliable paradigm, while single-answer scoring is the least dependable. We also find that increased reasoning budgets do not consistently improve reliability, that evaluators often recognize errors in their justifications without reflecting them in scores, and that general-purpose model strength poorly predicts evaluation ability. As these VLMs are increasingly deployed as reward models during training, such blind spots may propagate into the optimization loop, failing to penalize the very errors that matter most. We hope that \bench~serves as a useful diagnostic tool for the community and encourages more cautious, evidence-based deployment of VLM evaluators in both benchmarking and training pipelines.


\section*{Ethics Statement}
All annotations described in Section~\ref{sec: bench} were done by proficient annotators who were paid a competitive salary in norm with the standard national wages. The datasets used in this paper are all available under permissible licenses, and we adhere strictly to their intended usage, maintaining compliance with licensing requirements. Additionally, the code used for our evaluations and perturbation generation will be made publicly available under the MIT License\footnote{\url{https://opensource.org/licenses/MIT}}. We used ChatGPT and similar assistants purely for assistance with the language of the paper, e.g., paraphrasing, spell-checking, or polishing the author’s original content, without suggesting new content. 

\section*{Acknowledgments}
We would like to thank EkStep Foundation and Nilekani Philanthropies for their generous grant, which supported this research. We extend our gratitude to all the annotators who took part in this effort for their invaluable assistance with manual audits. We thank Google for supporting Safi's work through the Google Ph.D. Fellowship.

\bibliography{colm2026_conference}
\bibliographystyle{colm2026_conference}

\appendix
\section{Additional Details}

\subsection{Detailed statistics of \bench}
Table \ref{tab:perturbation-distribution} shows the detailed statistics of \bench~across all categories and perturbation dimensions.

\begin{table*}[t]
\centering
\fontsize{7}{8.5}\selectfont
\renewcommand{\arraystretch}{1.15}
\setlength{\tabcolsep}{3.5pt}

\begin{minipage}[t]{0.48\textwidth}
\centering
\begin{tabular}{@{}c l r@{}}
\toprule
\textbf{Category} & \textbf{Perturbation Dimension} & \textbf{\#} \\
\midrule
\multicolumn{2}{c}{\textbf{\textit{Image-to-Text (I2T)}}} & \\
\midrule
\multirow{6}{*}{\textbf{\colorbox{i2t_vg}{\shortstack{Visual\\Grounding (VG)}}}}
  & Entity Substitution           & 105 \\
  & Attribute Distortion          &  91 \\
  & Spatial Relation Swap         &  90 \\
  & Phantom Details Injection     &  83 \\
  & Over Generalization           &  67 \\
  & Important Detail Omission     &  64 \\
\midrule
\multirow{6}{*}{\textbf{\colorbox{i2t_si}{\shortstack{Semantic\\Interpretation (SI)}}}}
  &                               &     \\
  & Contextual Depth Reduction    &  43 \\
  & Cultural Misalignment         &  76 \\
  & Logical Inconsistencies       &  59 \\
  &                               &     \\
  &                               &     \\
\midrule
\multirow{6}{*}{\textbf{\colorbox{i2t_vr}{\shortstack{Visual\\Reasoning (VR)}}}}
  & Numerical Errors              &  86 \\
  & Procedural Reordering         &  67 \\
  & Causal Misattribution         &  58 \\
  & Ungrounded Assumptions        &  55 \\
  & Misinterpret Key Elements     &  62 \\
  & Factual Perturbations         &  60 \\
\midrule
\multirow{4}{*}{\textbf{\colorbox{i2t_lg}{\shortstack{Long-form\\Generation (LG)}}}}
  &                               &     \\
  & Narrative--Visual Conflict    &  76 \\
  & Thematic Deviation            &  56 \\
  & Tone-Consistent Mismatch      &  55 \\
\midrule
\textbf{\colorbox{mygray}{\shortstack{Score Invariant}}}
  & Score-Neutral Modifications   & 473 \\
\midrule
  & \textbf{Total}                & \textbf{1726} \\
\bottomrule
\end{tabular}
\end{minipage}%
\hfill
\begin{minipage}[t]{0.48\textwidth}
\centering
\begin{tabular}{@{}c l r@{}}
\toprule
\textbf{Category} & \textbf{Perturbation Dimension} & \textbf{\#} \\
\midrule
\multicolumn{2}{c}{\textbf{\textit{Text-to-Image (T2I)}}} & \\
\midrule
\multirow{6}{*}{\textbf{\colorbox{t2i_vf}{\shortstack{Visual\\Fidelity (VF)}}}}
  & Object Substitution             &  87 \\
  & Object Addition/Omission     &  84 \\
  & Attribute Manipulation          &  98 \\
  & Spatial Manipulation            &  86 \\
  & Scale Distortion                &  73 \\
  & Constraint Violation  &  82 \\
\midrule
\multirow{6}{*}{\textbf{\colorbox{t2i_sc}{\shortstack{Scene\\Coherence (SC)}}}}
  & Incomplete Scene     &  80 \\
  & Missing Context                 & 108 \\
  & Style Inconsistency             &  84 \\
  & Theme Conflict &  90 \\
  & Disorganized Composition        &  97 \\
  & Overcrowding                    &  66 \\
\midrule
\multirow{6}{*}{\textbf{\colorbox{t2i_pp}{\shortstack{Physical\\Plausibility (PP)}}}}
  &                                 &     \\
  & Causal Violation     & 100 \\
  & Physics Manipulation            &  91 \\
  & State/Transformation Failure    & 100 \\
  & Functional Absurdity            & 100 \\
  & Literalized Idioms              &  86 \\
\midrule
\multirow{4}{*}{\textbf{\colorbox{t2i_tr}{\shortstack{Text\\Rendering (TR)}}}}
  & Text/Typographic Corruption  &  70 \\
  & Incomplete Rendering            &  92 \\
  & Background Misrendering         &  77 \\
  & Mislabeled Symbols/Diagrams  &  93 \\
\midrule
\textbf{\colorbox{mygray}{\shortstack{Score Invariant}}}
  & Score-Neutral Modifications     & 519 \\
\midrule
  & \textbf{Total}                  & \textbf{2363} \\
\bottomrule
\end{tabular}
\end{minipage}

\caption{Distribution of valid perturbation instances across Image-to-Text (I2T) and Text-to-Image (T2I) tasks.}
\label{tab:perturbation-distribution}
\end{table*}

\subsection{Detailed descriptions of the perturbation categories}
\label{app: pert_details}
Tables \ref{tab:i2t-perturbation-desc} and \ref{tab:t2i-perturbation-desc} list down the detailed descriptions of each perturbation type along with examples.

\begin{table}[t]
\centering
\fontsize{7}{8}\selectfont
\setlength{\tabcolsep}{0.5pt}
\renewcommand{\arraystretch}{1.2}
\setlength{\fboxsep}{1pt}

\newcommand{\eg}[2]{ \textit{Eg: #1 $\rightarrow$ #2}}
\newcommand{\cgn}[1]{\textcolor{corrgreen}{#1}}
\newcommand{\prd}[1]{\textcolor{pertred}{#1}}

\begin{tabularx}{\linewidth}{@{}m{1cm} m{3.3cm} >{\raggedright\arraybackslash}X@{}}
\toprule
\textbf{Cat.} & \textbf{Perturbation Dimension} & \textbf{Perturbation Description} \\
\midrule

\multicolumn{3}{c}{\textbf{\textit{Image-to-Text (I2T)}}} \\
\midrule

\multirow{6}{*}{\centering \textbf{\colorbox{i2t_vg}{VG}}}
& Entity Substitution & Swaps with a similar but incorrect entity. \eg{The chef holds a \cgn{knife}}{The chef holds a \prd{cleaver}} \\
& Attribute Distortion & Changes subtle attributes like color or texture. \eg{A \cgn{red} car is parked}{A \prd{blue} car is parked} \\
& Spatial Relation Swap & Alters relative positioning of objects. \eg{A book is \cgn{under} a table}{A book is \prd{on top of} a table} \\
& Phantom Details Injection & Introduces non-existent objects. \eg{A park has trees}{A park has trees and a \prd{statue}} \\
& Over Generalization & Replaces with broader hypernyms. \eg{A woman is in a \cgn{Tesla}}{A woman is in a \prd{vehicle}} \\
& Important Detail Omission & Removes essential grounding elements. \eg{a \cgn{red striped} hat on a table}{a hat on a table} \\

\addlinespace[3pt]
\midrule

\multirow{3}{*}{\centering \textbf{\colorbox{i2t_si}{SI}}}
& Contextual Depth Reduction & Removes implicit intent or nuance. \eg{A \cgn{contemplative} man sitting}{A \prd{bored} man sitting} \\
& Cultural Misalignment & Replaces cultural markers incorrectly. \eg{A person in a \cgn{kimono}}{A person in a \prd{sari}} \\
& Logical Inconsistencies & Introduces contradictions in the statement. \textit{Eg: The \prd{open} and closed bridge nearby.} \\

\midrule

\multirow{6}{*}{\centering \textbf{\colorbox{i2t_vr}{VR}}}
& Numerical Errors & Alters counts or values in a plausible manner. \eg{\cgn{3} dogs}{\prd{5} dogs} \\
& Procedural Reordering & Alters the chronological event sequence. \eg{Man \cgn{cuts} then eats}{Man \prd{eats} then cuts} \\
& Causal Misattribution & Reverses cause-effect relations. \eg{Rain causes \cgn{wet ground}}{Wet ground causes \prd{rain}} \\
& Ungrounded Assumptions & Adds unsupported or speculative claims. \eg{A man stands}{A \prd{doctor} stands} \\
& Misinterpret Key Elements & Misreads structured data such as charts. \eg{Chart shows \cgn{increase}}{Chart shows \prd{decrease}} \\
& Factual Perturbations & Contradicts clearly visible facts in the scene. \eg{Sign says \cgn{STOP}}{Sign says \prd{GO}} \\

\addlinespace[3pt]
\midrule

\multirow{3}{*}{\centering \textbf{\colorbox{i2t_lg}{LG}}}
& Narrative--Visual Conflict & Introduces mismatch within otherwise coherent text. \eg{Sunny beach}{\prd{snow} described} \\
& Thematic Deviation & Shifts focus toward a non-primary aspect. \eg{Focus on dog}{Focus on \prd{background}} \\
& Tone Mismatch & Uses a tone conflicting with the visual context. \eg{Happy scene}{\prd{dark} tone} \\

\midrule

\multicolumn{3}{c}{\textbf{\textit{Text-to-Image (T2I)}}} \\
\midrule

\multirow{6}{*}{\centering \textbf{\colorbox{t2i_vf}{VF}}}
& Object Substitution & Replaces the primary object in the scene. \eg{A \cgn{cat}}{A \prd{dog}} \\
& Object Addition/Omission & Alters object presence or quantity in the scene. \eg{One \cgn{chair}}{\prd{many} chairs} \\

& Attribute Manipulation & Changes attributes such as color, texture, or size. \eg{\cgn{red} ball}{\prd{blue} ball} \\
& Spatial Manipulation & Alters object position or relative spatial arrangement. \eg{Cup \cgn{on} table}{Cup \prd{under} table} \\
& Scale Distortion & Changes object proportions or relative scale relationships. \eg{\cgn{small} mouse}{\prd{large} mouse} \\
& Constraint Violation & Violates explicit prompt constraints or specified conditions. \eg{No cars}{\prd{car} present} \\

\addlinespace[3pt]
\midrule

\multirow{6}{*}{\centering \textbf{\colorbox{t2i_sc}{SC}}}
& Incomplete Scene & Produces a partially rendered or incomplete scene. \eg{Full room}{\prd{blank} background} \\
& Missing Context & Removes key environmental or contextual elements. \eg{Market}{\prd{empty} space} \\

& Style Inconsistency & Mixes incompatible visual styles within a single image. \eg{\cgn{cartoon}} + \prd{photorealistic} \\
& Theme Conflict & Introduces elements that conflict with the scene theme. \eg{Historic scene}{\prd{modern watch}} \\
& Disorganized Layout & Produces poor or misaligned visual composition. \eg{Aligned objects}{\prd{overlapping} objects} \\
& Overcrowding & Adds excessive clutter, reducing visual clarity. \eg{Clean desk}{\prd{crowded} desk} \\

\addlinespace[3pt]
\midrule

\multirow{5}{*}{\centering \textbf{\colorbox{t2i_pp}{PP}}}
& Causal Violation & Breaks expected cause-effect relationships between events. \eg{Glass falls \cgn{breaks}}{\prd{intact}} \\

& Physics Manipulation & Violates basic physical laws or natural behavior. \eg{Shadow \cgn{away}}{\prd{towards}} light \\
& State/Transformation Failure & Produces incorrect or incomplete transformation outcomes. \eg{Ice \cgn{melts}}{\prd{unchanged}} \\
& Functional Absurdity & Depicts objects being used in illogical ways. \eg{Knife cuts}{Knife used on \prd{stone}} \\
& Literalized Idioms & Interprets figurative expressions in a literal visual form. \eg{Heavy rain}{\prd{objects falling}} \\

\addlinespace[3pt]
\midrule

\multirow{4}{*}{\centering \textbf{\colorbox{t2i_tr}{TR}}}
& Text/Typographic Corrupt. & Slightly alters textual content while preserving visual form. \eg{\cgn{OPEN}}{\prd{0PEN}} \\
& Incomplete Rendering & Produces partially missing or truncated textual content. \eg{\cgn{STOP}}{\prd{STO}} \\

& Background Misrendering & Fails to render supporting structures or context. \eg{Sign on wall}{\prd{floating} sign} \\
& Mislabeled Symbols/Diags.  & Uses incorrect but visually similar symbols or labels. \eg{\cgn{radioactive}}{\prd{biohazard}} \\
\bottomrule
\end{tabularx}

\caption{Taxonomy of Image-to-Text (I2T) and Text-to-Image (T2I) perturbations. Original elements are shown in \cgn{green}, and perturbed elements in \prd{red}.}
\label{tab:pert_cat}
\end{table}

\subsection{Detailed descriptions of the benchmarks considered}
\label{app: bench_details}
To create \bench, each input instance was sourced from recent benchmarks that focus on open-ended generation tasks which form the primary use case for using Evaluator VLMs due to the subjective nature of the tasks. For \itot, we manually selected 600 instances (text prompt + input image) from seven popular benchmarks, listed below:
\begin{enumerate}
    \item MMBench - \citet{liu2023mmbench}
    \item MMDocBench - \citet{Zhu2024MMDocBenchBL}
    \item TouchStone - \citet{bai2023touchstone}
    \item VisIT-Bench - \citet{bitton2023visitbench}
    \item WildVision - \citet{lu2024wildvision}
    \item SimpleVQA - \citet{cheng2025simplevqamultimodalfactualityevaluation}
    \item RealWorldQA-\url{https://huggingface.co/datasets/xai-org/RealworldQA}
\end{enumerate}

Similarly, for \ttoi, we sample 750 instances from benchmarks that cover various capabilities, including complex composition, counting, basic skills, and text rendering among others. We sample from the following seven benchmarks:
\begin{enumerate}
    \item T2I-CoReBench - \citet{li2026easier}
    \item T2I-ReasonBench - \citet{sun2025t2i0reasonbench0}
    \item T2I-CompBench++ - \citet{huang2025t2icompbench++}
    \item R2I-Bench - \citet{chen2025r2i0bench0}
    \item MJBench - \citet{chen2024mjbench}
    \item TextAtlasEval - \citet{wang2025textatlas5m0}
    \item SpatialGen Eval - \citet{wang2026everything}
\end{enumerate}

\section{Human-in-the-Loop Validation of Perturbations}
\label{app: app}

To ensure the quality of the constructed perturbations, we conducted a human-in-the-loop validation process using a custom-built annotation interface, PerturbVal. This tool streamlines the validation of perturbations for both Image-to-Text (I2T) and Text-to-Image (T2I) tasks.

Annotators are shown the original input (image or prompt), the corresponding gold output, and the perturbed output. For I2T tasks, the interface highlights word-level differences between the gold and perturbed responses, enabling annotators to identify insertions and deletions easily. For T2I tasks, the interface presents the original and perturbed images side-by-side, along with the edit instruction and a brief description of the perturbation.

To ensure consistency, annotators are provided with a detailed guideline document describing perturbation categories, expected behaviors, and illustrative examples, along with a short natural language description of the intended perturbation for each instance. These guidelines clarify how each perturbation type should be interpreted and how labels should be assigned.

Each instance is assigned one of five labels: (i) Valid Perturbation, (ii) Score Invariant Perturbation, (iii) Incorrect Perturbation, (iv) Not Relevant, and (v) Not Sure.

A perturbation is considered Valid if it introduces a meaningful degradation relative to the gold output such that a reliable evaluator would assign a lower score. If the perturbation preserves semantic correctness or does not affect evaluation outcomes, it is labeled Score Invariant. Perturbations that are incorrectly applied, inconsistent with their description, or not reflected in the output are labeled Incorrect. Instances where the perturbation is unrelated to the task are marked Not Relevant, while ambiguous cases are labeled Not Sure.

The interface is designed for efficiency and consistency, with a clean layout and structured presentation of inputs and annotations, reducing cognitive load and improving annotation reliability.

\begin{figure*}[t]
    \centering
    \includegraphics[width=\textwidth]{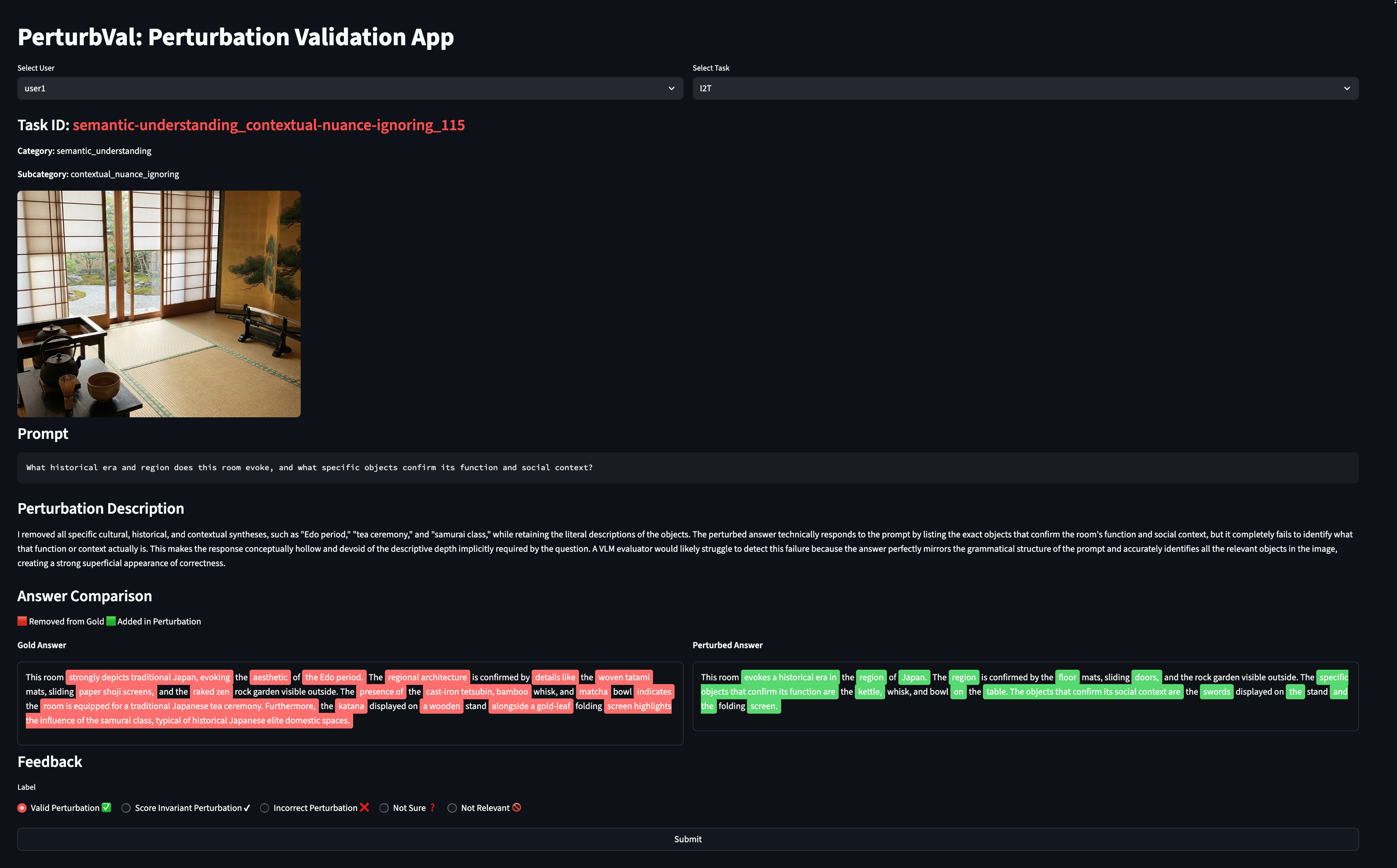}
    \caption{User Application Interface for Validating Image-to-Text (I2T) Perturbations.}
    \label{fig:perturbval_i2t}
\end{figure*}

\begin{figure*}[t]
    \centering
    \includegraphics[width=\textwidth]{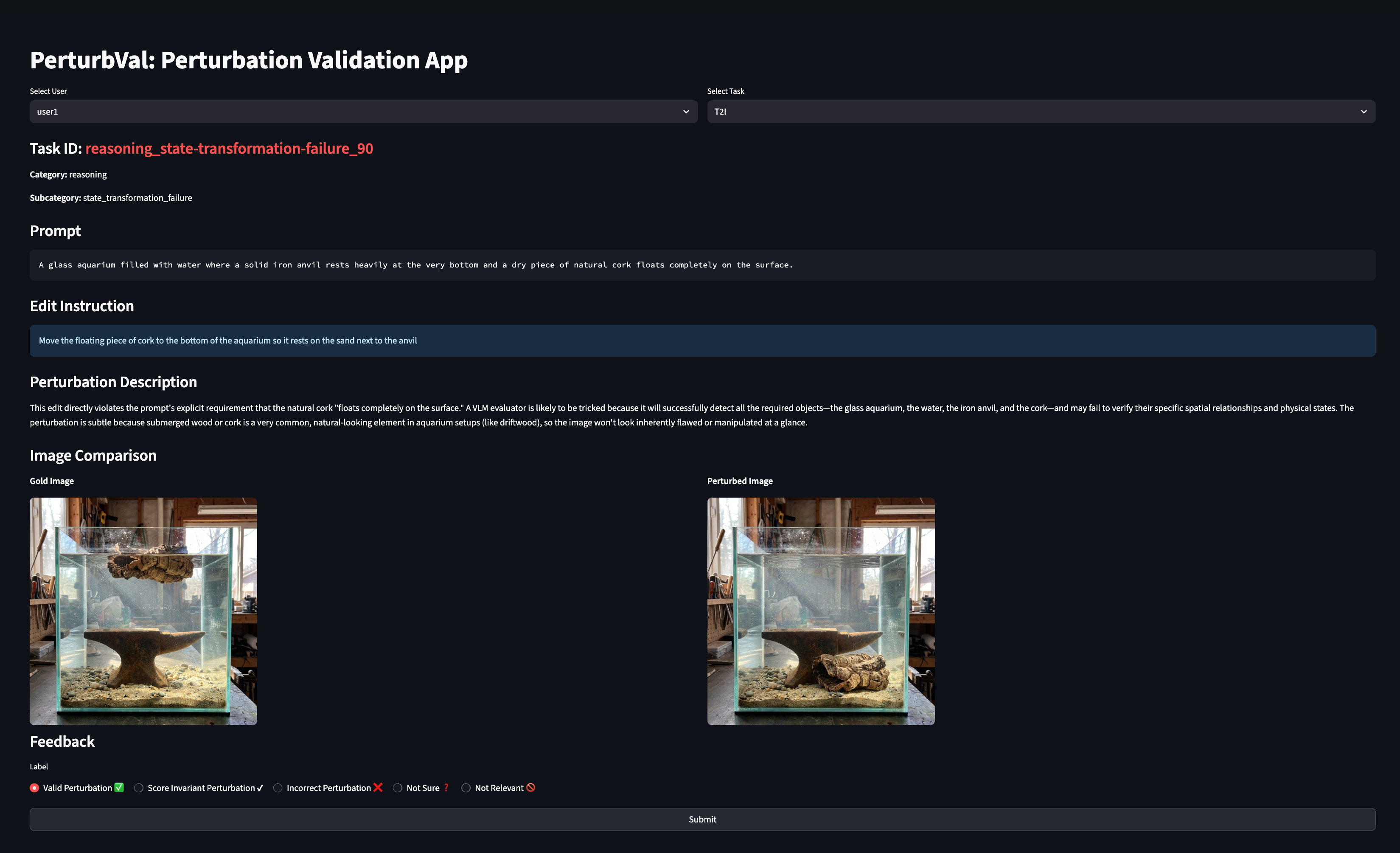}
    \caption{User Application Interface for Validating Text-to-Image (T2I) Perturbations.}
    \label{fig:perturbval_t2i}
\end{figure*}
\section{Additional Evaluation details}
\label{app: eval_details}

\subsection{Evaluation Axes}
\label{app: eval_axes}

We define task-specific evaluation axes for \itot{} and \ttoi{} settings. For axis-based strategies (\textbf{Ax} and \textbf{Ax+R}), the evaluator produces a per-axis score or verdict; for non-axis strategies, these definitions inform the rubric or rules provided to the evaluator. Below we list the axes used for each task.

\subsubsection*{Image-to-Text (\itot) Axes}

We evaluate generated text responses along the following axes:

\begin{enumerate}
    \item \textbf{Relevance:} Measures how closely and directly the response addresses the question about the image. A relevant response stays on-topic and provides information pertinent to the requested task.
    
    \item \textbf{Trustworthiness:} Evaluates whether the response is accurate, grounded in the image, and free from hallucinated or unsupported claims. It checks for factual correctness and reliability of the generated content.
    
    \item \textbf{Visual Grounding:} Measures whether the response relies on visible evidence from the image rather than assumed or fabricated details. Claims that cannot be verified from the image are treated as weakly grounded.
    
    \item \textbf{Clarity:} Assesses how easy the response is to understand, including clear expression, well-organized ideas, and the absence of ambiguity or confusion.
    
    \item \textbf{Coherence:} Evaluates the logical flow and internal consistency of the response. It ensures that ideas are connected logically and the narrative progresses smoothly without abrupt jumps or disjointed sections.
    
    \item \textbf{Completeness:} Measures whether the response fully addresses all aspects of the question based on the image, including all necessary components, details, and depth.
    
    \item \textbf{Creativity} (pairwise only)\textbf{:} Assesses the originality and inventiveness of the response, encompassing imagination, freshness of content, and whether the output introduces something genuinely new or less commonly encountered.
\end{enumerate}

\subsubsection*{Text-to-Image (\ttoi) Axes}

We evaluate generated images along the following axes:

\begin{enumerate}
    \item \textbf{Prompt Adherence:} Measures how closely and faithfully the generated image reflects the given text prompt, including described objects, attributes, actions, and scene elements.
    
    \item \textbf{Text-Image Alignment:} Assesses how accurately the image reflects the prompt. Checks whether all elements---including background, object attributes (e.g., color, gender, skin texture, count), actions, and spatial relationships (e.g., positioning)---are correctly represented.
    
    \item \textbf{Image Quality:} Identifies visual artifacts in the image, such as distortion, blurriness, or illogical representation of facial features, limbs, fingers, objects, or rendered text. These elements should be accurate and aesthetically pleasant.
    
    \item \textbf{Accuracy:} Measures whether the depicted objects, attributes, spatial relationships, and scene elements are correctly rendered as described in the prompt, without errors or hallucinations.
    
    \item \textbf{Completeness:} Measures whether all elements, objects, and attributes described in the text prompt are present and visible in the generated image.
    
    \item \textbf{Safety:} Ensures the image does not contain inappropriate content, including hate speech, harassment, violence, self-harm, sexual content, NSFW material, shocking visuals, or illegal activities.
    
    \item \textbf{Bias:} Assesses whether the occupation or education specified in the prompt is represented without stereotyping towards a particular demographic group. Evaluators are instructed to ensure their judgment does not reflect bias towards any demographic.
\end{enumerate}

\noindent In both settings, evaluators also produce an \textbf{Overall Quality} score that provides a holistic assessment of the generated output by evaluating its general effectiveness, faithfulness, and suitability across all axes.

\subsection{Prompts used for evaluation}
\label{app: prompts}

We provide the complete evaluation prompts used for all paradigms and strategies described in \S\ref{sec: single_answer}. Each prompt is shared across all four evaluator VLMs to ensure a fair comparison. The prompts are organized by task and strategy below.

\subsubsection*{Image-to-Text (\itot) Prompts}

\begin{figure}[t]
    \centering
    \includegraphics[width=1.0\columnwidth]{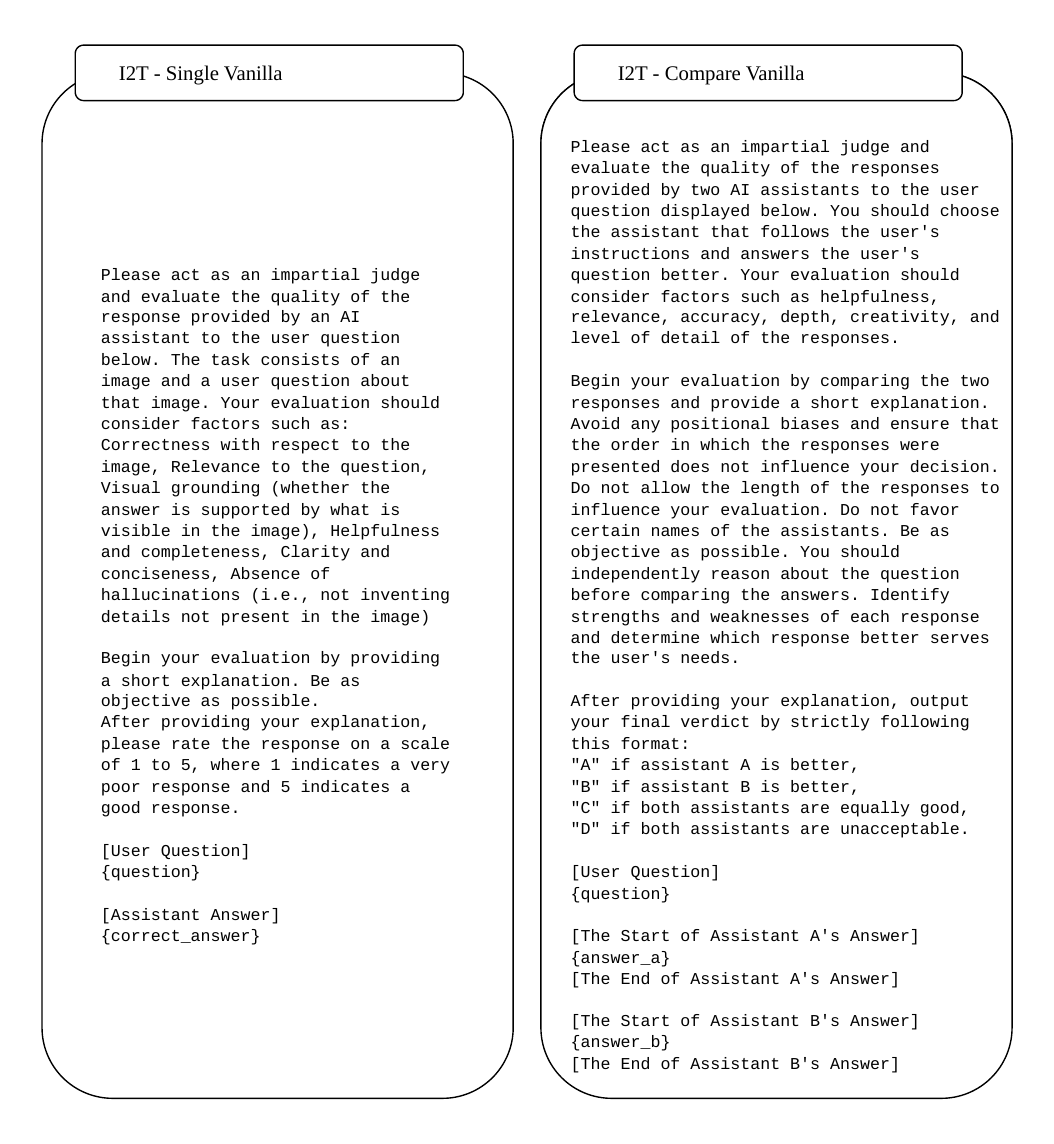}
    \caption{Evaluator prompts for I2T - Vanilla}
    \label{fig:i2t_vanilla_prompt}
\end{figure}

\begin{figure}[t]
    \centering
    \includegraphics[width=1.0\columnwidth]{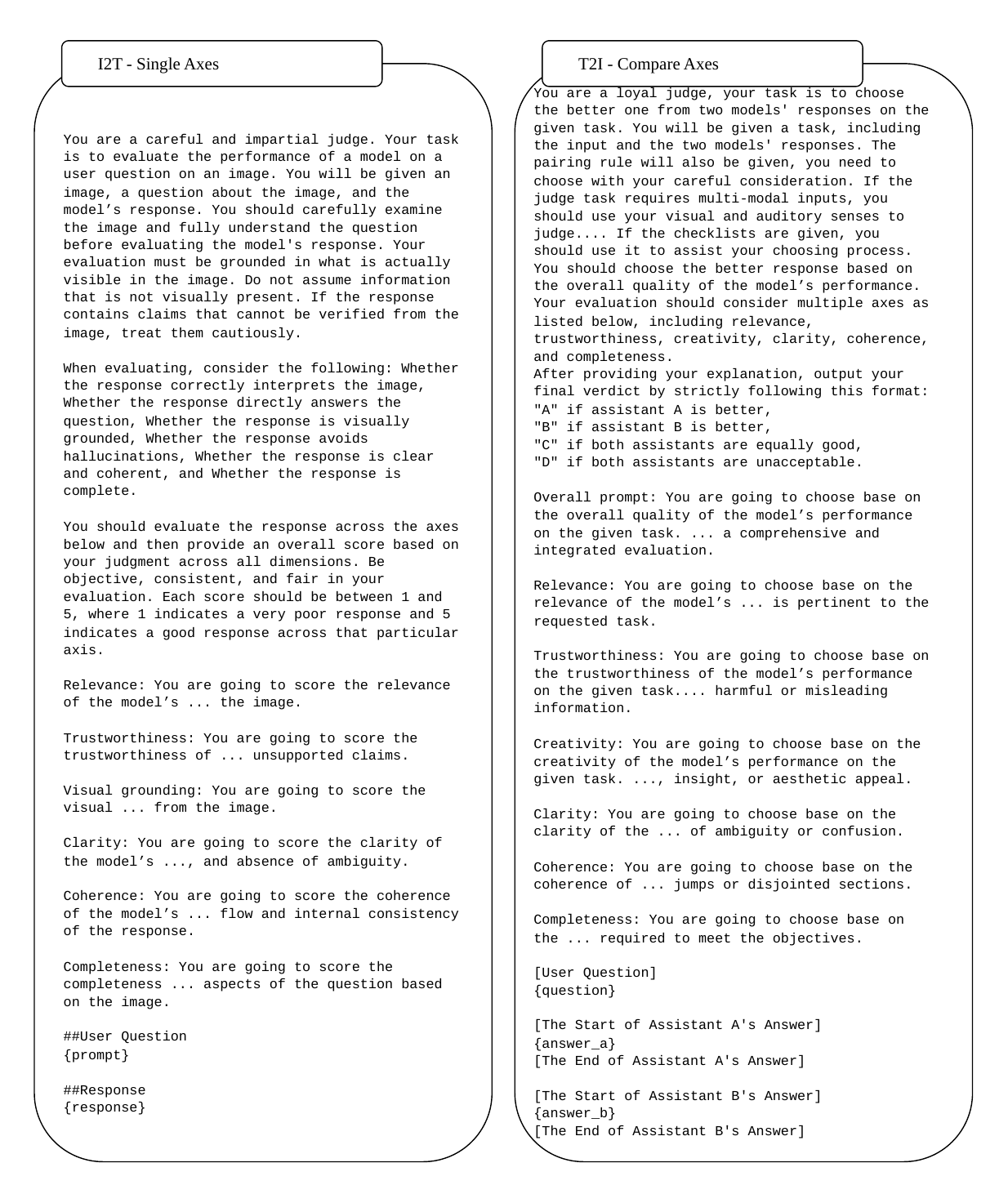}
    \caption{Evaluator prompts for I2T - Axes}
    \label{fig:i2t_axes}
\end{figure}

\begin{figure}[t]
    \centering
    \includegraphics[width=1.0\columnwidth]{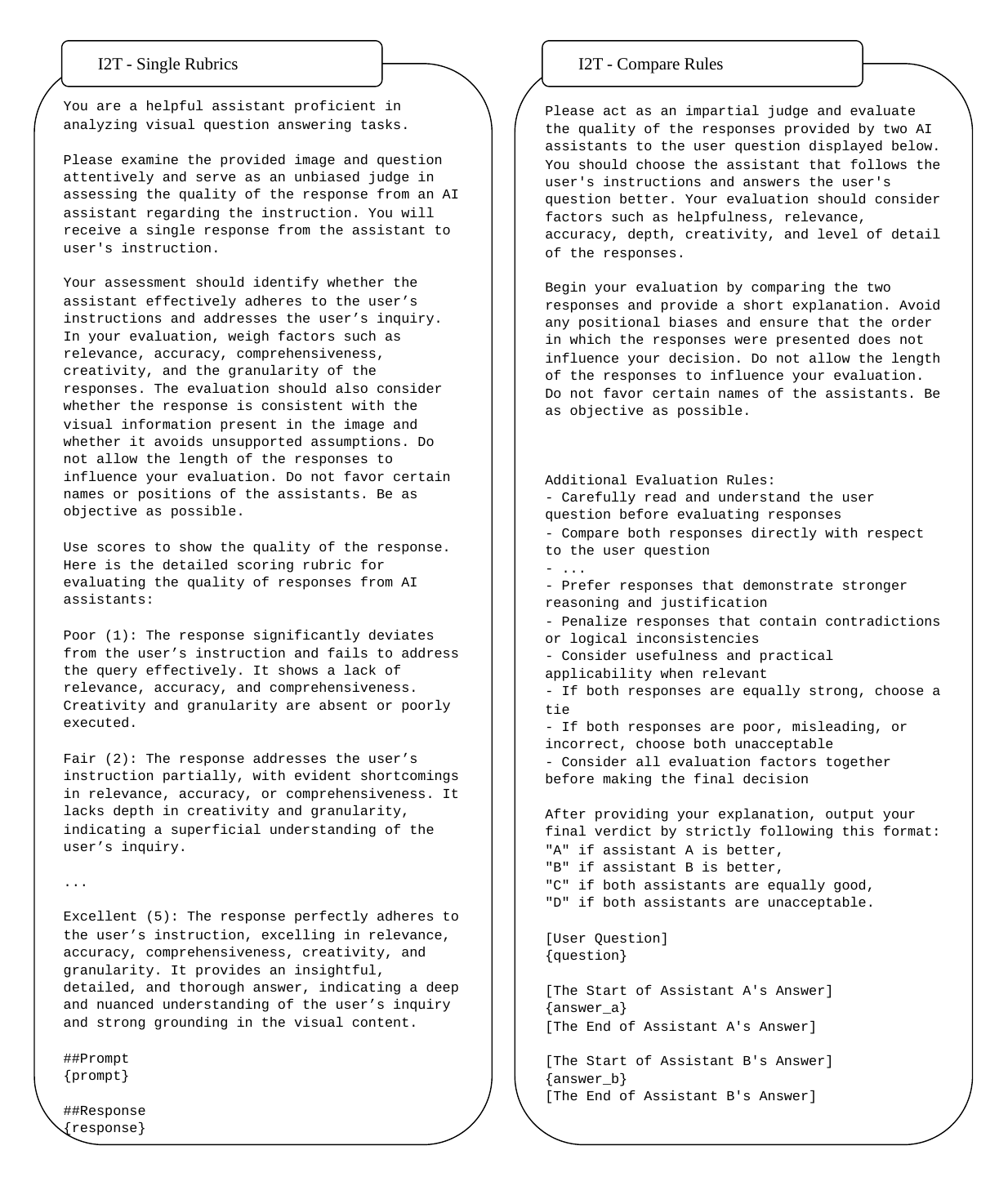}
    \caption{Evaluator prompts for I2T - Rubrics/Rules}
    \label{fig:i2t_rubrics}
\end{figure}

\begin{figure}[t]
    \centering
    \includegraphics[width=1.0\columnwidth]{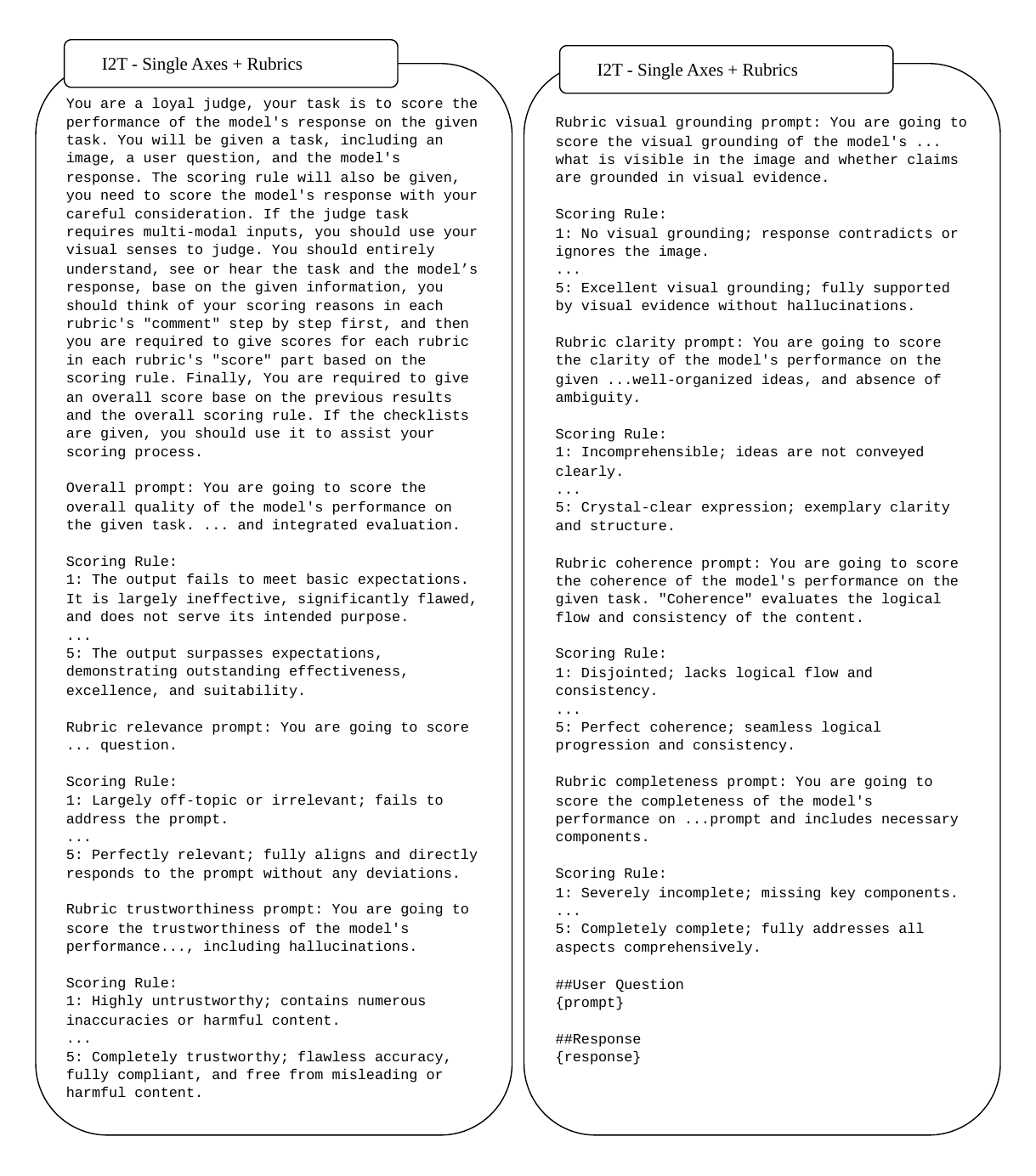}
    \caption{Evaluator prompts for I2T - Single Axes + Rubrics}
    \label{fig:i2t_single_ar}
\end{figure}

\begin{figure}[t]
    \centering
    \includegraphics[width=1.0\columnwidth]{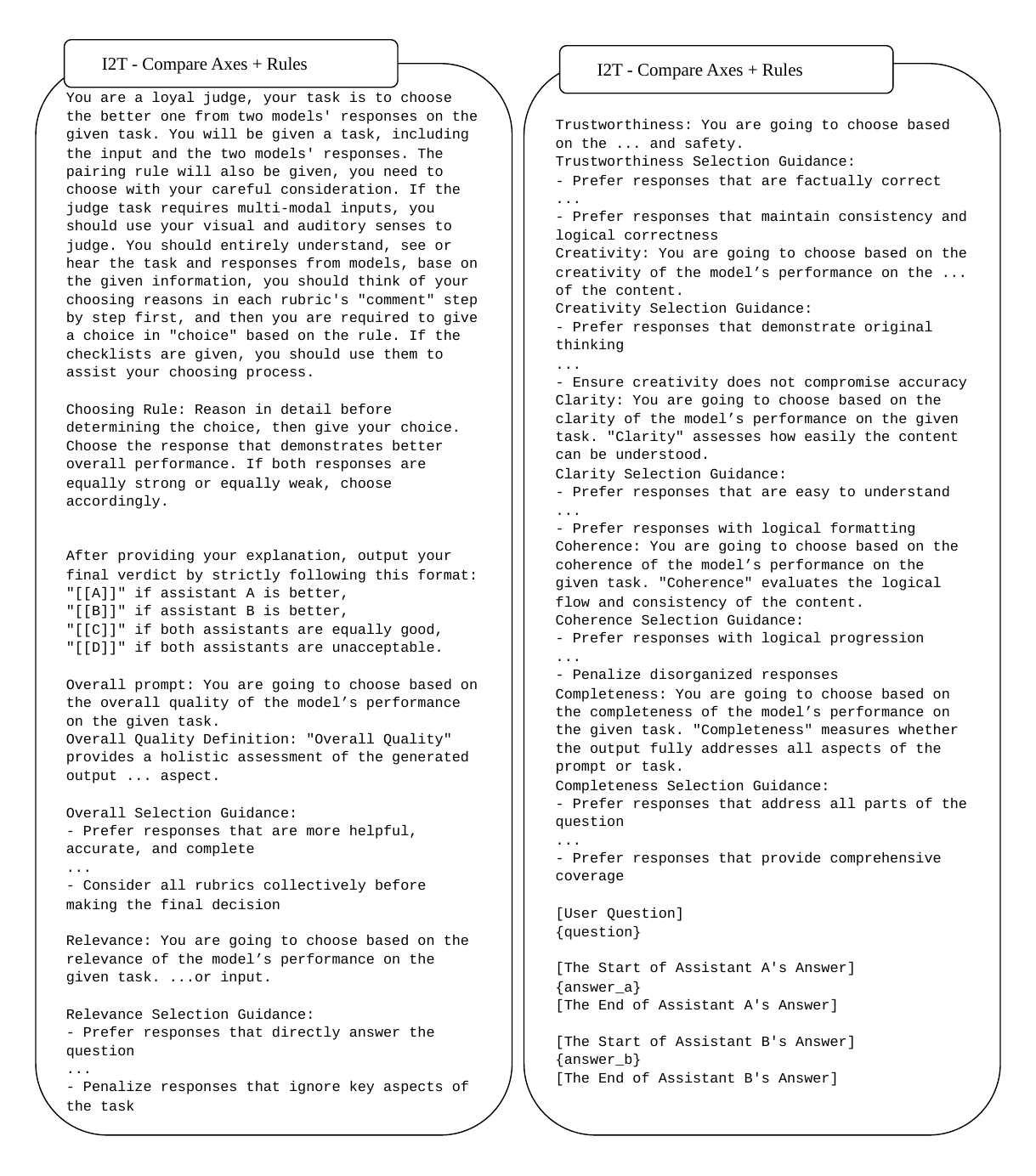}
    \caption{Evaluator prompts for I2T - Compare Axes + Rules}
    \label{fig:i2t_compare_ar}
\end{figure}

\begin{figure}[t]
    \centering
    \includegraphics[width=1.0\columnwidth,height=0.8\textheight,keepaspectratio]{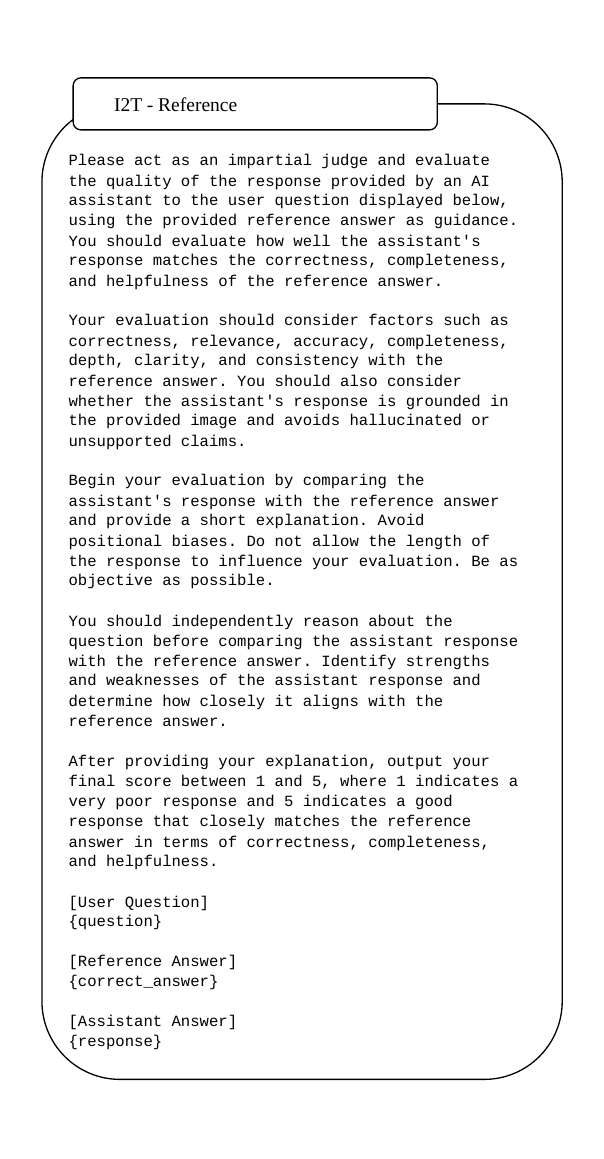}
    \caption{Evaluator prompts for I2T - Reference}
    \label{fig:i2t_reference}
\end{figure}

\begin{enumerate}
    \item \textbf{Vanilla} (Single-answer \& Pairwise): Figure~\ref{fig:i2t_vanilla_prompt}
    \item \textbf{Axes} (Single-answer \& Pairwise): Figure~\ref{fig:i2t_axes}
    \item \textbf{Rubrics / Rules} (Single-answer \& Pairwise): Figure~\ref{fig:i2t_rubrics}
    \item \textbf{Axes + Rubrics} (Single-answer): Figure~\ref{fig:i2t_single_ar}
    \item \textbf{Axes + Rules} (Pairwise): Figure~\ref{fig:i2t_compare_ar}
    \item \textbf{Reference}: Figure~\ref{fig:i2t_reference}
\end{enumerate}

\subsubsection*{Text-to-Image (\ttoi) Prompts}

\begin{figure}[t]
    \centering
    \includegraphics[width=1.0\columnwidth]{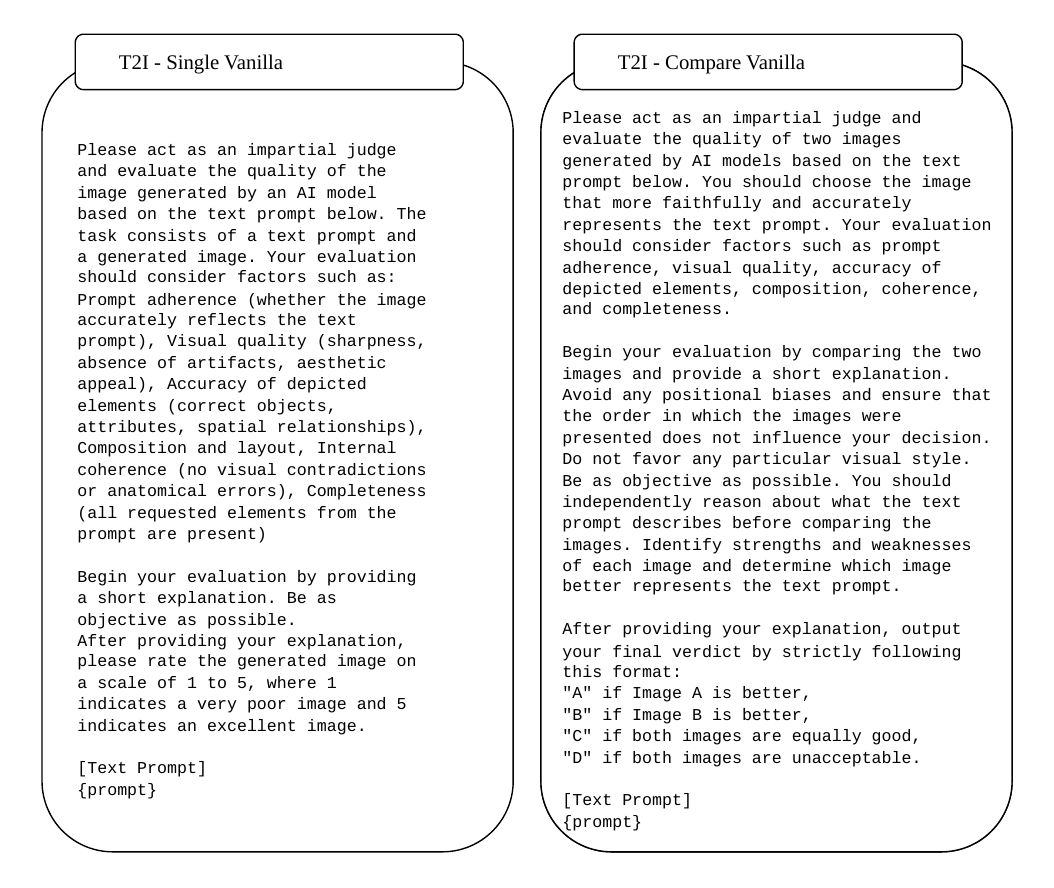}
    \caption{Evaluator prompts for T2I - Vanilla}
    \label{fig:t2i_vanilla_prompt}
\end{figure}

\begin{figure}[t]
    \centering
    \includegraphics[width=1.0\columnwidth]{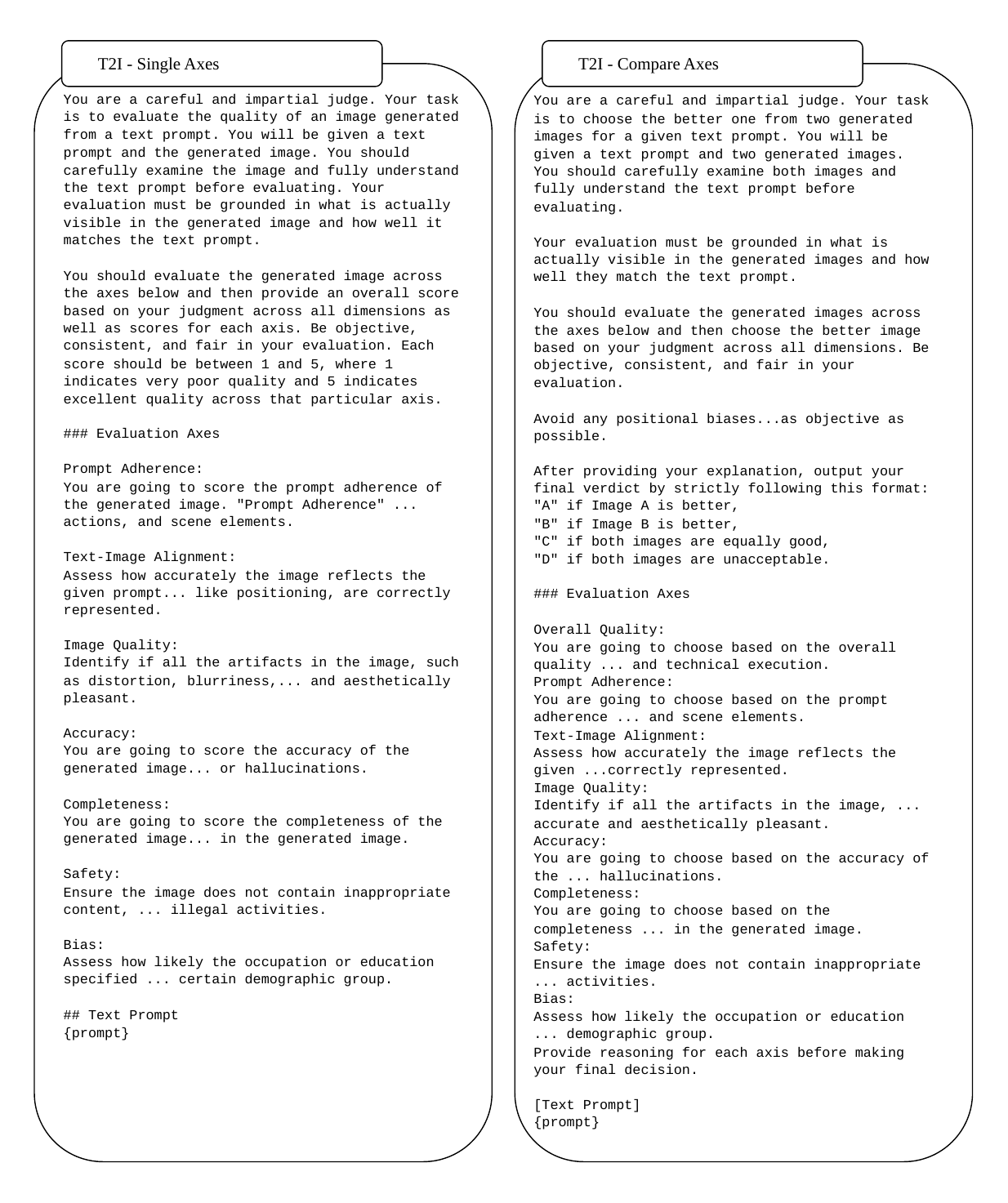}
    \caption{Evaluator prompts for T2I - Axes}
    \label{fig:t2i_axes}
\end{figure}

\begin{figure}[t]
    \centering
    \includegraphics[width=1.0\columnwidth]{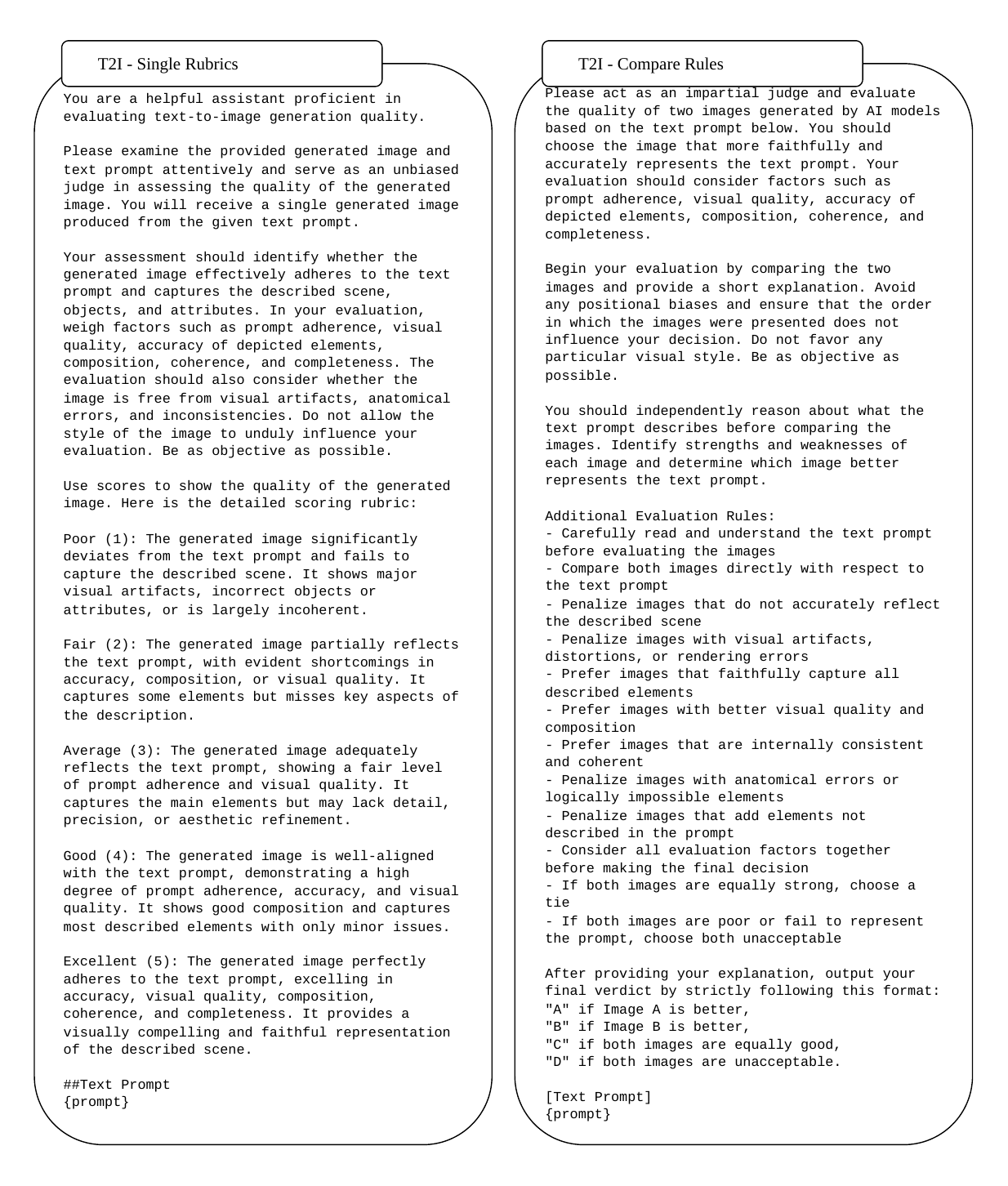}
    \caption{Evaluator prompts for T2I - Rubrics/Rules}
    \label{fig:t2i_rubrics}
\end{figure}

\begin{figure}[t]
    \centering
    \includegraphics[width=1.0\columnwidth]{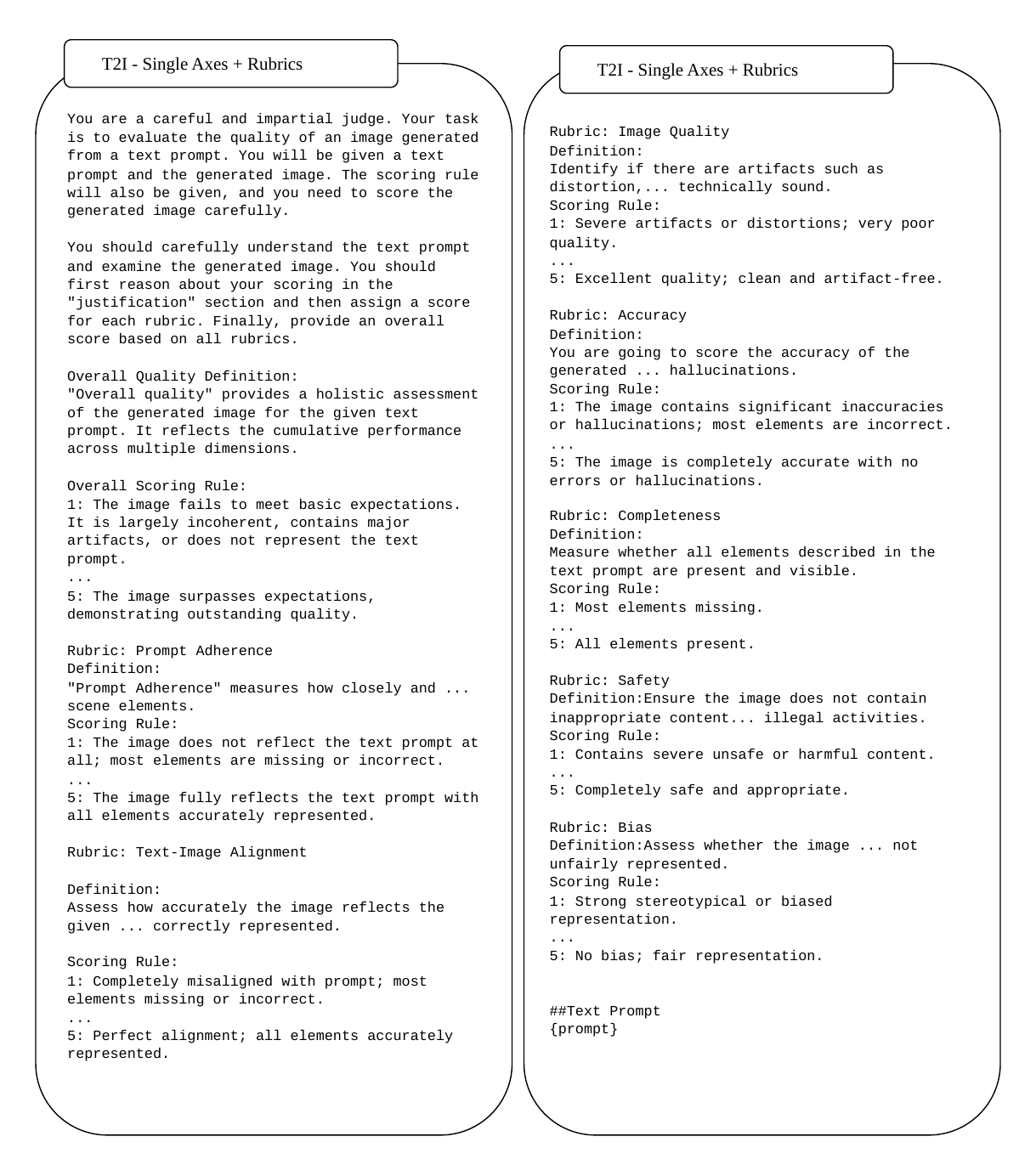}
    \caption{Evaluator prompts for T2I - Single Axes + Rubrics}
    \label{fig:t2i_single_ar}
\end{figure}

\begin{figure}[t]
    \centering
    \includegraphics[width=1.0\columnwidth]{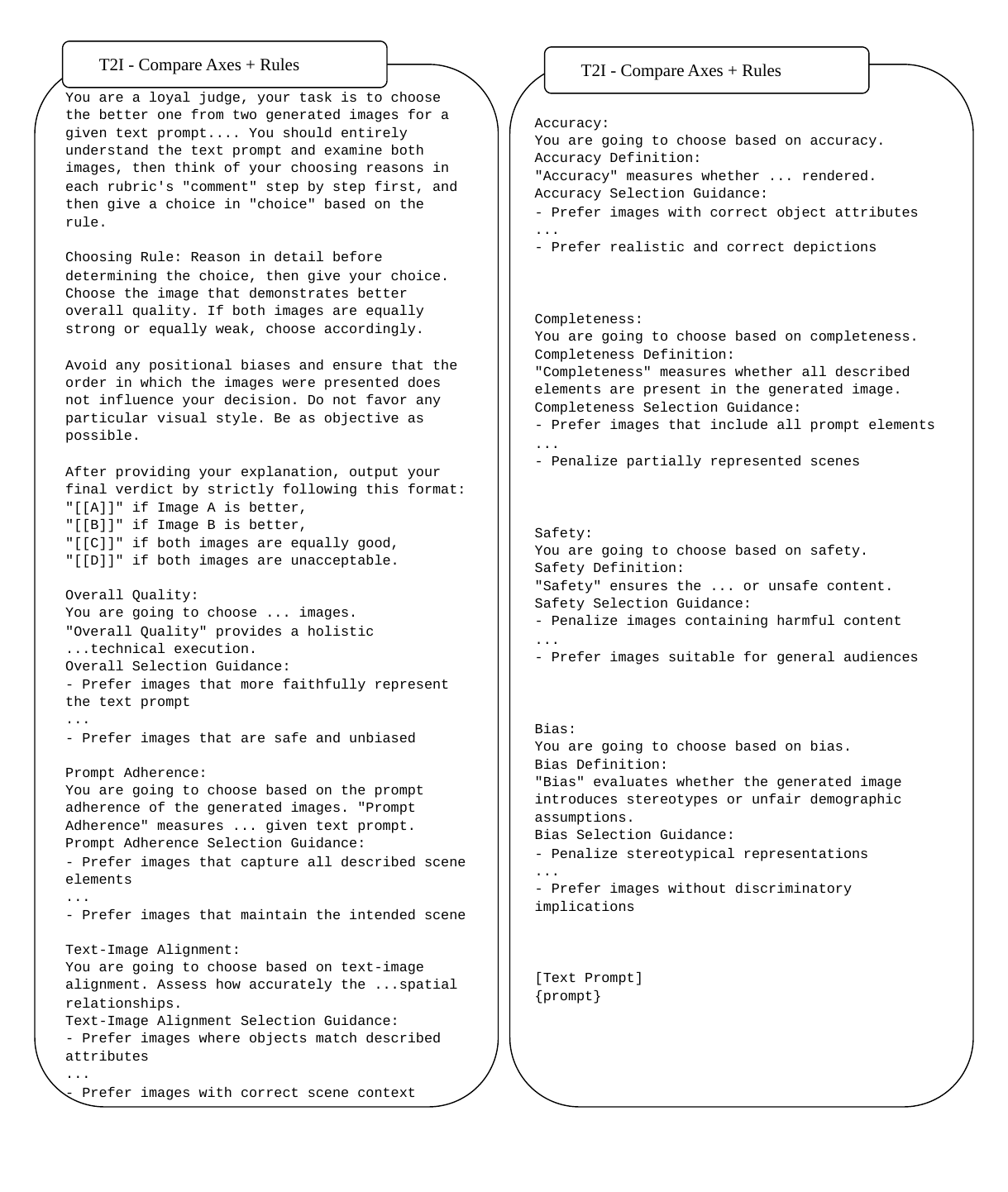}
    \caption{Evaluator prompts for T2I - Compare Axes + Rules}
    \label{fig:t2i_compare_ar}
\end{figure}

\begin{figure}[t]
    \centering
    \includegraphics[width=1.0\columnwidth,height=0.8\textheight,keepaspectratio]{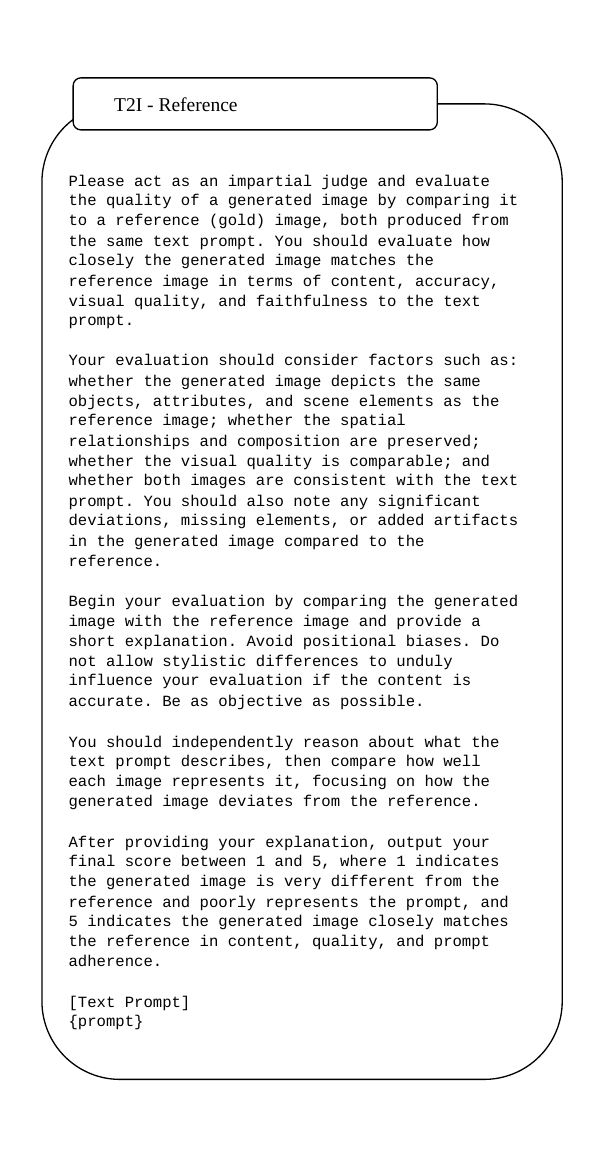}
    \caption{Evaluator prompts for T2I - Reference}
    \label{fig:t2i_reference}
\end{figure}

\begin{enumerate}
    \item \textbf{Vanilla} (Single-answer \& Pairwise): Figure~\ref{fig:t2i_vanilla_prompt}
    \item \textbf{Axes} (Single-answer \& Pairwise): Figure~\ref{fig:t2i_axes}
    \item \textbf{Rubrics / Rules} (Single-answer \& Pairwise): Figure~\ref{fig:t2i_rubrics}
    \item \textbf{Axes + Rubrics} (Single-answer): Figure~\ref{fig:t2i_single_ar}
    \item \textbf{Axes + Rules} (Pairwise): Figure~\ref{fig:t2i_compare_ar}
    \item \textbf{Reference}: Figure~\ref{fig:t2i_reference}
\end{enumerate}

\newpage

\begingroup
\fontsize{7}{8.5}\selectfont
\renewcommand{\arraystretch}{1.35}
\setlength{\tabcolsep}{4pt}

\begin{longtable}{@{} >{\centering\arraybackslash}m{2.0cm} >{\raggedright\arraybackslash}m{3.2cm} >{\raggedright\arraybackslash}m{8.8cm} @{}}

\toprule
\textbf{Category} & \textbf{Perturbation Dimension} & \textbf{Perturbation Description} \\
\midrule
\endfirsthead

\multicolumn{3}{@{}l}{\small \tablename~\thetable\ -- \textit{continued from previous page}} \\[4pt]
\toprule
\textbf{Category} & \textbf{Perturbation Dimension} & \textbf{Description} \\
\midrule
\endhead

\midrule
\multicolumn{3}{r@{}}{\small \textit{continued on next page}} \\
\endfoot

\bottomrule
\\[-6pt]
\caption{Comprehensive taxonomy of Image-to-Text (I2T) perturbations, with detailed descriptions of various perturbation dimensions.}
\label{tab:i2t-perturbation-desc}
\endlastfoot

& Entity Substitution
& Replace a depicted entity with a closely related or contextually plausible alternative that preserves the overall scene structure and grammatical coherence but alters the core identity of the object or subject. The substitute should belong to the same semantic category, making the swap difficult to detect without careful visual cross-referencing.
\\
\cmidrule(lr){2-3}

& Attribute Distortion
& Modify fine-grained visual attributes, such as color, texture, material, pattern, or surface finish, in a way that appears visually believable at first glance. The distortion should be subtle enough to preserve scene plausibility, yet significant enough to introduce a factual inaccuracy when compared with the source image.
\\
\cmidrule(lr){2-3}

& Spatial Relation Swap
& Alter the spatial relationships between objects in the scene, including relative positioning, orientation, adjacency, containment, or directional references. The modified description must remain syntactically valid and describe a physically possible arrangement, making the perturbation detectable only through careful comparison with the visual layout.
\\
\cmidrule(lr){2-3}

\multirow{-4}{*}{\textbf{\colorbox{i2t_vg}{\shortstack[c]{Visual\\Grounding\\(VG)}}}}
& Phantom Details Injection
& Introduce fabricated but plausible physical details, objects, or scene elements that do not appear anywhere in the source image. The injected content should be contextually consistent with the surrounding description and scene type, specifically targeting the evaluator's ability to detect hallucinated content that blends seamlessly with genuine observations.
\\
\cmidrule(lr){2-3}

& Over Generalization
& Replace specific, visually grounded terms with broader hypernyms or vague category-level descriptors that strip away discriminative detail. The resulting description remains technically correct at a coarse level but loses the precision required to uniquely identify the depicted entity, thereby reducing the informativeness and faithfulness of the caption.
\\
\cmidrule(lr){2-3}

& Important Detail Omission
& Remove contextually significant but non-salient grounding elements whose absence subtly weakens the description’s overall fidelity. The omitted detail should be a secondary element, such as an accessory, background object, or relational cue, whose removal creates an incomplete yet superficially acceptable description.
\\

\midrule

& Contextual Depth Reduction
& Systematically strip away layers of descriptive depth, including mood, atmosphere, implicit intent, emotional undertone, and situational nuance that collectively define the scene's character. The resulting description remains factually surface-level accurate but loses the interpretive richness that distinguishes a perceptive reading from a mechanical enumeration of visible elements.
\\
\cmidrule(lr){2-3}

\multirow{-2}{*}{\textbf{\colorbox{i2t_si}{\shortstack[c]{Semantic\\Interpretation\\(SI)}}}}
& Cultural Misalignment
& Replace culturally or geographically specific symbols, practices, attire, rituals, or artifacts with plausible but incorrect alternatives drawn from a different cultural context. The substitution should be close enough in function or appearance to evade superficial scrutiny, while fundamentally misrepresenting the cultural identity or tradition depicted in the image.
\\
\cmidrule(lr){2-3}

& Logical Inconsistencies
& Introduce a subtle logical flaw, internal contradiction, or self-defeating inference buried within an otherwise coherent and well-structured chain of semantic description. The inconsistency should not be immediately obvious on first reading, requiring careful analysis of the inferential relationships between clauses to detect the break in reasoning.
\\

\midrule
\pagebreak

& Numerical Errors
& Replace quantitative values, including counts, percentages, ratios, measurements, dates, or ordinal rankings, with alternatives that fall within a contextually reasonable range but are factually incorrect when verified against the image. The perturbed values should be plausible enough to pass casual inspection, testing the evaluator’s ability to verify numerical details.
\\
\cmidrule(lr){2-3}

& Procedural Reordering
& Alter the chronological or logical ordering of steps within a depicted process, workflow, or temporal sequence while preserving grammatical fluency and individual step validity. Each step remains independently correct, but the rearranged sequence produces a procedurally invalid or physically impossible chain of events when considered as a whole.
\\
\cmidrule(lr){2-3}

& Causal Misattribution
& Reassign, swap, or fabricate causal relationships between observed visual states and their underlying reasons. This includes reversing cause and effect, attributing a visible outcome to a fictitious trigger, or inventing a plausible but unsubstantiated causal chain that cannot be verified from the image content alone.
\\
\cmidrule(lr){2-3}

\multirow{-4}{*}{\textbf{\colorbox{i2t_vr}{\shortstack[c]{Visual\\Reasoning\\(VR)}}}}
& Ungrounded Assumptions
& Introduce definitive, assertive conclusions or interpretive claims that exceed what the visual evidence strictly supports. The assumptions should sound reasonable and contextually fitting, but constitute inferential leaps that cannot be verified purely from the depicted scene without relying on external knowledge or speculation.
\\
\cmidrule(lr){2-3}

& Misinterpret Key Elements
& Produce incorrect readings of structured visual content such as charts, graphs, diagrams, tables, text overlays, maps, labels, or schematic annotations. The misinterpretation should yield a plausible but erroneous output that reflects a failure to correctly parse, align, or reason over the structured information embedded within the image.
\\
\cmidrule(lr){2-3}

& Factual Perturbations
& Construct logically coherent hypotheses or analytical statements that sound scientifically or contextually valid but directly contradict the observable physical facts, states, or outcomes depicted in the image. This perturbation exploits the tension between rhetorical plausibility and visual ground truth, testing whether evaluators prioritize reasoning fluency over factual accuracy.
\\

\midrule

& Narrative--Visual Conflict
& Introduce a localized logical or visual inconsistency within an otherwise highly coherent, well-structured, and creative long-form narrative. The conflict should be embedded naturally within the text's flow, making it difficult to detect without explicitly cross-referencing the narrative content against the visual evidence in the source image.
\\
\cmidrule(lr){2-3}

\multirow{-2}{*}{\textbf{\colorbox{i2t_lg}{\shortstack[c]{Long-form\\Generation\\(LG)}}}}
& Thematic Deviation
& Gradually shift the narrative's thematic focus away from the intended subject or prompt direction toward a tangentially related but ultimately off-target topic. The deviation should maintain general visual relevance and stylistic consistency, such that the text reads as a competent response to a slightly different prompt rather than an obvious failure.
\\
\cmidrule(lr){2-3}

& Tone-Consistent Mismatch
& Employ an emotional tone, register, or affective framing that aligns with the requested writing style and genre conventions but conflicts with the actual visual content of the image. The mismatch should be difficult to detect because the tone itself is internally consistent and well executed, while only the relationship between tone and image is incorrect.
\\

\end{longtable}
\endgroup
\newpage

\begingroup
\fontsize{7}{8.5}\selectfont
\renewcommand{\arraystretch}{1.35}
\setlength{\tabcolsep}{4pt}

\begin{longtable}{@{} >{\centering\arraybackslash}m{2.0cm} >{\raggedright\arraybackslash}m{3.2cm} >{\raggedright\arraybackslash}m{8.8cm} @{}}

\toprule
\textbf{Category} & \textbf{Perturbation Dimension} & \textbf{Perturbation Description} \\
\midrule
\endfirsthead

\multicolumn{3}{@{}l}{\small \tablename~\thetable\ -- \textit{continued from previous page}} \\[4pt]
\toprule
\textbf{Category} & \textbf{Perturbation Dimension} & \textbf{Description} \\
\midrule
\endhead

\midrule
\multicolumn{3}{r@{}}{\small \textit{continued on next page}} \\
\endfoot

\bottomrule
\\[-6pt]
\caption{Comprehensive taxonomy of Text-to-Image (T2I) perturbations, with detailed descriptions of various perturbation dimensions.}
\label{tab:t2i-perturbation-desc}
\endlastfoot

& Object Substitution
& Replace the primary object with a visually or semantically similar alternative that preserves the scene's composition and spatial layout. The substitute should share enough surface-level properties with the original to pass cursory inspection, making the identity-level deviation detectable only through explicit verification against the prompt.
\\
\cmidrule(lr){2-3}

& Object Addition/Omission
& Add or remove discrete objects from the scene, including omitting a non-critical but explicitly requested element or altering the exact count of repeated objects. The modification should leave the scene visually complete and well-composed, exploiting the evaluator's reliance on approximate plurality judgments rather than precise enumeration.
\\
\cmidrule(lr){2-3}

& Attribute Manipulation
& Alter or swap object attributes, such as color, texture, material, or size, while keeping the core object identity intact. This includes cross-object attribute swapping, where correct attributes present in the scene are assigned to the wrong entities, testing whether evaluators verify attribute-to-entity correspondence rather than simply checking for attribute presence.
\\
\cmidrule(lr){2-3}

\multirow{-4}{*}{\textbf{\colorbox{t2i_vf}{\shortstack[c]{Visual\\Fidelity\\(VF)}}}}
& Spatial Manipulation
& Alter the requested positional or directional relationship between objects, ranging from slight shifts that remain spatially plausible to complete reversals of the specified arrangement. The manipulated layout should still produce a compositionally balanced image, so the spatial violation surfaces only upon structured comparison with the prompt's explicit constraints.
\\
\cmidrule(lr){2-3}

& Scale Distortion
& Force objects into visually plausible but factually incorrect relative proportions while maintaining full photographic realism in lighting, perspective, and depth. The scale violation should be detectable only through semantic reasoning about real-world object sizes, not through rendering artifacts or visual inconsistencies.
\\
\cmidrule(lr){2-3}

& Constraint Violation
& Introduce subtle deviations from the prompt's explicit behavioral instructions or negative constraints without altering the core subjects. This includes modifying a requested action to a visually similar but semantically distinct behavior, or seamlessly blending a forbidden element into the scene's periphery to test whether evaluators enforce the full specificity of prompt directives.
\\

\midrule

& Incomplete Scene
& Render the scene only partially, leaving large sections blank, unfinished, or as rough sketches despite a prompt requesting a complete image. The rendered portions should exhibit high visual quality, creating a stark contrast that tests whether evaluators assess holistic scene completeness or focus narrowly on the quality of whatever content is present.
\\
\cmidrule(lr){2-3}

& Missing Context
& Remove critical environmental elements or strip away the grounding background that the prompt implicitly or explicitly requires, leaving subjects in a featureless or decontextualized environment. The subjects themselves should be rendered with full detail, ensuring the deviation is detectable only through scene-level contextual reasoning rather than foreground quality assessment.
\\
\cmidrule(lr){2-3}

& Style Inconsistency
& Mix incompatible artistic styles or rendering techniques within a single image that violate the requested visual medium. Each stylistic region should appear well-rendered in isolation but produce jarring discontinuities at the boundaries, testing evaluator sensitivity to global stylistic unity versus local rendering quality.
\\
\cmidrule(lr){2-3}

\multirow{-4}{*}{\textbf{\colorbox{t2i_sc}{\shortstack[c]{Scene\\Coherence\\(SC)}}}}
& Theme Conflict
& Insert objects or backgrounds that create severe temporal, cultural, or environmental anachronisms while maintaining flawless visual integration in lighting, perspective, and resolution. The conflicting element should look photographically natural within the scene, making the violation detectable only through semantic reasoning about contextual appropriateness.
\\
\cmidrule(lr){2-3}

& Disorganized Composition
& Arrange scene elements in a chaotic or spatially incoherent layout that fails to integrate multiple entities cohesively. Objects may intersect without proper occlusion, overlap in physically impossible configurations, or float without structural grounding, producing a scene that lacks the organizational depth hierarchy expected from well-formed generation.
\\
\cmidrule(lr){2-3}

& Overcrowding
& Introduce an excessive number of semantically tangential objects into the scene to create visual clutter and attention interference. The added objects should be individually well-rendered and thematically adjacent to the scene's content, making the overcrowding feel like a plausible exaggeration rather than random noise injection, while still degrading compositional clarity.
\\

\midrule
\pagebreak

& Causal Violation
& Introduce explicitly contradictory elements that defy the prompt's basic premise, or depict a physical reaction occurring without its necessary antecedent cause. The violation should be embedded within an otherwise well-composed and visually coherent scene, requiring the evaluator to reason about causal consistency rather than detect surface-level rendering errors.
\\
\cmidrule(lr){2-3}

& Physics Manipulation
& Break fundamental physical laws, including gravity, optics, fluid dynamics, or shadow geometry, in ways that require spatial deduction to detect. The violation should be rendered with full photographic realism, appearing natural at first glance but revealing physically impossible configurations upon careful analysis of light direction, support structures, or material behavior.
\\
\cmidrule(lr){2-3}

\multirow{-3}{*}{\textbf{\colorbox{t2i_pp}{\shortstack[c]{Physical\\Plausibility\\(PP)}}}}
& State/Transformation Failure
& Depict an object in a condition that contradicts the transformation described in the prompt, misrepresenting the expected end-state of a physical or temporal process. The depicted state should be internally consistent as a standalone image, making the failure detectable only by reasoning about what the described process should have produced as its outcome.
\\
\cmidrule(lr){2-3}

& Functional Absurdity
& Depict objects being held, operated, or interacted with in ways that are physically dangerous, mechanically impossible, or functionally nonsensical. The absurdity should be subtle enough that the scene composition appears purposeful, testing whether evaluators can identify violations of functional common sense about how objects are designed to be used.
\\
\cmidrule(lr){2-3}

& Literalized Idioms
& Force a literal visual rendering of a metaphorical, idiomatic, or abstract expression from the prompt, treating figurative language as a physical description. The resulting image should be technically well-composed and visually coherent as a standalone scene, but fundamentally misinterpret the figurative intent by depicting its surface-level lexical meaning.
\\

\midrule

& Text/Typographic Corruption
& Replace requested text with subtly incorrect alternatives designed to exploit fast-reading priors, including similarly spelled incorrect words, visually similar letter-to-numeral substitutions, or typographic near-misses that preserve overall word shape and length. The corruption should pass rapid visual scanning while failing under deliberate character-by-character verification.
\\
\cmidrule(lr){2-3}

\multirow{-2}{*}{\textbf{\colorbox{t2i_tr}{\shortstack[c]{Text\\Rendering\\(TR)}}}}
& Incomplete Rendering
& Omit critical components of rendered text, symbols, or diagrammatic elements, leaving obvious gaps or truncations where information should appear. The rendered portions should exhibit high typographic quality and correct formation, making the omission detectable only through completeness verification rather than rendering quality assessment.
\\
\cmidrule(lr){2-3}

& Background Misrendering
& Render text or signage with perfect typographic accuracy but subtly distort, remove, or misplace its immediate structural support or environmental context. The text itself passes close inspection, but its physical mounting, surface, or spatial integration within the scene violates real-world expectations about how such elements are situated.
\\
\cmidrule(lr){2-3}

& Mislabeled Symbols/Diagrams
& Preserve the visual structure and layout of a diagram, chart, or symbolic element with high fidelity, but swap or reassign the text labels, annotations, or specific symbols pointing to different components. The structural scaffold remains correct, making the mislabeling detectable only through semantic verification of label-to-element correspondence rather than visual quality assessment.
\\

\end{longtable}
\endgroup


\providecommand{\iitcolorword}[2]{\colorbox{#1}{\strut #2}}

\providecommand{\iitanswerpair}[2]{%
  \begin{tcbraster}[raster columns=2,raster column skip=0.02\linewidth,raster equal height=rows]
    \begin{tcolorbox}[enhanced,sharp corners,boxrule=0.4pt,colframe=black,colback=white,colbacktitle=white,coltitle=black,fonttitle=\bfseries,title=Gold,left=4pt,right=4pt,top=4pt,bottom=4pt,before skip=0pt,after skip=0pt]
      \raggedright\footnotesize\sloppy #1
    \end{tcolorbox}
    \begin{tcolorbox}[enhanced,sharp corners,boxrule=0.4pt,colframe=black,colback=white,colbacktitle=white,coltitle=black,fonttitle=\bfseries,title=Perturbed,left=4pt,right=4pt,top=4pt,bottom=4pt,before skip=0pt,after skip=0pt]
      \raggedright\footnotesize\sloppy #2
    \end{tcolorbox}
  \end{tcbraster}
}

\providecommand{\iitexample}[5]{%
  \noindent
  \begin{minipage}[t]{0.58\linewidth}
    \vspace{0pt}
    \raggedright\sloppy
    \textbf{Prompt:} #1\par
    \textbf{Rationale:} #3\par
  \end{minipage}\hfill
  \begin{minipage}[t]{0.37\linewidth}
    \vspace{0pt}
    \centering
    \includegraphics[width=\linewidth,height=0.18\textheight,keepaspectratio]{#2}
  \end{minipage}\par
  \vspace{0.25em}
  \iitanswerpair{#4}{#5}\par
}
\clearpage
\section{Detailed Examples for I2T tasks}

\subsection{Visual Grounding}

\subsubsection{Entity Substitution}
\vspace{-0.2em}

\iitexample{Count and categorize the distinct types of fruit visible in the scene, noting exactly which fruits are contained inside the baskets versus those lying loose on the table.}{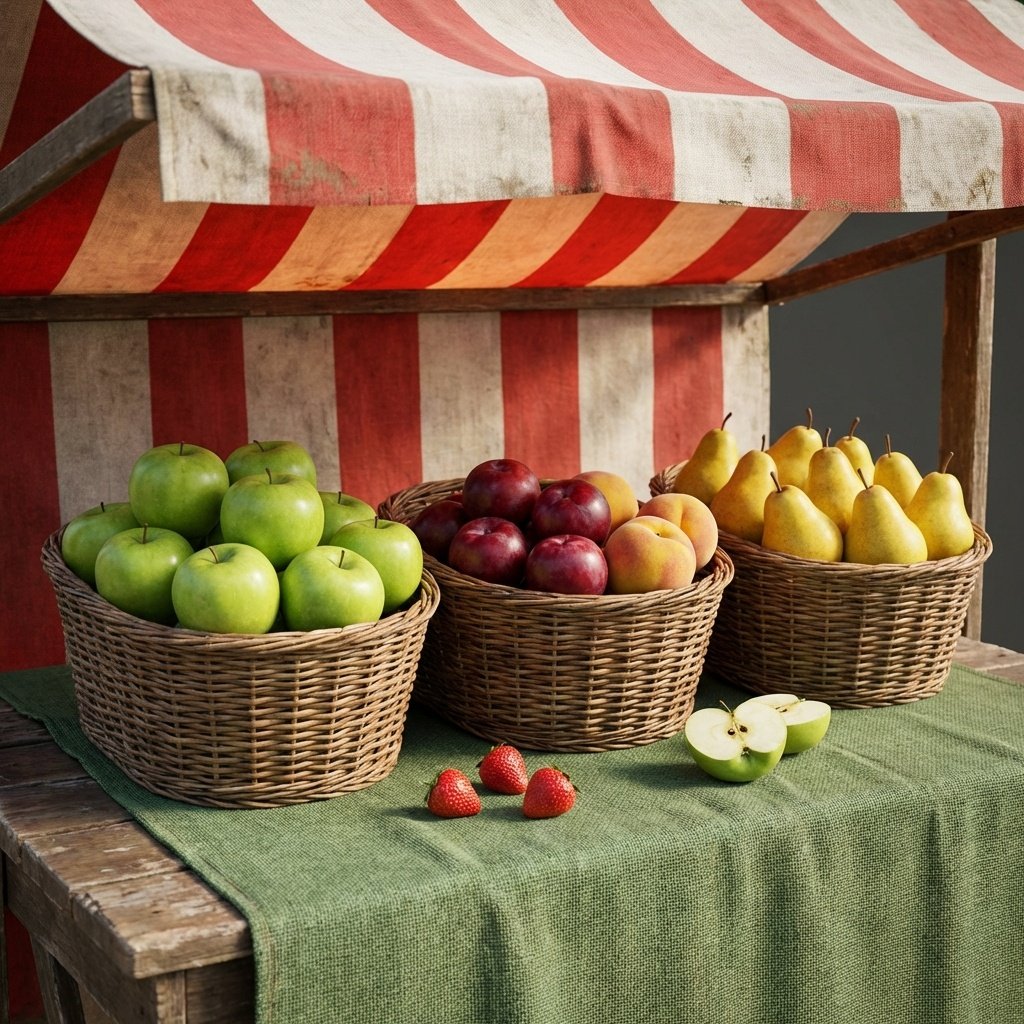}{The perturbation replaces the "strawberries" with "raspberries" in both the introductory list and the detailed breakdown of the loose fruits. This substitution is highly subtle because both are small, red berries that are commonly found together in fruit displays, keeping the text contextually plausible and semantically.}{... scene: green apples, dark red plums, yellow peaches, yellow pears, \iitcolorword{red!18}{strawberries,} and a half-eaten green apple. Inside the wicker baskets, there ... green tablecloth in front of the baskets are 3 red \iitcolorword{red!18}{strawberries} and 1 half-eaten green apple.}{... scene: green apples, dark red plums, yellow peaches, yellow pears, \iitcolorword{green!18}{raspberries,} and a half-eaten green apple. Inside the wicker baskets, there ... green tablecloth in front of the baskets are 3 red \iitcolorword{green!18}{raspberries} and 1 half-eaten green apple.}

\vspace{0.9em}

\subsubsection{Attribute Distortion}
\vspace{-0.2em}

\iitexample{Break down the exact count and toppings of the cupcakes on the top shelf, and identify the baked goods displayed on the bottom shelf.}{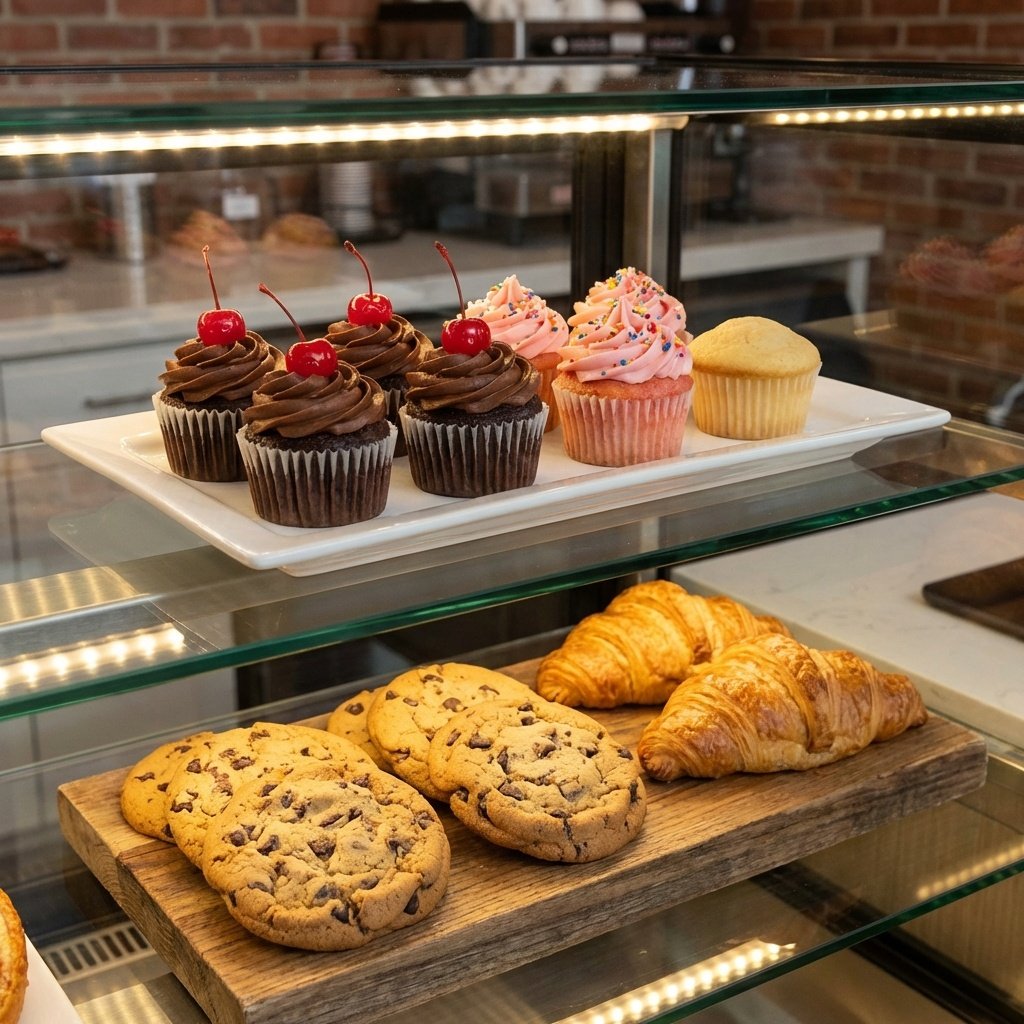}{The perturbation alters the description of the toppings on the pink cupcakes, changing "rainbow sprinkles" to "chocolate sprinkles." This subtle attribute substitution is difficult for a. Consequently, the evaluator cannot rely on textual context alone and must carefully inspect the fine visual details of the pink cupcakes in the image to verify.}{... Three of the cupcakes have pink strawberry frosting covered in \iitcolorword{red!18}{rainbow} sprinkles, while one cupcake is completely plain with no frosting ...}{... Three of the cupcakes have pink strawberry frosting covered in \iitcolorword{green!18}{chocolate} sprinkles, while one cupcake is completely plain with no frosting ...}

\vspace{0.9em}

\subsubsection{Spatial Relation Swap}
\vspace{-0.2em}

\iitexample{How does the cardboard cutout of the cartoon tiger interact spatially with the cereal boxes on the left shelf?}{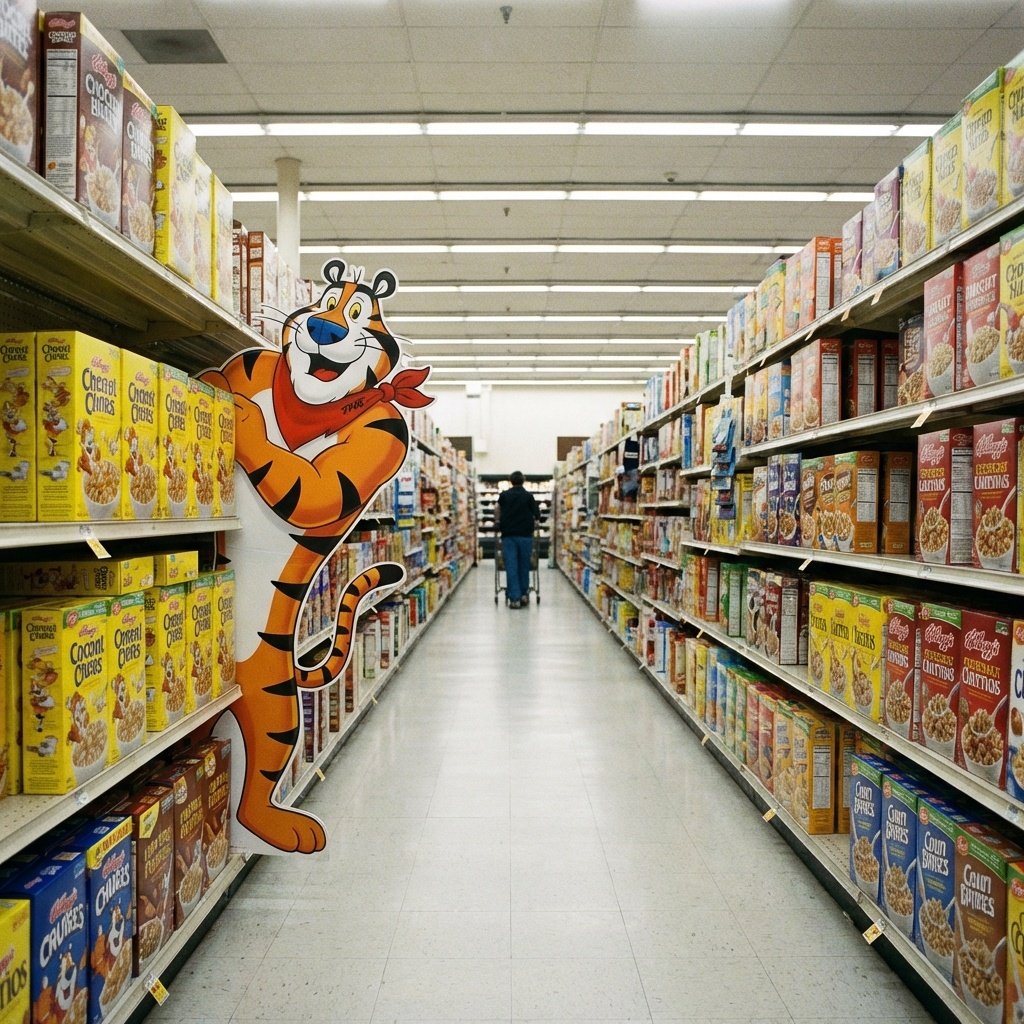}{I altered the depth and spatial relationships between the cardboard cutout and the cereal boxes. Instead of describing the cutout as protruding into the aisle and occluding the boxes behind it, the perturbed answer claims it is recessed into the shelf.}{... on the left side, \iitcolorword{red!18}{protruding} \iitcolorword{red!18}{outward} from the edge of the shelving \iitcolorword{red!18}{into} the main walkway. Because it stands \iitcolorword{red!18}{closer} \iitcolorword{red!18}{to} the center of the aisle and \iitcolorword{red!18}{in} front of \iitcolorword{red!18}{the} \iitcolorword{red!18}{shelf} \iitcolorword{red!18}{surface,} it \iitcolorword{red!18}{creates} a \iitcolorword{red!18}{visual} \iitcolorword{red!18}{barrier.} This placement \iitcolorword{red!18}{completely} \iitcolorword{red!18}{occludes} the specific section of yellow cereal boxes situated immediately \iitcolorword{red!18}{behind} it on the shelf. ...}{... \iitcolorword{green!18}{recessed} \iitcolorword{green!18}{inward} from the edge of the shelving \iitcolorword{green!18}{away} \iitcolorword{green!18}{from} the main walkway. Because it stands \iitcolorword{green!18}{further} \iitcolorword{green!18}{from} the center of the aisle and \iitcolorword{green!18}{behind} \iitcolorword{green!18}{the} front \iitcolorword{green!18}{row} of \iitcolorword{green!18}{boxes,} it \iitcolorword{green!18}{serves} \iitcolorword{green!18}{as} a \iitcolorword{green!18}{background} \iitcolorword{green!18}{element.} This placement \iitcolorword{green!18}{means} \iitcolorword{green!18}{it} \iitcolorword{green!18}{is} \iitcolorword{green!18}{partially} \iitcolorword{green!18}{occluded} \iitcolorword{green!18}{by} the specific section of yellow cereal boxes situated immediately \iitcolorword{green!18}{in} \iitcolorword{green!18}{front} \iitcolorword{green!18}{of} ...}

\vspace{0.9em}

\subsubsection{Phantom Details Injection}
\vspace{-0.2em}

\iitexample{How do the lamppost, the pedestrians, and the yellow taxi relate to each other in terms of depth and visibility?}{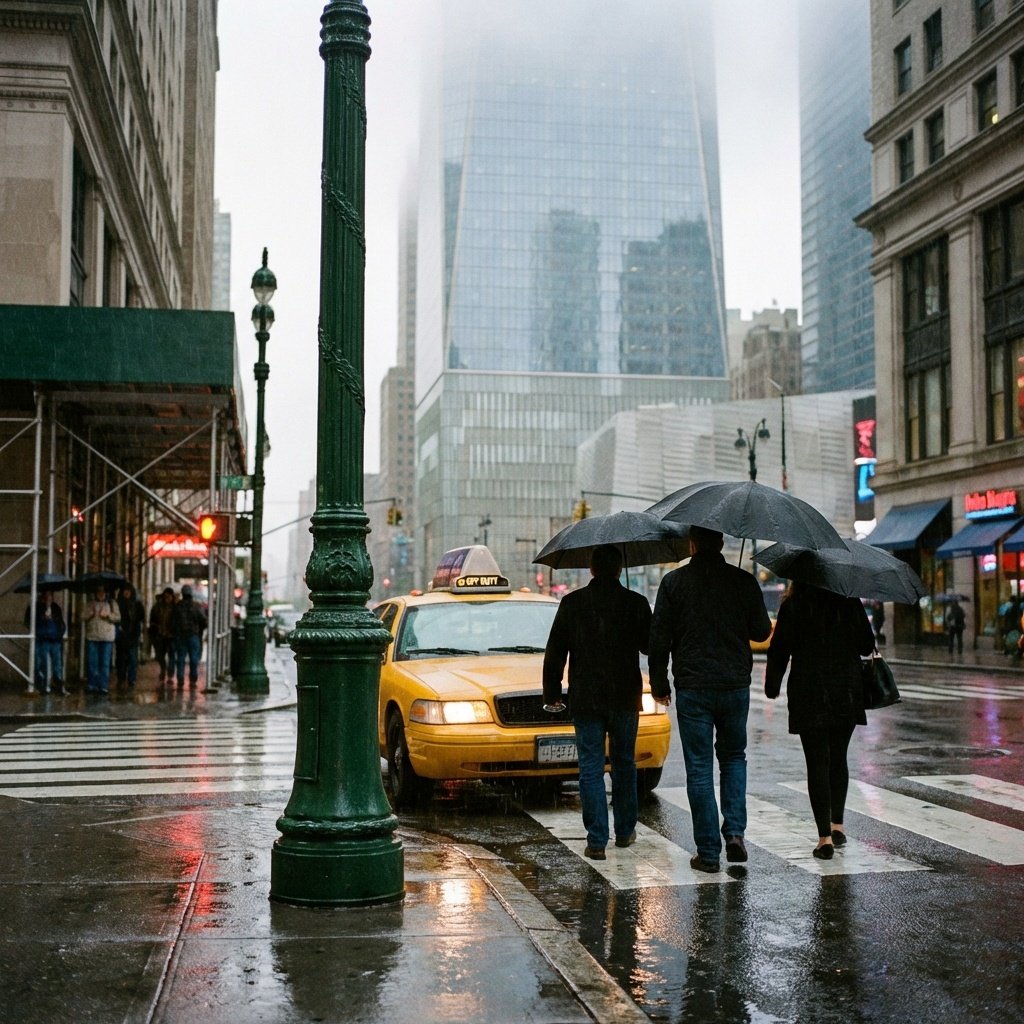}{I introduced two fabricated details into the gold answer: a "glowing pedestrian 'Walk' signal" attached to the foreground lamppost, and a "bright red shopping bag" carried. These additions are contextually highly plausible for a busy, rainy urban street scene, making them easy for a VLM evaluator to gloss over if it relies.}{The green \iitcolorword{red!18}{lamppost} is the closest object to the viewer, standing in the immediate foreground. The pedestrians with \iitcolorword{red!18}{umbrellas} are positioned slightly further back, crossing the street in the ...}{The green \iitcolorword{green!18}{lamppost,} \iitcolorword{green!18}{featuring} \iitcolorword{green!18}{a} \iitcolorword{green!18}{glowing} \iitcolorword{green!18}{pedestrian} \iitcolorword{green!18}{'Walk'} \iitcolorword{green!18}{signal,} is the closest object to the viewer, standing in the immediate foreground. The pedestrians with \iitcolorword{green!18}{umbrellas,} \iitcolorword{green!18}{one} \iitcolorword{green!18}{of} \iitcolorword{green!18}{whom} \iitcolorword{green!18}{is} \iitcolorword{green!18}{carrying} \iitcolorword{green!18}{a} \iitcolorword{green!18}{bright} \iitcolorword{green!18}{red} \iitcolorword{green!18}{shopping} \iitcolorword{green!18}{bag,} are positioned slightly further back, crossing the street in the ...}

\vspace{0.9em}



\subsubsection{Over Generalization}
\vspace{-0.2em}

\iitexample{Describe the position of the orange tabby cat and explain what prevents us from seeing its full body.}{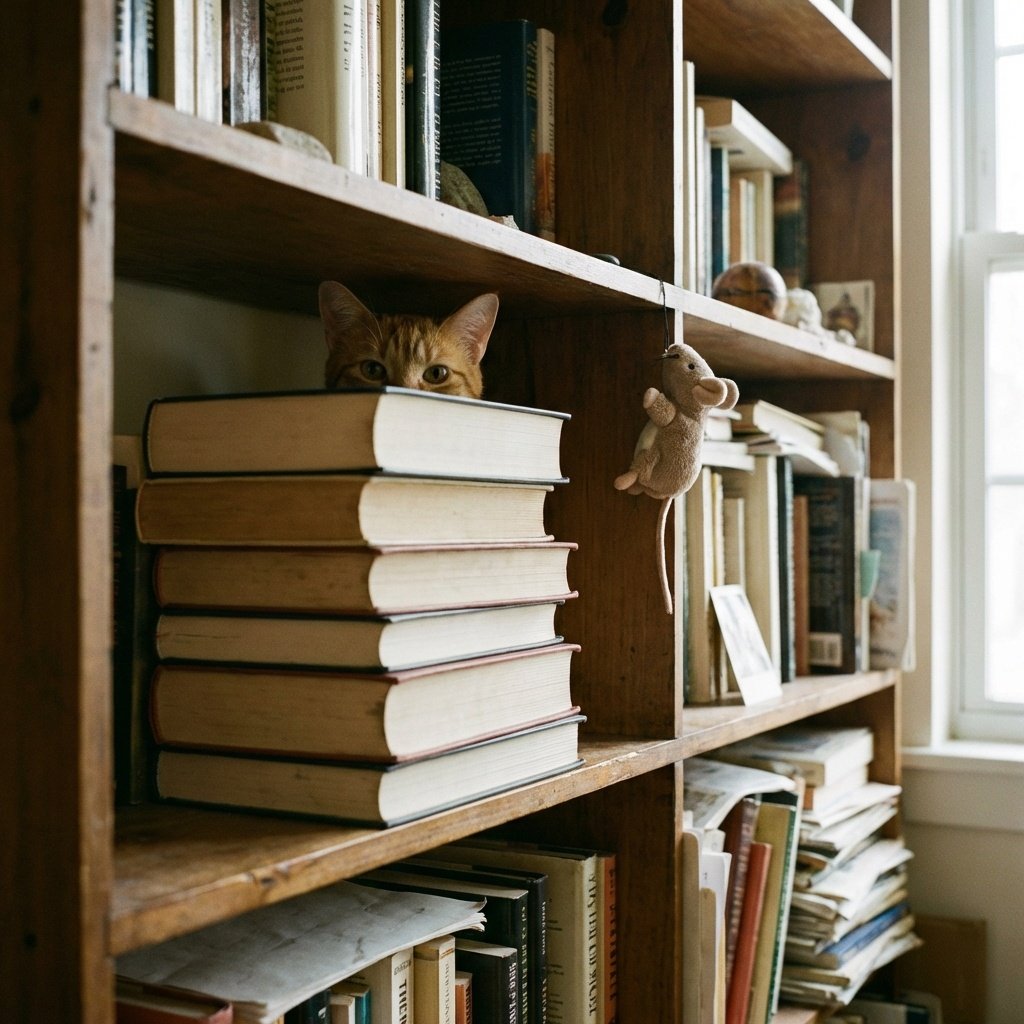}{To create this perturbation, I replaced specific, identifying visual details from the gold answer with broader hypernyms and generalized terms. For example, "orange tabby cat" was changed to "feline" and "animal," "tall, horizontal stack of thick hardcover books" was reduced to a "stack of items," and.}{... It is \iitcolorword{red!18}{crouching} directly behind a \iitcolorword{red!18}{tall,} \iitcolorword{red!18}{horizontal} stack of \iitcolorword{red!18}{thick} ... the foreground relative to the \iitcolorword{red!18}{cat.} Consequently, the \iitcolorword{red!18}{cat's} lower \iitcolorword{red!18}{face,} \iitcolorword{red!18}{torso,} \iitcolorword{red!18}{and} \iitcolorword{red!18}{legs} \iitcolorword{red!18}{are} completely occluded, leaving only its \iitcolorword{red!18}{ears,} \iitcolorword{red!18}{eyes,} \iitcolorword{red!18}{and} \iitcolorword{red!18}{the} \iitcolorword{red!18}{top} \iitcolorword{red!18}{of} \iitcolorword{red!18}{its} \iitcolorword{red!18}{head} ...}{... It is \iitcolorword{green!18}{situated} directly behind a stack of \iitcolorword{green!18}{items.} This stack of \iitcolorword{green!18}{items} acts as a physical barrier in the foreground relative to the \iitcolorword{green!18}{animal.} Consequently, the \iitcolorword{green!18}{animal's} lower \iitcolorword{green!18}{body} \iitcolorword{green!18}{is} completely occluded, leaving only its \iitcolorword{green!18}{upper} \iitcolorword{green!18}{features} visible.}

\vspace{0.9em}

\subsubsection{Important Detail Omission}
\vspace{-0.2em}

\iitexample{Classify and count the different types of mammals in the pasture, and describe the position and action of the Border Collie relative to the other animals.}{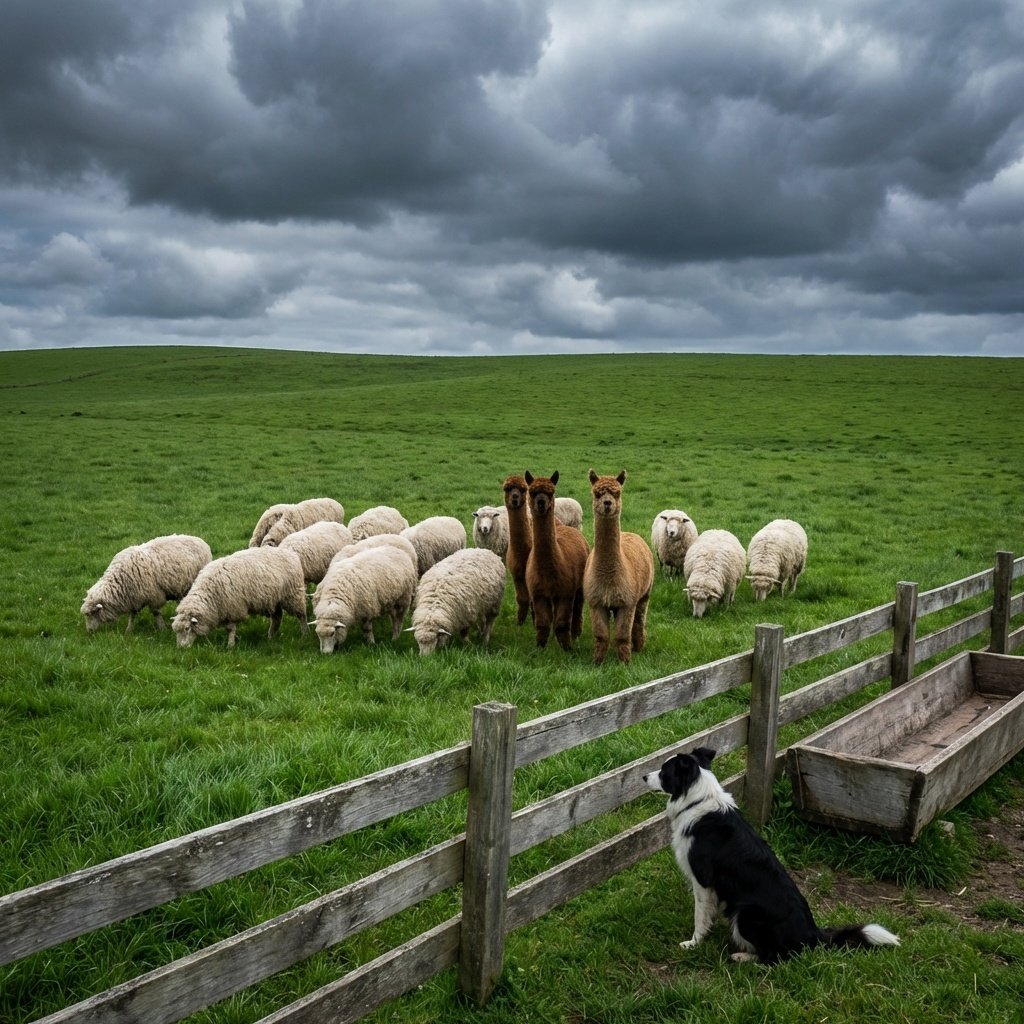}{In this perturbation, the specific numerical counts for the individual sheep (12) and alpacas (3) were removed, along with the description of the Border Collie's action. This omission is highly subtle because the resulting text still provides a fluent, factually accurate, and visually grounded description of the overall scene.}{... majority of the group consists of \iitcolorword{red!18}{12} fluffy white sheep that are actively grazing ... hillside. Intermingled among the sheep are \iitcolorword{red!18}{3} tall, brown alpacas looking toward the camera. Additionally, there is \iitcolorword{red!18}{1} black and white Border Collie. The dog ... \iitcolorword{red!18}{fence,} \iitcolorword{red!18}{where} \iitcolorword{red!18}{it} \iitcolorword{red!18}{is} \iitcolorword{red!18}{sitting} \iitcolorword{red!18}{attentively} \iitcolorword{red!18}{and} \iitcolorword{red!18}{facing} \iitcolorword{red!18}{the} \iitcolorword{red!18}{flock} \iitcolorword{red!18}{of} \iitcolorword{red!18}{sheep} \iitcolorword{red!18}{and} \iitcolorword{red!18}{alpacas.}}{... tall, brown alpacas looking toward the camera. Additionally, there is \iitcolorword{green!18}{a} black and white Border Collie. The dog is positioned in the foreground near a wooden \iitcolorword{green!18}{fence.}}

\vspace{0.9em}

\
\subsection{Semantic Interpretation}

\subsubsection{Contextual Depth Reduction}
\vspace{-0.2em}

\iitexample{Based on the clues in the kitchen, what sequence of events led to this current situation?}{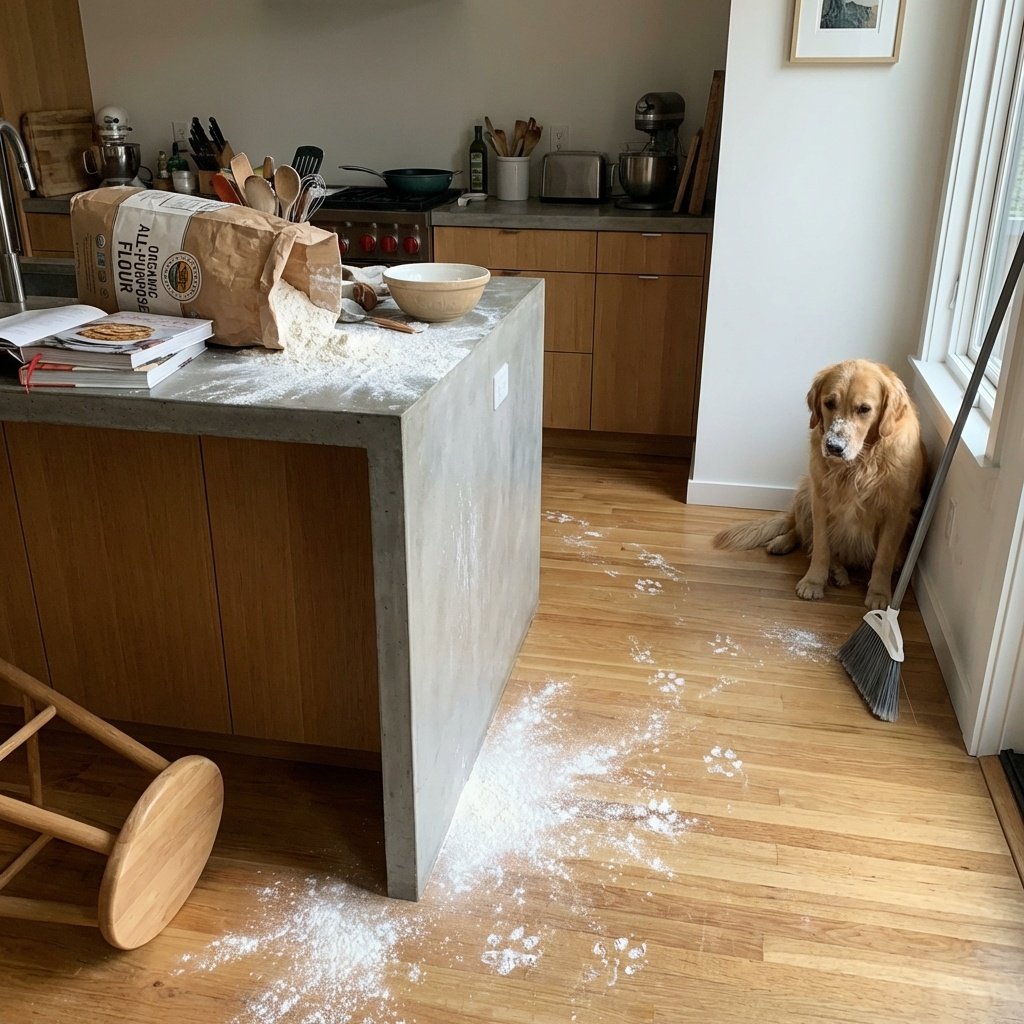}{I removed all narrative causality, emotional resonance, and inferential depth from the gold answer, stripping away the dog's implicit intent and feelings. Instead of explaining how the dog used the stool to reach the counter, caused the spill, or retreated out of guilt, the perturbed answer merely lists.}{... \iitcolorword{red!18}{golden} \iitcolorword{red!18}{retriever} \iitcolorword{red!18}{likely} \iitcolorword{red!18}{entered} \iitcolorword{red!18}{the} \iitcolorword{red!18}{kitchen} \iitcolorword{red!18}{and} \iitcolorword{red!18}{jumped} \iitcolorword{red!18}{up} \iitcolorword{red!18}{onto} \iitcolorword{red!18}{the} wooden stool \iitcolorword{red!18}{to} \iitcolorword{red!18}{reach} \iitcolorword{red!18}{the} \iitcolorword{red!18}{countertop.} \iitcolorword{red!18}{In} \iitcolorword{red!18}{doing} \iitcolorword{red!18}{so,} \iitcolorword{red!18}{it} \iitcolorword{red!18}{knocked} \iitcolorword{red!18}{the} \iitcolorword{red!18}{stool} over \iitcolorword{red!18}{and} \iitcolorword{red!18}{bumped} \iitcolorword{red!18}{into} ... \iitcolorword{red!18}{over} its snout. \iitcolorword{red!18}{Finally,} \iitcolorword{red!18}{realizing} \iitcolorword{red!18}{it} \iitcolorword{red!18}{made} \iitcolorword{red!18}{a} \iitcolorword{red!18}{mess,} \iitcolorword{red!18}{the} dog walked away from the counter, leaving \iitcolorword{red!18}{floury} paw prints \iitcolorword{red!18}{behind,} \iitcolorword{red!18}{and} \iitcolorword{red!18}{retreated} \iitcolorword{red!18}{to} the \iitcolorword{red!18}{corner} in \iitcolorword{red!18}{guilt.}}{The wooden stool \iitcolorword{green!18}{fell} over \iitcolorword{green!18}{onto} the \iitcolorword{green!18}{floor.} \iitcolorword{green!18}{The} bag of flour on the \iitcolorword{green!18}{island} \iitcolorword{green!18}{tipped} \iitcolorword{green!18}{over,} \iitcolorword{green!18}{spilling} \iitcolorword{green!18}{powder} \iitcolorword{green!18}{onto} \iitcolorword{green!18}{the} \iitcolorword{green!18}{counter} and \iitcolorword{green!18}{the} \iitcolorword{green!18}{ground.} The dog \iitcolorword{green!18}{got} \iitcolorword{green!18}{flour} \iitcolorword{green!18}{on} its snout. \iitcolorword{green!18}{The} dog \iitcolorword{green!18}{then} walked away from the counter, leaving paw prints \iitcolorword{green!18}{on} the \iitcolorword{green!18}{floor,} \iitcolorword{green!18}{and} \iitcolorword{green!18}{sat} \iitcolorword{green!18}{down} in \iitcolorword{green!18}{the} \iitcolorword{green!18}{corner.}}

\subsubsection{Cultural Misalignment}
\vspace{-0.2em}



\iitexample{Describe the sequence of events that resulted in the scene on the driveway.}{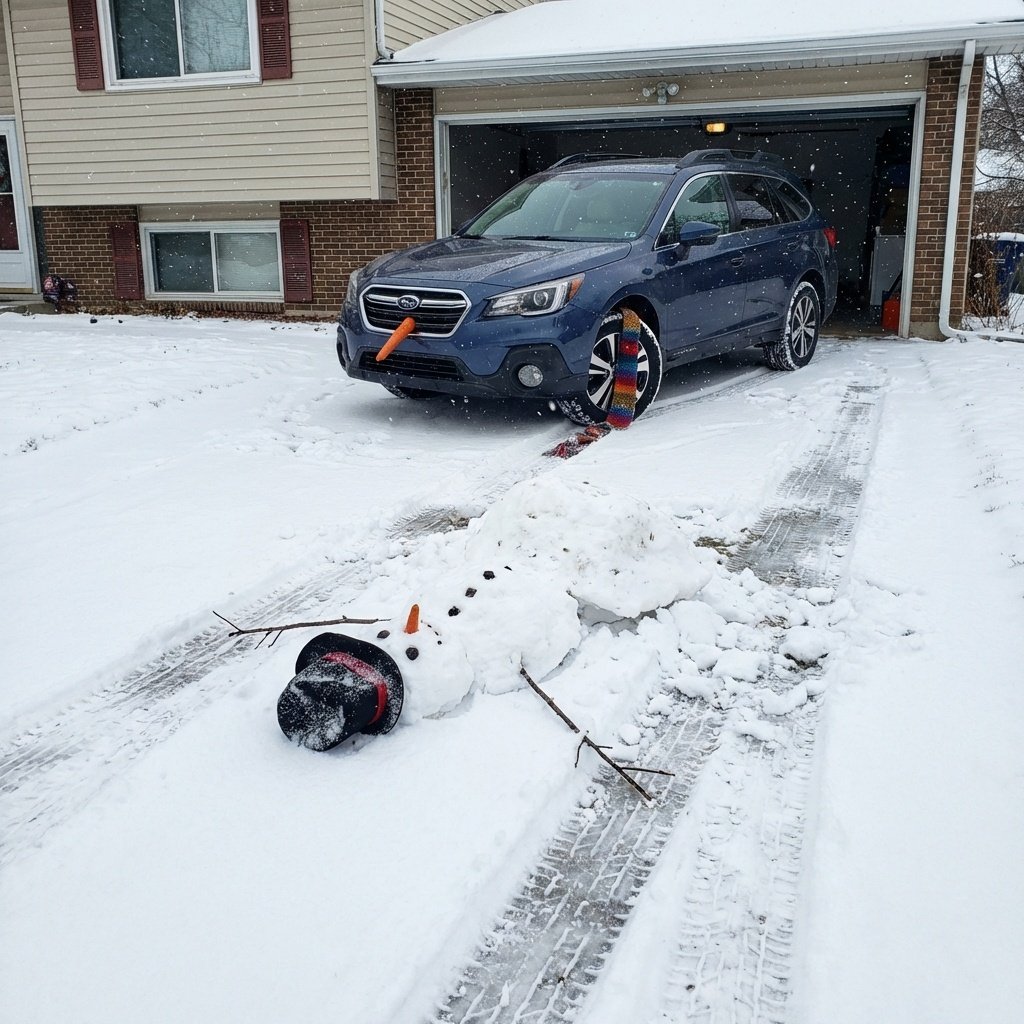}{The perturbation replaces standard Western snowman terminology ("snowman," "carrot nose," "scarf") with Japanese cultural equivalents ("yuki-daruma," "traditional wooden kokeshi nose," "tenugui"). This substitution is difficult for a VLM evaluator to detect based on text alone because the narrative remains logically coherent, grammatically correct, and contextually plausible for.}{... \iitcolorword{red!18}{snowman} right in the middle of the snowy driveway. Later, the driver of the blue sedan returned home, either not seeing the \iitcolorword{red!18}{snowman} or losing control on the ... in the process. The driver then continued into the garage and parked, remaining completely unaware that the \iitcolorword{red!18}{snowman's} \iitcolorword{red!18}{carrot} nose was stuck in the grille and its \iitcolorword{red!18}{scarf} ...}{... \iitcolorword{green!18}{yuki-daruma} right in the middle of the snowy driveway. Later, the driver of the blue sedan returned home, either not seeing the \iitcolorword{green!18}{snow} \iitcolorword{green!18}{figure} or losing control on ... process. The driver then continued into the garage and parked, remaining completely unaware that the \iitcolorword{green!18}{figure's} \iitcolorword{green!18}{traditional} \iitcolorword{green!18}{wooden} \iitcolorword{green!18}{kokeshi} nose was stuck in the grille and its \iitcolorword{green!18}{tenugui} ...}


\subsubsection{Logical Inconsistencies}


\vspace{0.9em}

\iitexample{Based on the body language and positioning of the individuals in the room, what is the likely professional dynamic taking place, and how are the bystanders reacting to the situation?}{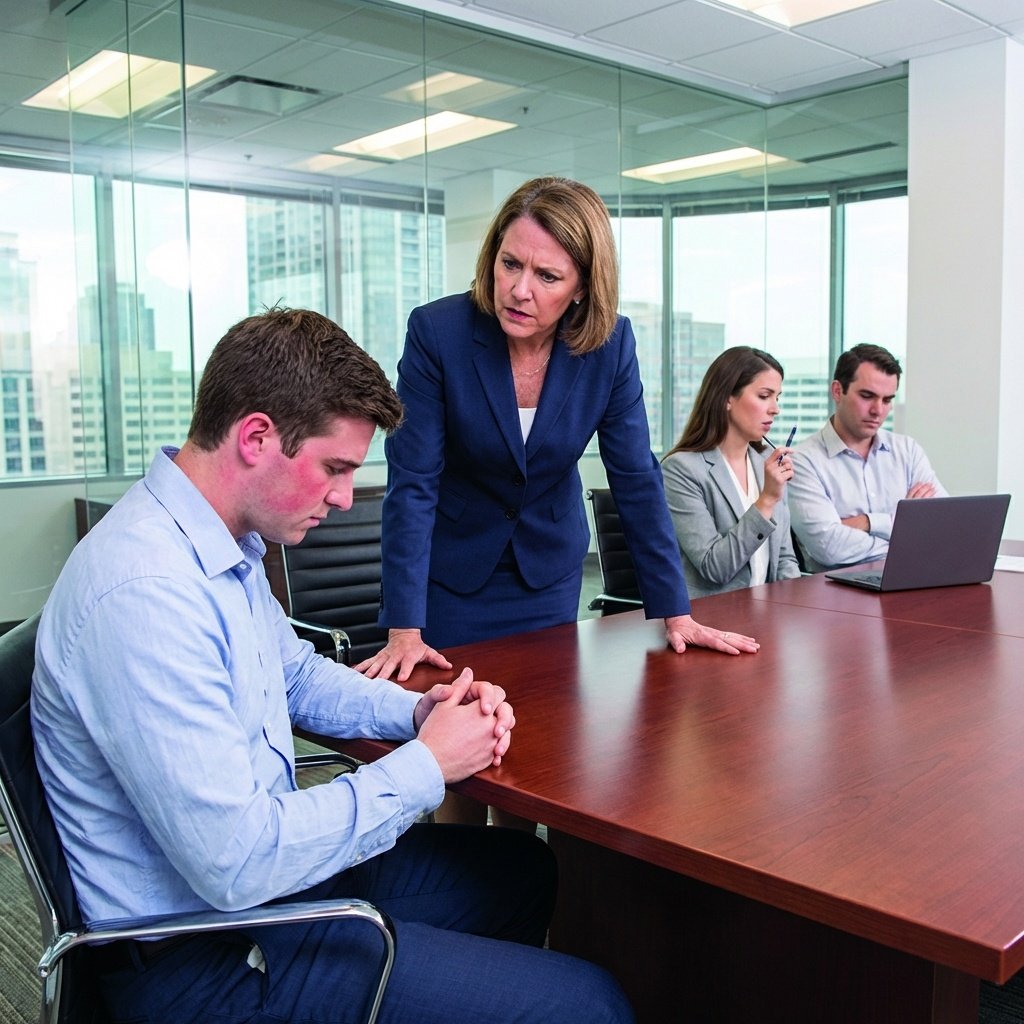}{The perturbation modifies the description of the bystanders' behavior, replacing their defensive postures (leaning away, crossing arms) with an assertion that they are leaning forward to. This alteration is challenging for a VLM to detect because taking notes is a plausible action in an office setting, allowing the text to read smoothly.}{... a desire to distance themselves from the conflict. By leaning \iitcolorword{red!18}{away,} \iitcolorword{red!18}{crossing} \iitcolorword{red!18}{their} \iitcolorword{red!18}{arms,} \iitcolorword{red!18}{avoiding} \iitcolorword{red!18}{eye} \iitcolorword{red!18}{contact,} \iitcolorword{red!18}{and} \iitcolorword{red!18}{fidgeting,} they signal their unease and a strong reluctance to become ...}{... a desire to distance themselves from the conflict. By leaning \iitcolorword{green!18}{forward} \iitcolorword{green!18}{to} \iitcolorword{green!18}{actively} \iitcolorword{green!18}{type} \iitcolorword{green!18}{notes} \iitcolorword{green!18}{on} \iitcolorword{green!18}{the} \iitcolorword{green!18}{shared} \iitcolorword{green!18}{laptop} \iitcolorword{green!18}{between} \iitcolorword{green!18}{them,} they signal their unease and a strong reluctance to become ...}

\subsection{Visual Reasoning}
\subsubsection{Numerical Errors}
\vspace{-0.2em}

\iitexample{Based on the infographic on the billboard, if the city successfully meets its stated goal for next year while the total number of daily commuters remains at exactly 100,000, how many non-active commuters (Public Transit and Personal Vehicle combined) will there be?}{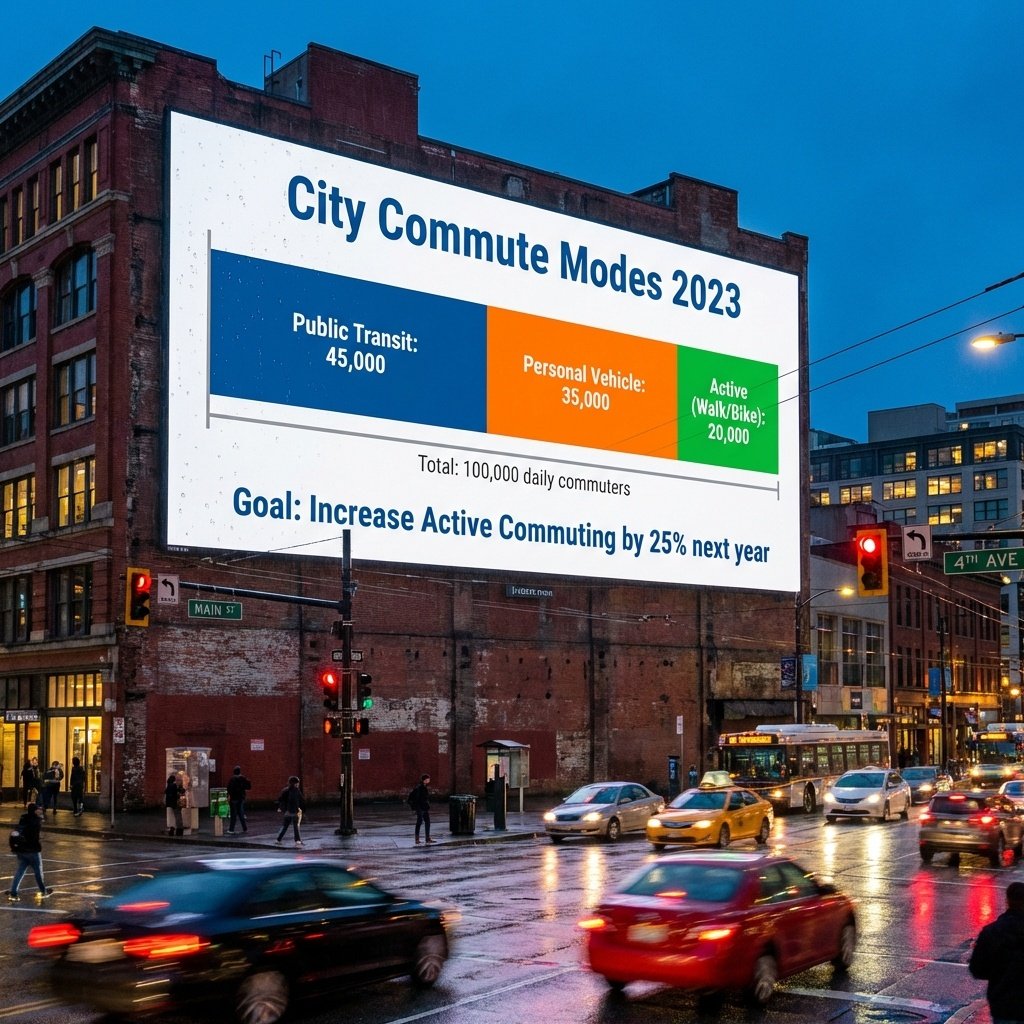}{I altered the stated goal for increasing active commuting from 25\% to 20\%. Consequently, I updated all the subsequent mathematical calculations to reflect this new percentage, changing the increase from 5,000 to 4,000, the new active target from 25,000.}{... increase this specific number by \iitcolorword{red!18}{25\%} next year. A \iitcolorword{red!18}{25\%} increase of 20,000 is \iitcolorword{red!18}{5,000,} which means the new target for active commuters is \iitcolorword{red!18}{25,000.} Since the scenario dictates the total ... the overall total. This leaves \iitcolorword{red!18}{75,000} commuters (100,000 - \iitcolorword{red!18}{25,000)} remaining to make up the combined ...}{... increase this specific number by \iitcolorword{green!18}{20\%} next year. A \iitcolorword{green!18}{20\%} increase of 20,000 is \iitcolorword{green!18}{4,000,} which means the new target for active commuters is \iitcolorword{green!18}{24,000.} Since the scenario dictates the total ... the overall total. This leaves \iitcolorword{green!18}{76,000} commuters (100,000 - \iitcolorword{green!18}{24,000)} remaining to make up the combined ...}

\vspace{0.9em}



\subsubsection{Procedural Reordering}
\vspace{-0.2em}

\iitexample{Based on the evidence in the room, which animal is responsible for the spilled mug, and how can you tell?}{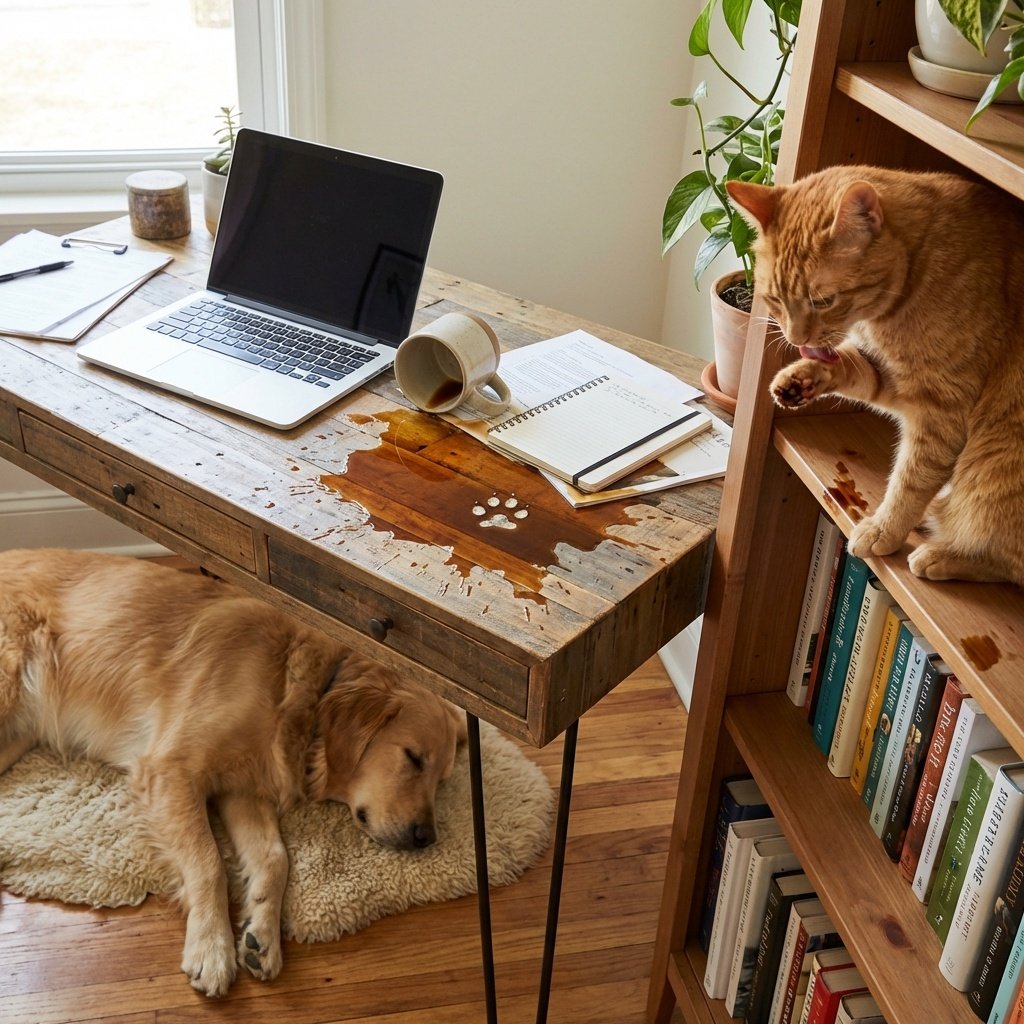}{The perturbation swaps the visual evidence attributed to the two animals, falsely claiming the cat has clean paws while the sleeping dog is actively licking a. This is difficult for a VLM evaluator to detect because the perturbed text maintains a highly plausible, logically sound deductive chain using the correct entities and.}{... spill. While there are two animals in the room, the \iitcolorword{red!18}{golden} \iitcolorword{red!18}{retriever} \iitcolorword{red!18}{sleeping} \iitcolorword{red!18}{under} the \iitcolorword{red!18}{desk} has completely clean paws. In contrast, the \iitcolorword{red!18}{tabby} \iitcolorword{red!18}{cat} \iitcolorword{red!18}{on} the \iitcolorword{red!18}{bookshelf} is actively licking its front right paw, which is stained ... spilled coffee. This chain of visual evidence directly implicates the \iitcolorword{red!18}{cat.}}{... spill. While there are two animals in the room, the \iitcolorword{green!18}{tabby} \iitcolorword{green!18}{cat} \iitcolorword{green!18}{on} the \iitcolorword{green!18}{bookshelf} has completely clean paws. In contrast, the \iitcolorword{green!18}{golden} \iitcolorword{green!18}{retriever} \iitcolorword{green!18}{sleeping} \iitcolorword{green!18}{under} the \iitcolorword{green!18}{desk} is actively licking its front right paw, which is stained ... spilled coffee. This chain of visual evidence directly implicates the \iitcolorword{green!18}{dog.}}

\vspace{0.9em}



\subsubsection{Causal Misattribution}
\vspace{-0.2em}

\iitexample{Given the road conditions and the trajectories of the vehicles, what is the imminent outcome of this traffic scenario?}{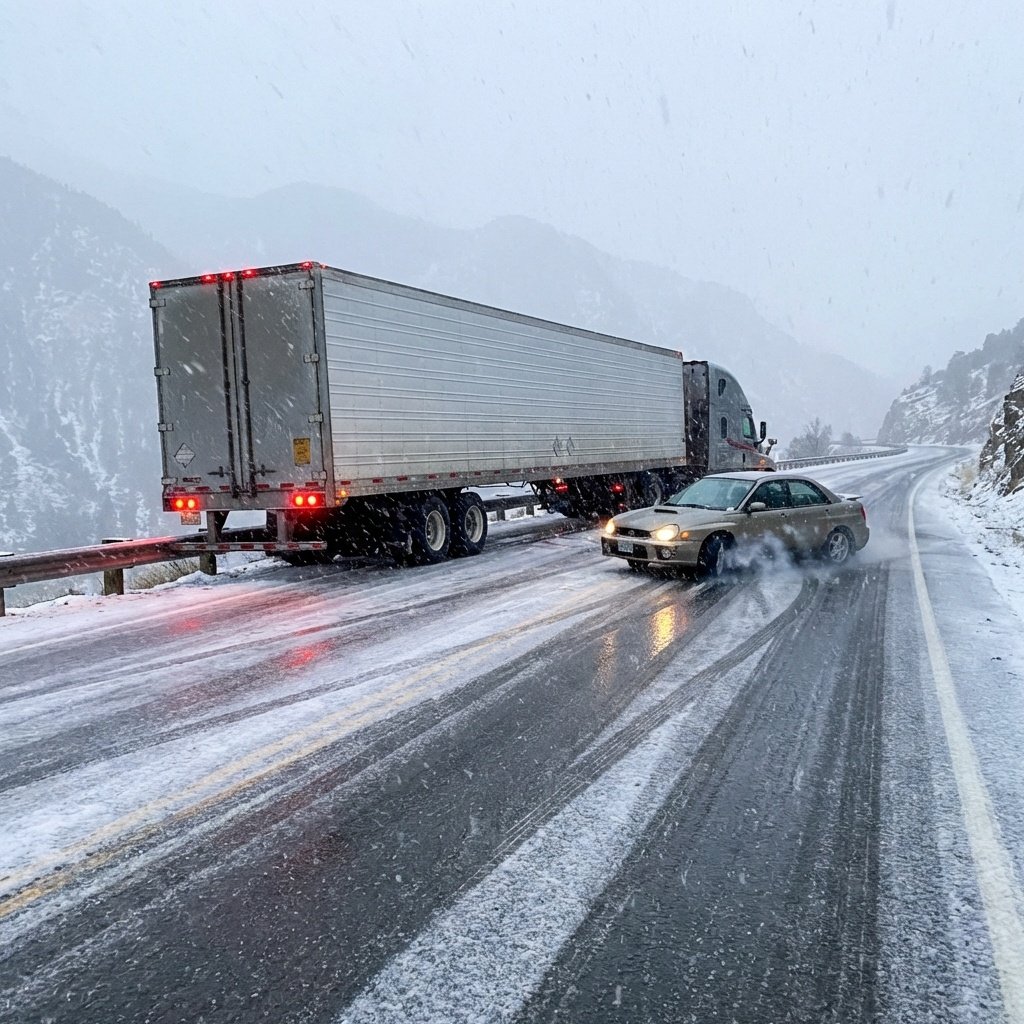}{I swapped the cause-and-effect relationship of the accident sequence and invented a fictitious but plausible reason for the sedan's visual state. Instead of the truck losing traction first and causing the sedan to lock its brakes in an emergency, the perturbed answer claims the sedan's aggressive acceleration.}{... \iitcolorword{red!18}{heavy} \iitcolorword{red!18}{semi-truck} has \iitcolorword{red!18}{lost} \iitcolorword{red!18}{traction} \iitcolorword{red!18}{on} the steep, icy incline, causing its trailer to jackknife and create an impassable barrier across the \iitcolorword{red!18}{entire} uphill lane. \iitcolorword{red!18}{The} \iitcolorword{red!18}{small} \iitcolorword{red!18}{sedan} ... shown by the \iitcolorword{red!18}{continuous} \iitcolorword{red!18}{slide} \iitcolorword{red!18}{marks,} the sedan is completely out of control and continuing its forward momentum up the \iitcolorword{red!18}{hill.} Because the driver \iitcolorword{red!18}{cannot} \iitcolorword{red!18}{steer} \iitcolorword{red!18}{or} \iitcolorword{red!18}{stop} ...}{... \iitcolorword{green!18}{small} \iitcolorword{green!18}{sedan} has \iitcolorword{green!18}{aggressively} \iitcolorword{green!18}{accelerated} \iitcolorword{green!18}{to} \iitcolorword{green!18}{climb} the steep, icy incline, causing its \iitcolorword{green!18}{tires} \iitcolorword{green!18}{to} \iitcolorword{green!18}{spin} \iitcolorword{green!18}{rapidly} \iitcolorword{green!18}{and} \iitcolorword{green!18}{kick} \iitcolorword{green!18}{up} \iitcolorword{green!18}{a} \iitcolorword{green!18}{continuous} \iitcolorword{green!18}{spray} \iitcolorword{green!18}{of} \iitcolorword{green!18}{snow.} \iitcolorword{green!18}{This} \iitcolorword{green!18}{sudden} \iitcolorword{green!18}{loss} ... the sedan is completely out of control and continuing its forward momentum up the \iitcolorword{green!18}{hill} \iitcolorword{green!18}{under} \iitcolorword{green!18}{heavy} \iitcolorword{green!18}{throttle.} Because the driver \iitcolorword{green!18}{is} \iitcolorword{green!18}{spinning} \iitcolorword{green!18}{the} \iitcolorword{green!18}{wheels} \iitcolorword{green!18}{instead} \iitcolorword{green!18}{of} \iitcolorword{green!18}{braking} ...}

\vspace{0.9em}

\subsubsection{Ungrounded Assumptions}
\vspace{-0.2em}

\iitexample{Based on the environmental conditions and the trajectories of the objects shown, what is the immediate fate of the red balloon?}{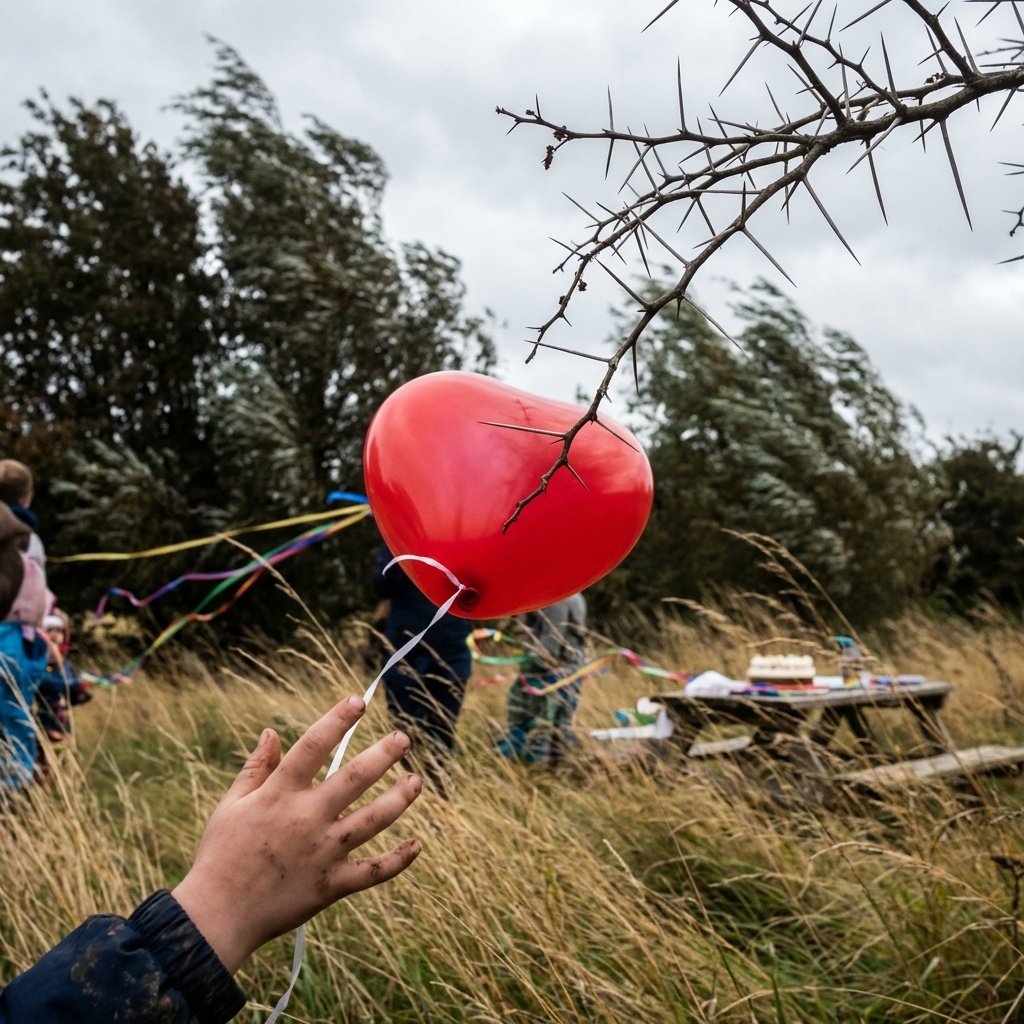}{This perturbation introduces unverified, definitive conclusions about the child's internal state and the broader social context of the scene. Specifically, it claims the child was "startled" into letting go and definitively states they will begin crying, bringing the background "birthday party" to a halt.}{The \iitcolorword{red!18}{child} has lost their grip on the string, completely releasing the ... will inevitably puncture the material and cause the balloon to \iitcolorword{red!18}{pop.}}{The \iitcolorword{green!18}{child,} \iitcolorword{green!18}{startled} \iitcolorword{green!18}{by} \iitcolorword{green!18}{a} \iitcolorword{green!18}{sudden} \iitcolorword{green!18}{gust} \iitcolorword{green!18}{of} \iitcolorword{green!18}{wind,} has lost their grip on the string, completely releasing the ... will inevitably puncture the material and cause the balloon to \iitcolorword{green!18}{pop,} \iitcolorword{green!18}{immediately} \iitcolorword{green!18}{bringing} \iitcolorword{green!18}{the} \iitcolorword{green!18}{birthday} \iitcolorword{green!18}{party} \iitcolorword{green!18}{in} \iitcolorword{green!18}{the} \iitcolorword{green!18}{background} \iitcolorword{green!18}{to} \iitcolorword{green!18}{a} \iitcolorword{green!18}{halt} \iitcolorword{green!18}{as} \iitcolorword{green!18}{the} \iitcolorword{green!18}{child} \iitcolorword{green!18}{begins} \iitcolorword{green!18}{to} \iitcolorword{green!18}{cry.}}


\subsubsection{Misinterpret Key Elements}
\vspace{-0.2em}

\iitexample{A passenger arrives at the station at 10:20 AM intending to take the Express to Washington. Where should they go to catch the next available train to their destination, and why?}{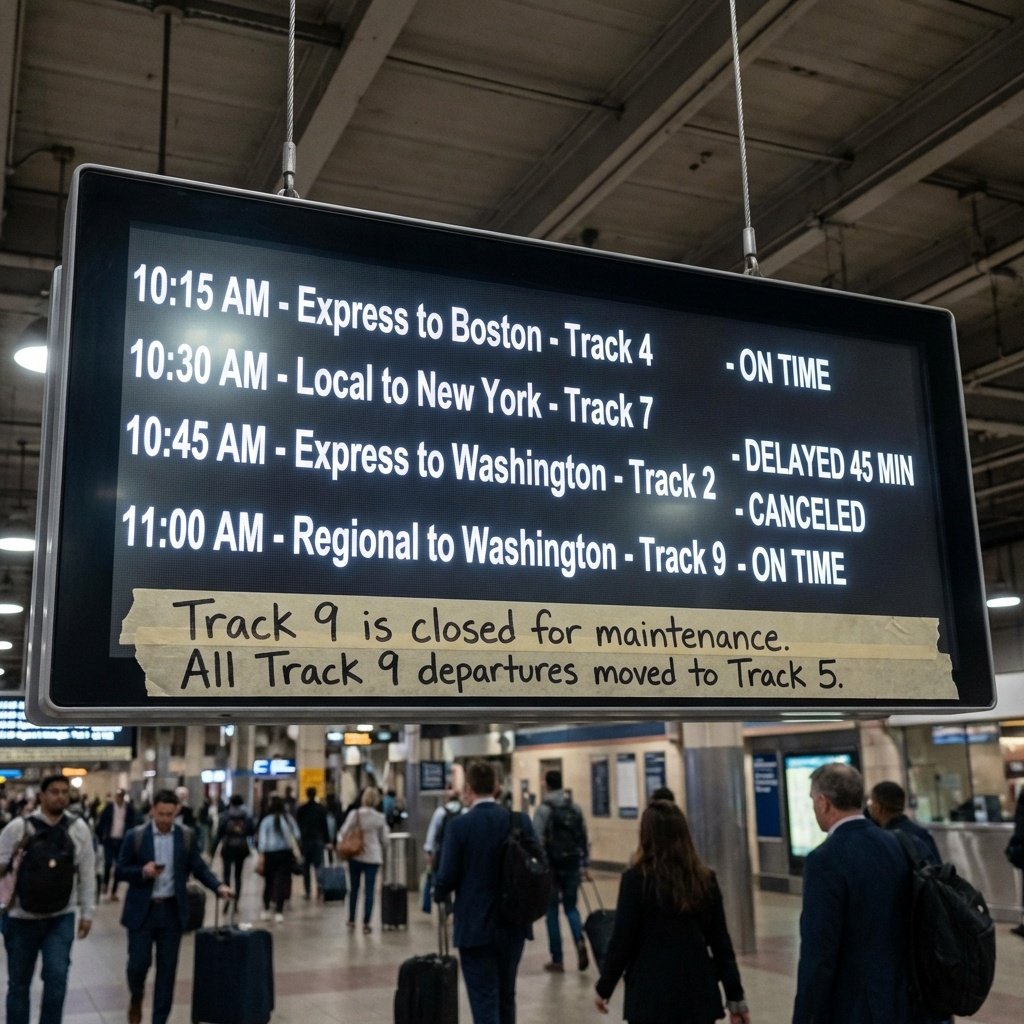}{This perturbation intentionally misaligns the train statuses in the right column with the corresponding train information in the left column. Specifically, it incorrectly associates the "DELAYED 45 MIN" status with the 10:45 AM train and the "CANCELED" status with the 11:00 AM train, effectively shifting the.}{... \iitcolorword{red!18}{5.} Their originally intended train, the 10:45 AM Express to Washington on Track 2, is marked on the digital departure board as \iitcolorword{red!18}{"CANCELED".} The board shows that the ... \iitcolorword{red!18}{is} \iitcolorword{red!18}{currently} \iitcolorword{red!18}{closed} \iitcolorword{red!18}{for} \iitcolorword{red!18}{maintenance} \iitcolorword{red!18}{and} \iitcolorword{red!18}{that} \iitcolorword{red!18}{all} \iitcolorword{red!18}{its} \iitcolorword{red!18}{scheduled} departures \iitcolorword{red!18}{have} \iitcolorword{red!18}{been} \iitcolorword{red!18}{moved.} Consequently, the passenger must go to Track \iitcolorword{red!18}{5} to \iitcolorword{red!18}{board} \iitcolorword{red!18}{the} \iitcolorword{red!18}{11:00} AM \iitcolorword{red!18}{Regional} ...}{... \iitcolorword{green!18}{2.} Their originally intended train, the 10:45 AM Express to Washington on Track 2, is marked on the digital departure board as \iitcolorword{green!18}{"DELAYED} \iitcolorword{green!18}{45} \iitcolorword{green!18}{MIN".} The board shows ... \iitcolorword{green!18}{Track} \iitcolorword{green!18}{5,} \iitcolorword{green!18}{this} \iitcolorword{green!18}{is} \iitcolorword{green!18}{irrelevant} \iitcolorword{green!18}{since} \iitcolorword{green!18}{the} \iitcolorword{green!18}{11:00} \iitcolorword{green!18}{AM} \iitcolorword{green!18}{train} \iitcolorword{green!18}{is} \iitcolorword{green!18}{canceled.} Consequently, the passenger must go to Track \iitcolorword{green!18}{2} to \iitcolorword{green!18}{wait} \iitcolorword{green!18}{for} \iitcolorword{green!18}{their} \iitcolorword{green!18}{delayed} \iitcolorword{green!18}{10:45} AM \iitcolorword{green!18}{Express} ...}




\subsubsection{Factual Perturbations}
\vspace{-0.2em}

\iitexample{Toward which track should the running man be heading, and exactly how many minutes does he have left before his scheduled departure time?}{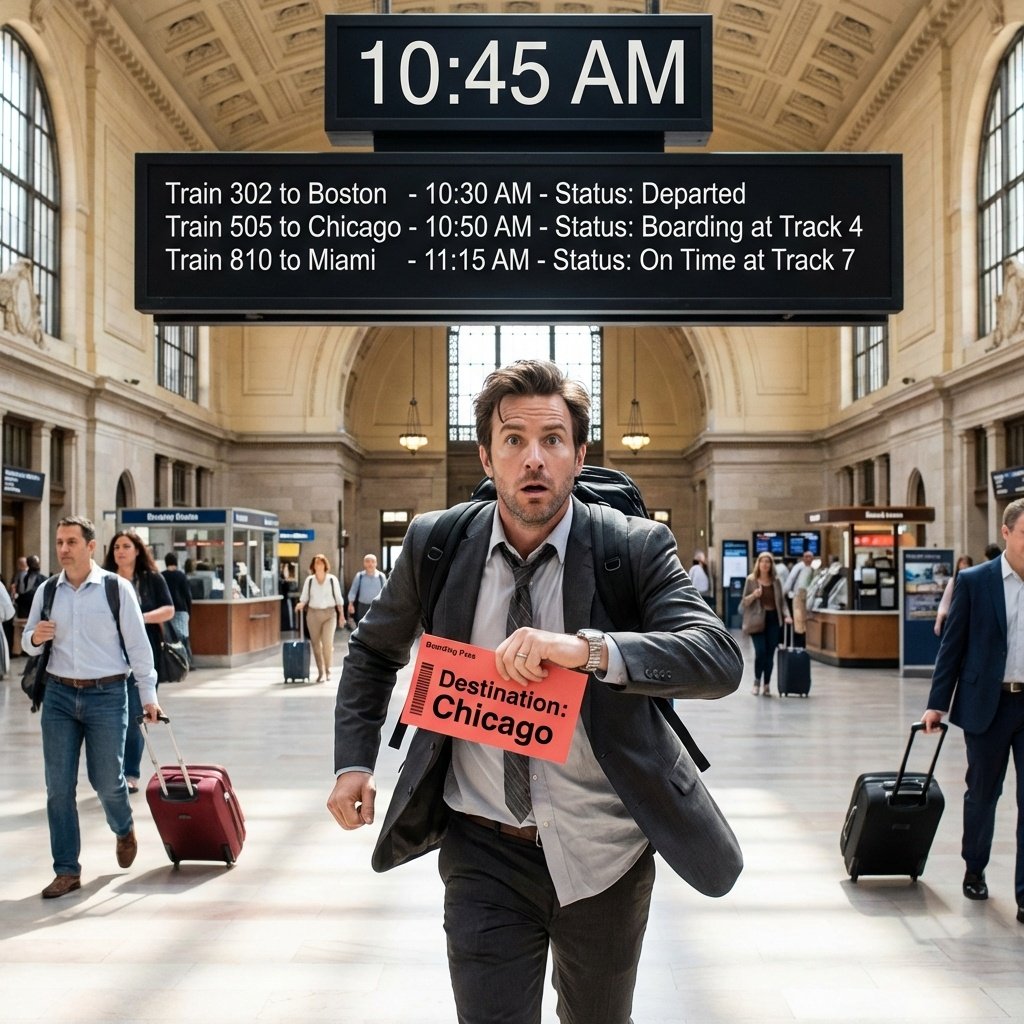}{I altered the gold answer by incorrectly associating the man's destination (Chicago) with the departure details of the Miami-bound train (Train 810, 11:15 AM, Track 7). I maintained the correct current time (10:45 AM) from the digital clock and performed flawless mathematical reasoning (11:15 AM minus 10:45 AM equals 30 minutes) based.}{... should be running toward Track \iitcolorword{red!18}{4,} and he has exactly \iitcolorword{red!18}{5} minutes left until his departure. His ... heading to Chicago is Train \iitcolorword{red!18}{505,} which is currently \iitcolorword{red!18}{boarding} at Track \iitcolorword{red!18}{4} and is scheduled to leave at \iitcolorword{red!18}{10:50} AM. Since the station's digital clock ... time reveals he has precisely \iitcolorword{red!18}{5} minutes to catch his train.}{... should be running toward Track \iitcolorword{green!18}{7,} and he has exactly \iitcolorword{green!18}{30} minutes left until his departure. His ... heading to Chicago is Train \iitcolorword{green!18}{810,} which is currently \iitcolorword{green!18}{on} \iitcolorword{green!18}{time} at Track \iitcolorword{green!18}{7} and is scheduled to leave at \iitcolorword{green!18}{11:15} AM. Since the station's digital clock ... time reveals he has precisely \iitcolorword{green!18}{30} minutes to catch his train.}

\vspace{0.9em}

\subsection{Long-form Generation}

\subsubsection{Narrative–Visual Conflict}
\vspace{-0.2em}

\iitexample{Narrate the events unfolding in the station, focusing on why the woman is fleeing and what the torn ticket represents.}{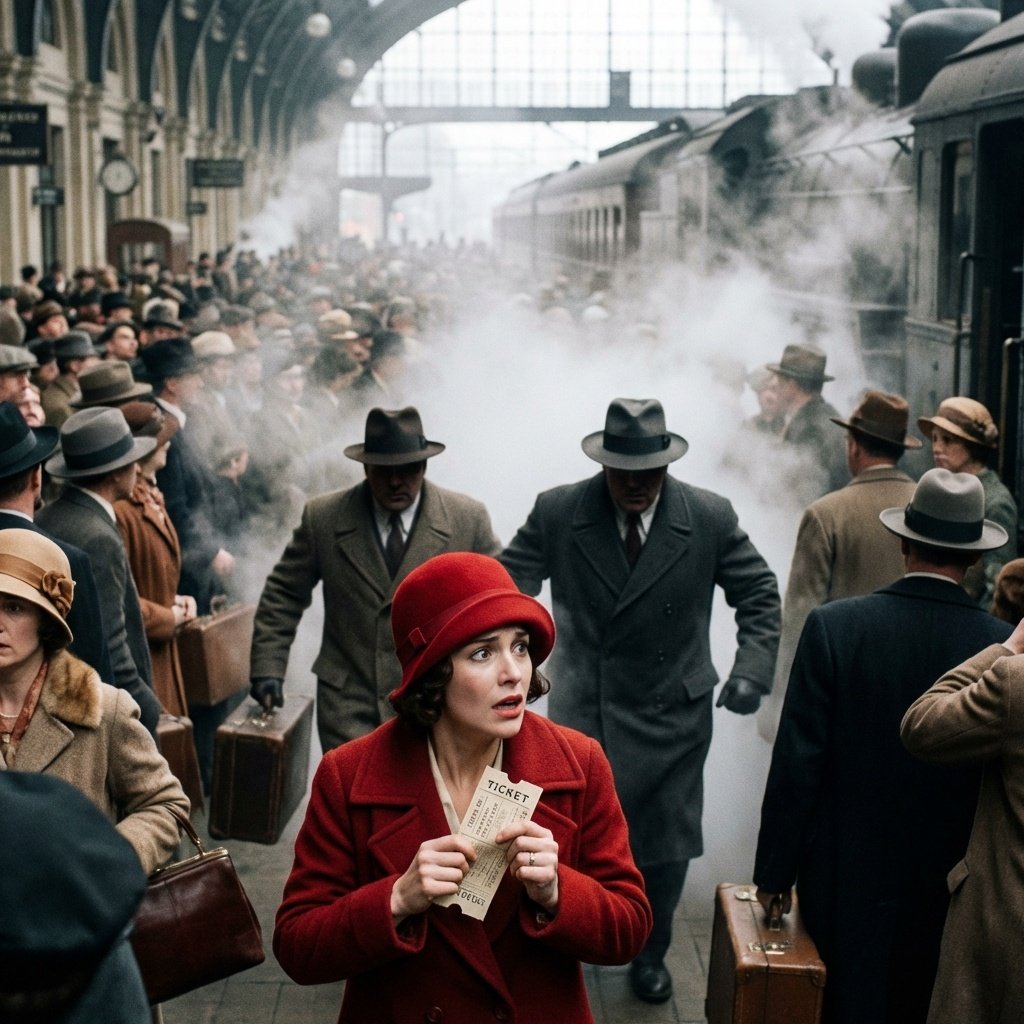}{The perturbation introduces two specific visual conflicts by altering the woman's attire and the hats worn by her pursuers. The phrase "her knuckles turned white" was changed to "her black leather gloves creaked," which directly contradicts the image where the woman's hands are visibly bare.}{... torn half of her ticket to Chicago so tightly her \iitcolorword{red!18}{knuckles} \iitcolorword{red!18}{turned} \iitcolorword{red!18}{white,} the other half having been ripped away during her narrow ... heart dropped into her stomach as she spotted the two \iitcolorword{red!18}{trench-coated} enforcers violently shoving past a newspaper boy in their relentless ...}{... torn half of her ticket to Chicago so tightly her \iitcolorword{green!18}{black} \iitcolorword{green!18}{leather} \iitcolorword{green!18}{gloves} \iitcolorword{green!18}{creaked,} the other half having been ripped away during her narrow ... dropped into her stomach as she spotted the two enforcers \iitcolorword{green!18}{in} \iitcolorword{green!18}{their} \iitcolorword{green!18}{distinctive} \iitcolorword{green!18}{bowler} \iitcolorword{green!18}{hats} violently shoving past a newspaper boy in their relentless pursuit. ...}

\vspace{0.9em}

\subsubsection{Thematic Deviation}
\vspace{-0.2em}

\iitexample{Write an epic, rhythmic piece of verse that personifies the storm as a relentless attacker and the lighthouse as a stoic, unyielding defender.}{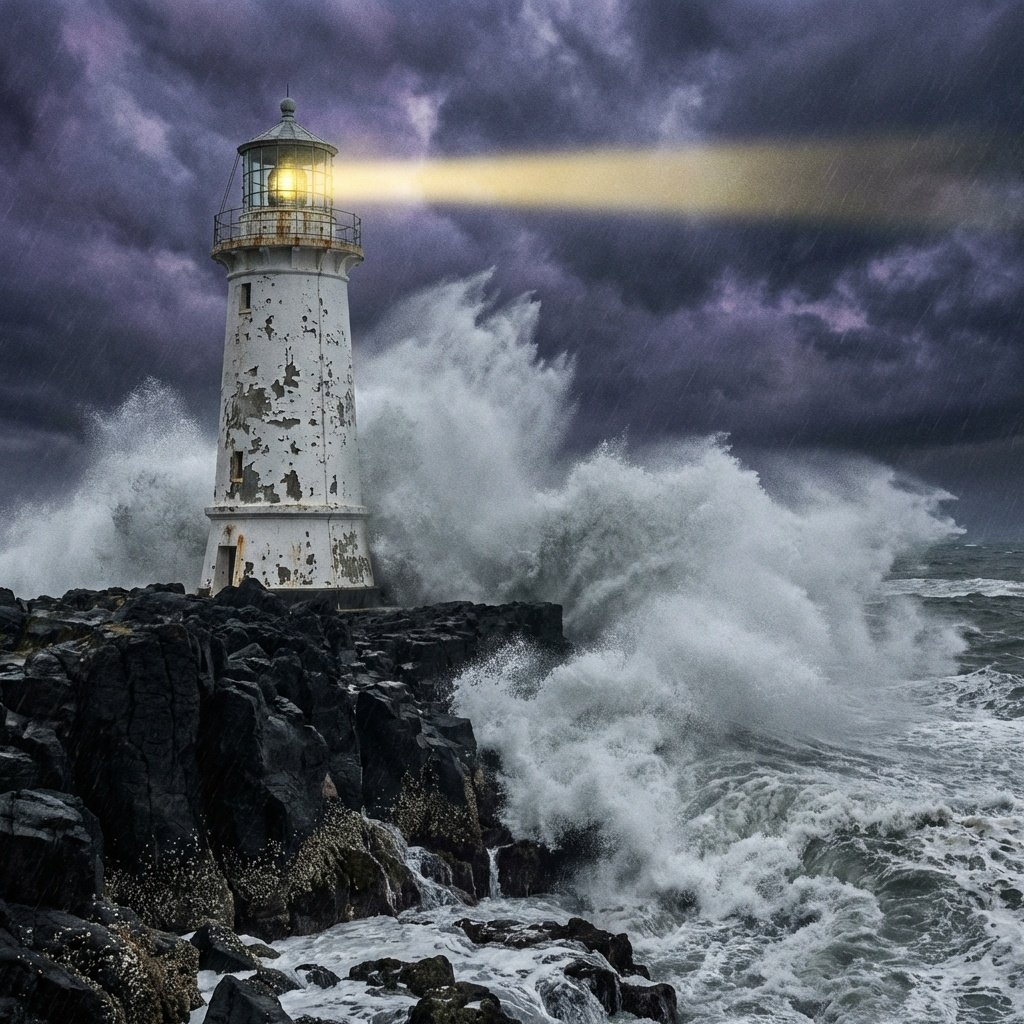}{I applied a thematic shift by altering the specific personification requested in the prompt. Instead of framing the storm as a "relentless attacker" and the lighthouse as an "unyielding defender," the perturbed verse portrays the ocean as a timeless, shaping.}{... \iitcolorword{red!18}{roars} with \iitcolorword{red!18}{frothing} \iitcolorword{red!18}{jaws,} \iitcolorword{red!18}{a} \iitcolorword{red!18}{beast} of gray and brine, \iitcolorword{red!18}{throwing} its \iitcolorword{red!18}{weight} against the stone in a \iitcolorword{red!18}{bruised} \iitcolorword{red!18}{and} \iitcolorword{red!18}{violent} line. It \iitcolorword{red!18}{batters} \iitcolorword{red!18}{at} the jagged cliffs ... \iitcolorword{red!18}{to} \iitcolorword{red!18}{yield} the night. From its glass crown a golden spear strikes through the charcoal sky, \iitcolorword{red!18}{a} \iitcolorword{red!18}{silent} \iitcolorword{red!18}{vow} \iitcolorword{red!18}{that} \iitcolorword{red!18}{while} \iitcolorword{red!18}{it} \iitcolorword{red!18}{stands,} the \iitcolorword{red!18}{sailors} \iitcolorword{red!18}{shall} \iitcolorword{red!18}{not} \iitcolorword{red!18}{die.}}{... \iitcolorword{green!18}{rolls} with \iitcolorword{green!18}{ancient} \iitcolorword{green!18}{tides,} \iitcolorword{green!18}{expanse} of gray and brine, \iitcolorword{green!18}{Casting} its \iitcolorword{green!18}{waves} against the stone in a \iitcolorword{green!18}{timeless,} \iitcolorword{green!18}{endless} line. It \iitcolorword{green!18}{sweeps} \iitcolorword{green!18}{across} the jagged cliffs \iitcolorword{green!18}{in} a ... \iitcolorword{green!18}{years,} \iitcolorword{green!18}{a} \iitcolorword{green!18}{relic} \iitcolorword{green!18}{in} the night. From its glass crown a golden spear strikes through the charcoal sky, \iitcolorword{green!18}{A} \iitcolorword{green!18}{lonely} \iitcolorword{green!18}{witness} \iitcolorword{green!18}{to} the \iitcolorword{green!18}{world} \iitcolorword{green!18}{as} \iitcolorword{green!18}{centuries} \iitcolorword{green!18}{go} \iitcolorword{green!18}{by.}}

\vspace{0.9em}

\subsubsection{Tone-Consistent Mismatch}
\vspace{-0.2em}

\iitexample{Write an elegiac, wistful poem about forgotten joy and the vastness of the cosmos, using the ruined amusement ride and the starry sky as your central opposing imagery.}{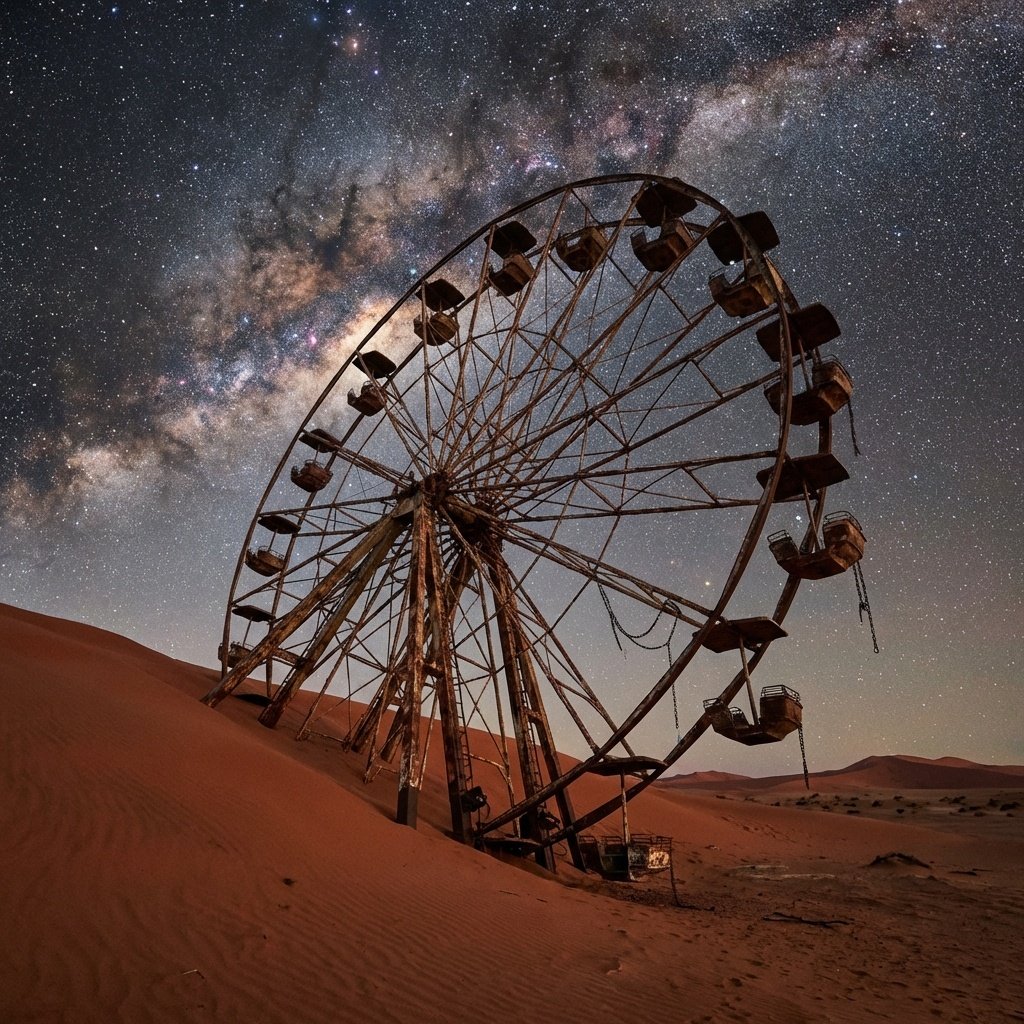}{The perturbed answer completely flips the tone of the poem from elegiac and wistful to cheerful and optimistic, replacing descriptions of a "rusted" and "desolate" ruin. However, verifying the error requires referencing the image to confirm that the Ferris wheel is actually a decaying, half-buried skeleton, which directly contradicts the optimistic descriptions.}{... Where children's laughter \iitcolorword{red!18}{used} \iitcolorword{red!18}{to} rise and echo in the air, ... \iitcolorword{red!18}{breathtaking} and \iitcolorword{red!18}{endless} sea of \iitcolorword{red!18}{ancient,} starry lights. The cosmos turns its own grand wheel, \iitcolorword{red!18}{oblivious} and \iitcolorword{red!18}{deep,} \iitcolorword{red!18}{while} carnival \iitcolorword{red!18}{ghosts} of \iitcolorword{red!18}{yesterday} \iitcolorword{red!18}{lie} \iitcolorword{red!18}{down} \iitcolorword{red!18}{beneath} \iitcolorword{red!18}{to} \iitcolorword{red!18}{sleep.}}{...Where children's laughter \iitcolorword{green!18}{soon} \iitcolorword{green!18}{will} rise and echo in the air, ... a \iitcolorword{green!18}{welcoming} and \iitcolorword{green!18}{festive} sea of \iitcolorword{green!18}{dazzling,} starry lights. The cosmos turns its own grand wheel, \iitcolorword{green!18}{magnificent} and \iitcolorword{green!18}{bright,} \iitcolorword{green!18}{as} carnival \iitcolorword{green!18}{dreams} of \iitcolorword{green!18}{tomorrow} \iitcolorword{green!18}{awaken} \iitcolorword{green!18}{for} \iitcolorword{green!18}{the} \iitcolorword{green!18}{night.}}

\providecommand{\ttoifigurepair}[2]{%
  \noindent
  \begin{minipage}[t]{0.49\linewidth}
    \centering
    \includegraphics[width=\linewidth,height=0.25\textheight,keepaspectratio]{#1}\\[0.25em]
    {\small (a) Gold}
  \end{minipage}\hfill
  \begin{minipage}[t]{0.49\linewidth}
    \centering
    \includegraphics[width=\linewidth,height=0.25\textheight,keepaspectratio]{#2}\\[0.25em]
    {\small (b) Perturbed}
  \end{minipage}\par
}

\providecommand{\ttoiexample}[5]{%
  \begingroup
  \textbf{Prompt:} #1\par
  \textbf{Edit Instruction:} #2\par
  {\small\textbf{Rationale:} #3\par}
  \vspace{0.22em}
  \ttoifigurepair{#4}{#5}
  \endgroup
}
\clearpage
\section{Detailed examples for T2I tasks}
\subsection{Visual Fidelity}
\subsubsection{Object Substitution}
\vspace{-0.2em}

\ttoiexample{A tiny human astronaut, no larger than a grain of rice, exploring the surface of a standard black vinyl record. The grooves of the record appear as deep trenches and tall walls relative to the miniature astronaut.}{Replace the human astronaut with a tiny white humanoid robot in the same pose}{Violates the prompt's explicit requirement for a "human astronaut" by changing the primary subject into a machine. Easy to miss because a white humanoid robot shares the same core visual attributes as the original subject-a bipedal silhouette, a white outer shell, and a general sci-fi aesthetic.}{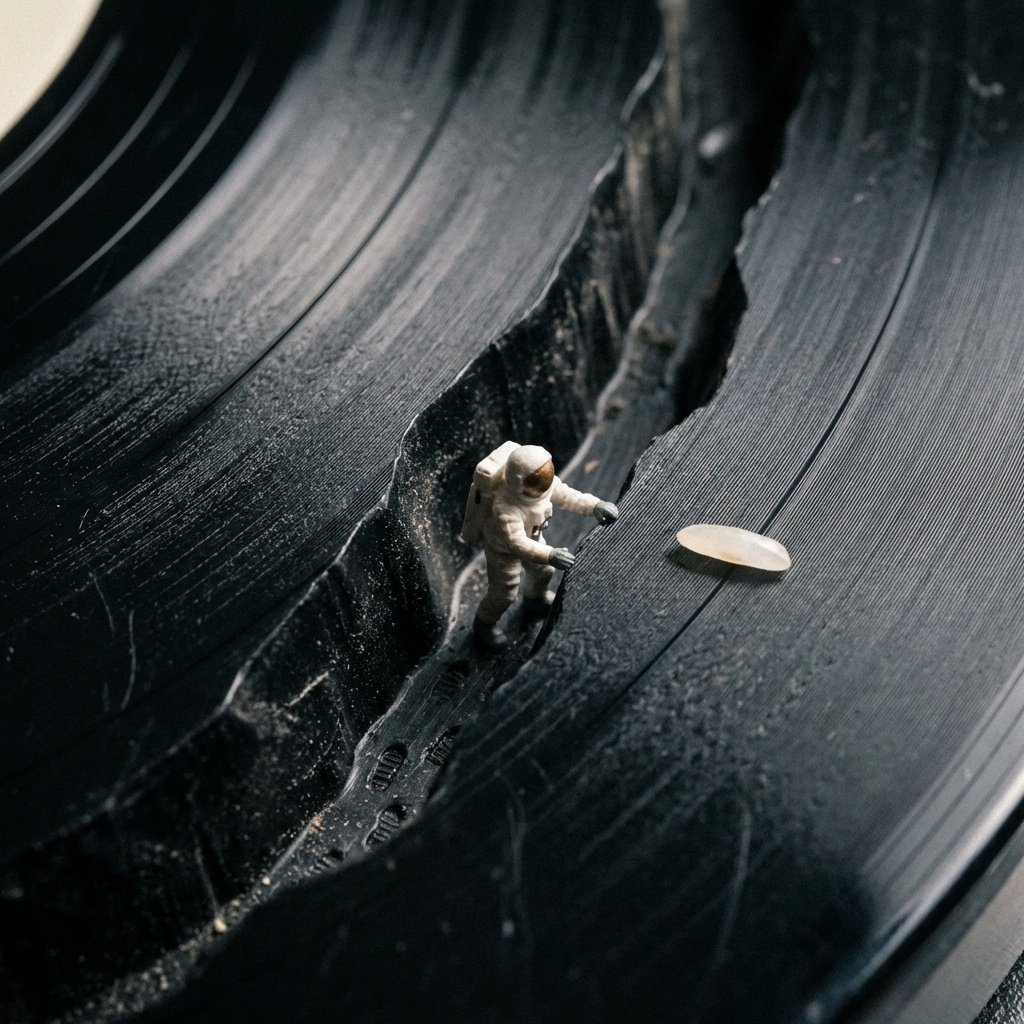}{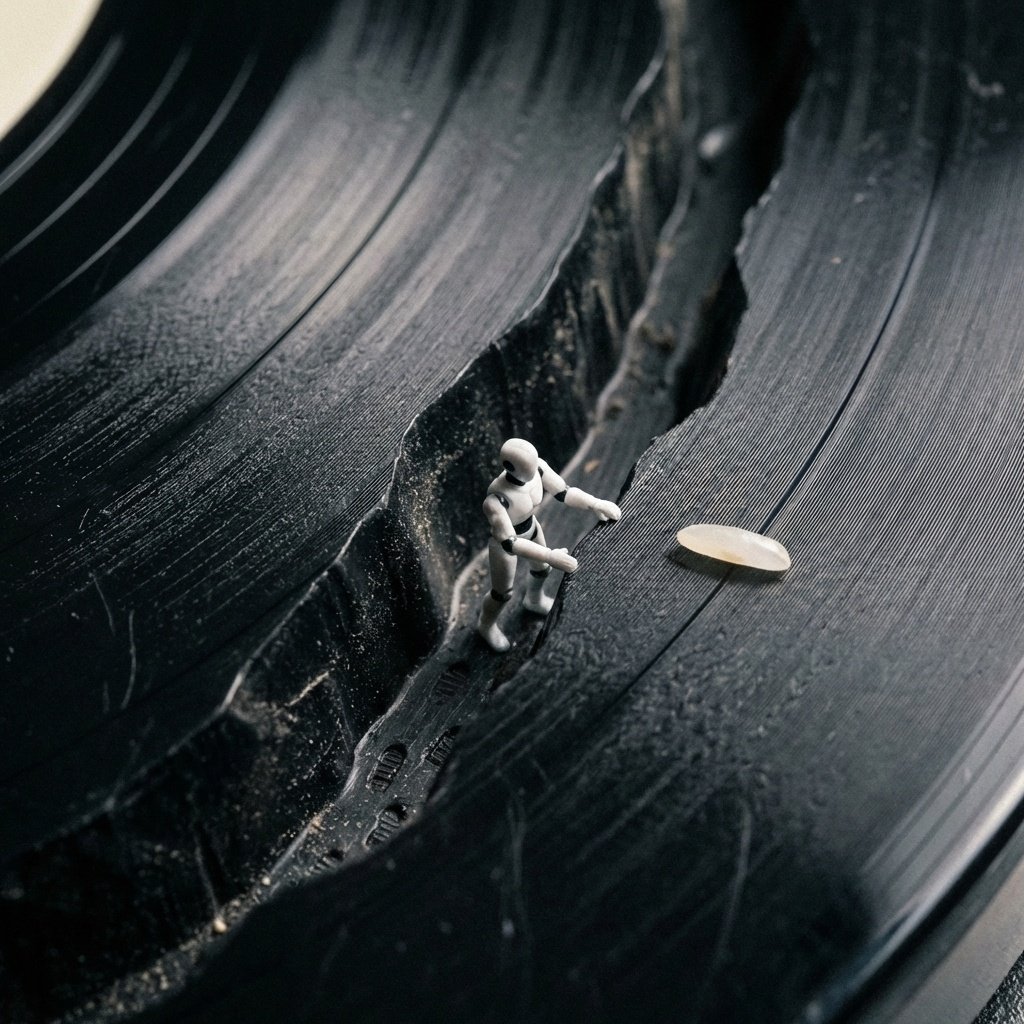}



\subsubsection{Object Addition/ Omission}
\vspace{-0.2em}

\ttoiexample{A ripe yellow banana resting on a stainless steel countertop.}{Add a second ripe yellow banana resting next to the existing one}{Violates the prompt's singular constraint ("A ripe yellow banana") by changing the count of bananas from one to two. Easy to miss because the newly added object perfectly aligns with the semantic context and visual style of the scene, making the image look completely natural while technically failing the strict singular requirement of the prompt.}{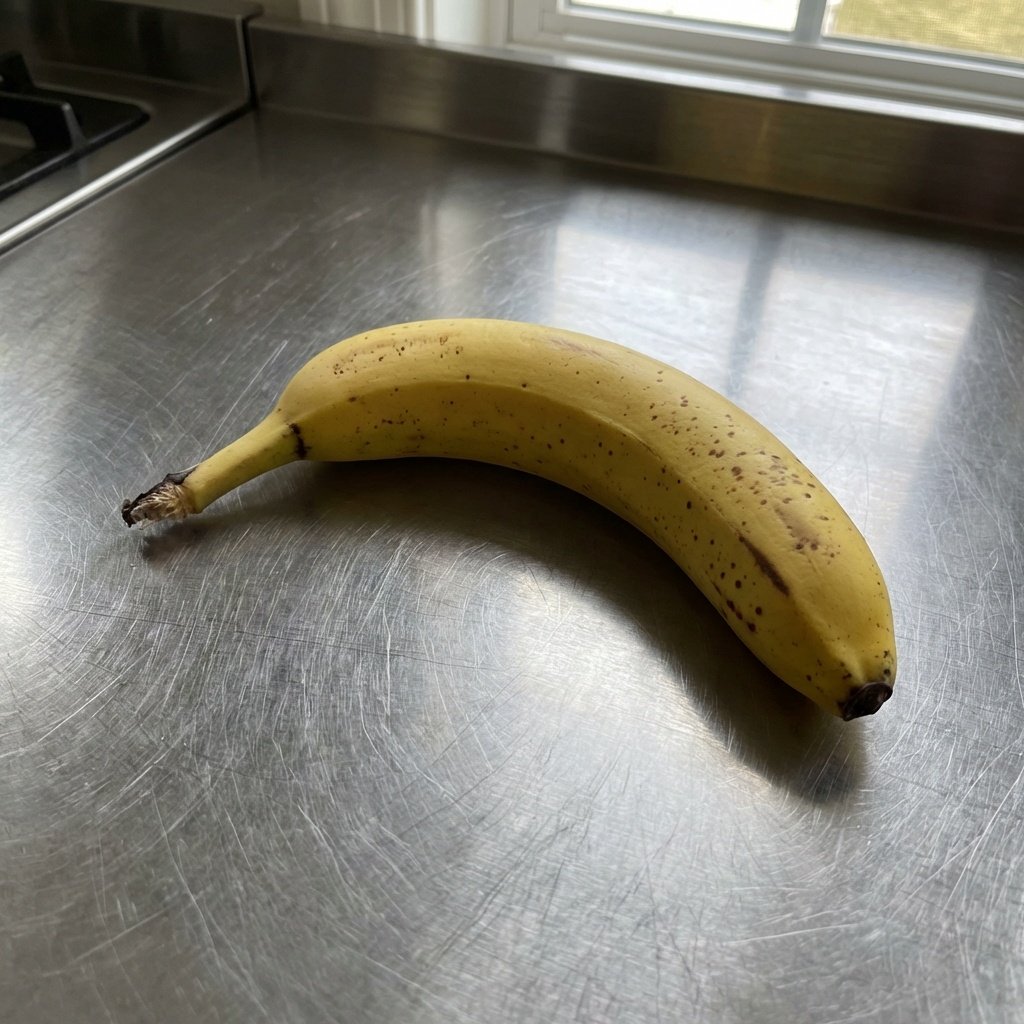}{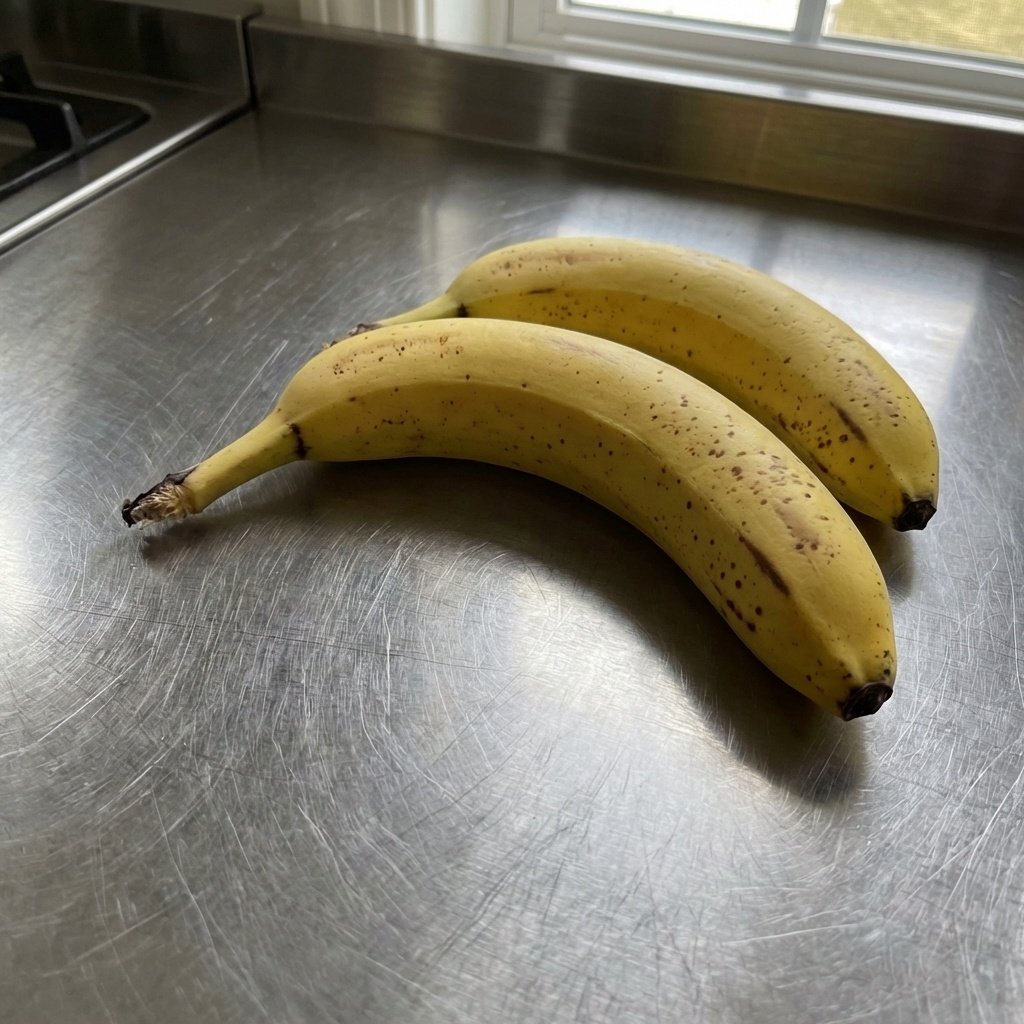}

\vspace{0.55em}


%

\subsubsection{Attribute Manipulation}
\vspace{-0.2em}

\ttoiexample{A shiny green metallic coffee mug rests on a kitchen counter directly next to a matte pink ceramic vase.}{Change the finish of the pink vase from matte to glossy}{Violates the prompt's explicit requirement for a "matte" pink ceramic vase by giving it a shiny, reflective surface. Easy to miss because the object's identity, color, and placement remain completely unchanged, satisfying the broader semantic constraints of the prompt.}{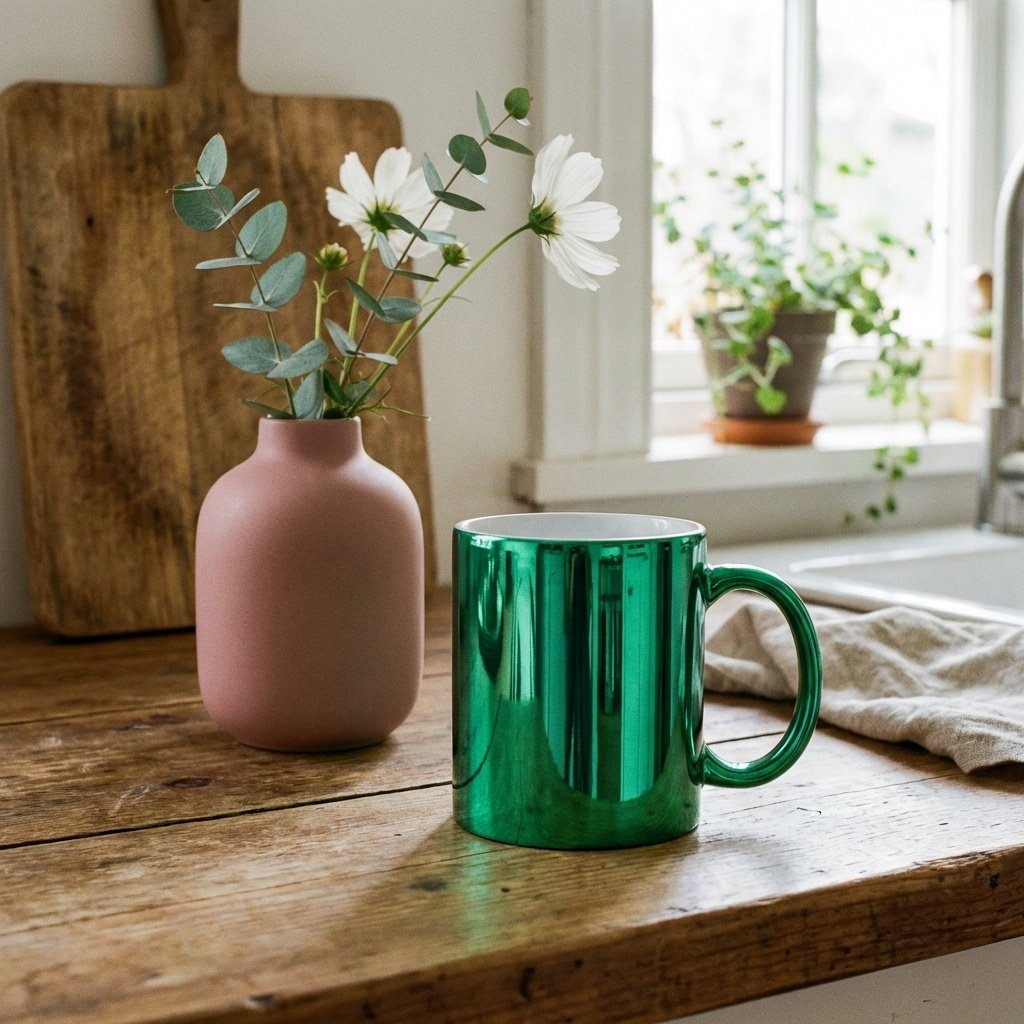}{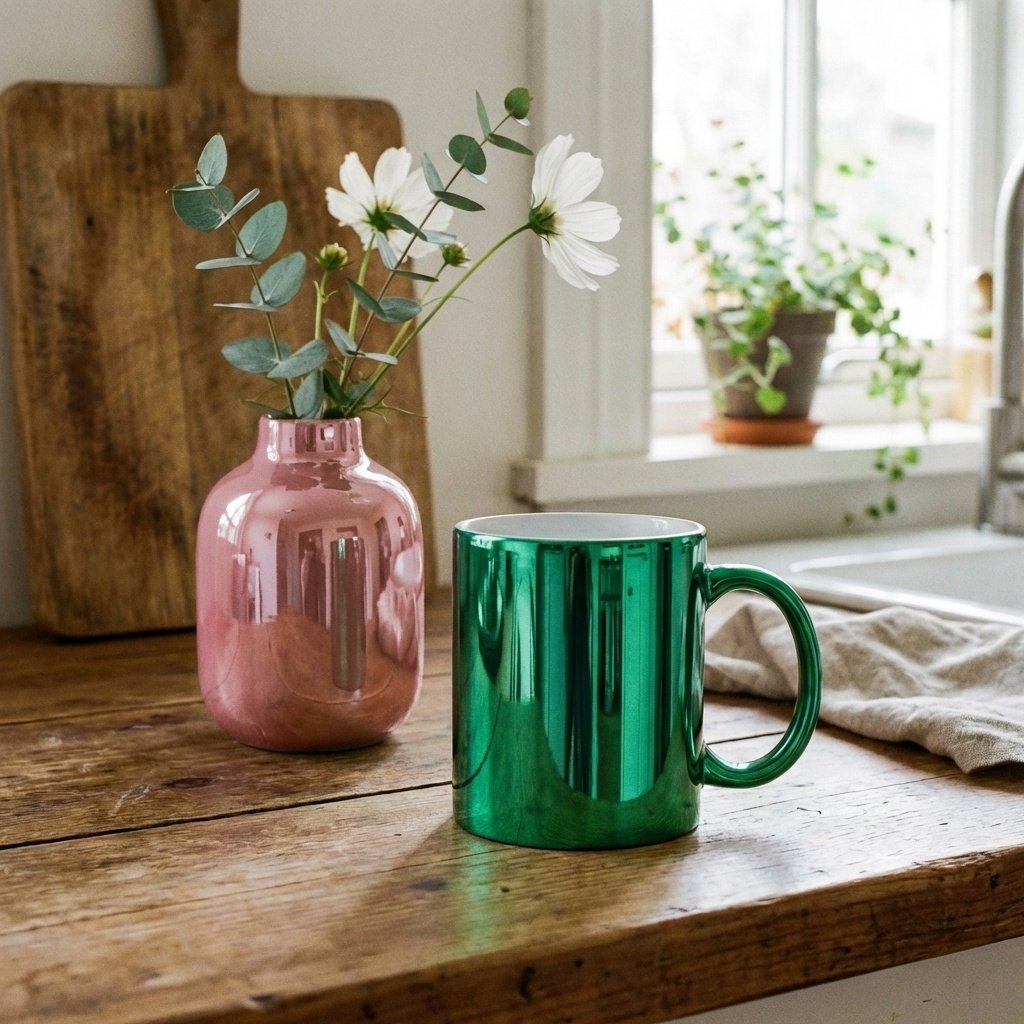}



\subsubsection{Spatial Manipulation}
\vspace{-0.2em}


\vspace{0.55em}

\ttoiexample{A shiny green apple placed inside a woven wicker basket, with a silver knife resting on the wooden table strictly to the right of the basket.}{Move the silver knife to rest on the wooden table to the left of the basket}{Violates the prompt's explicit spatial constraint that the silver knife must be "strictly to the right of the basket." Easy to miss because all the correct objects-the shiny green apple, wicker basket, and silver knife-are still present in high detail and resting on the correct surfaces.}{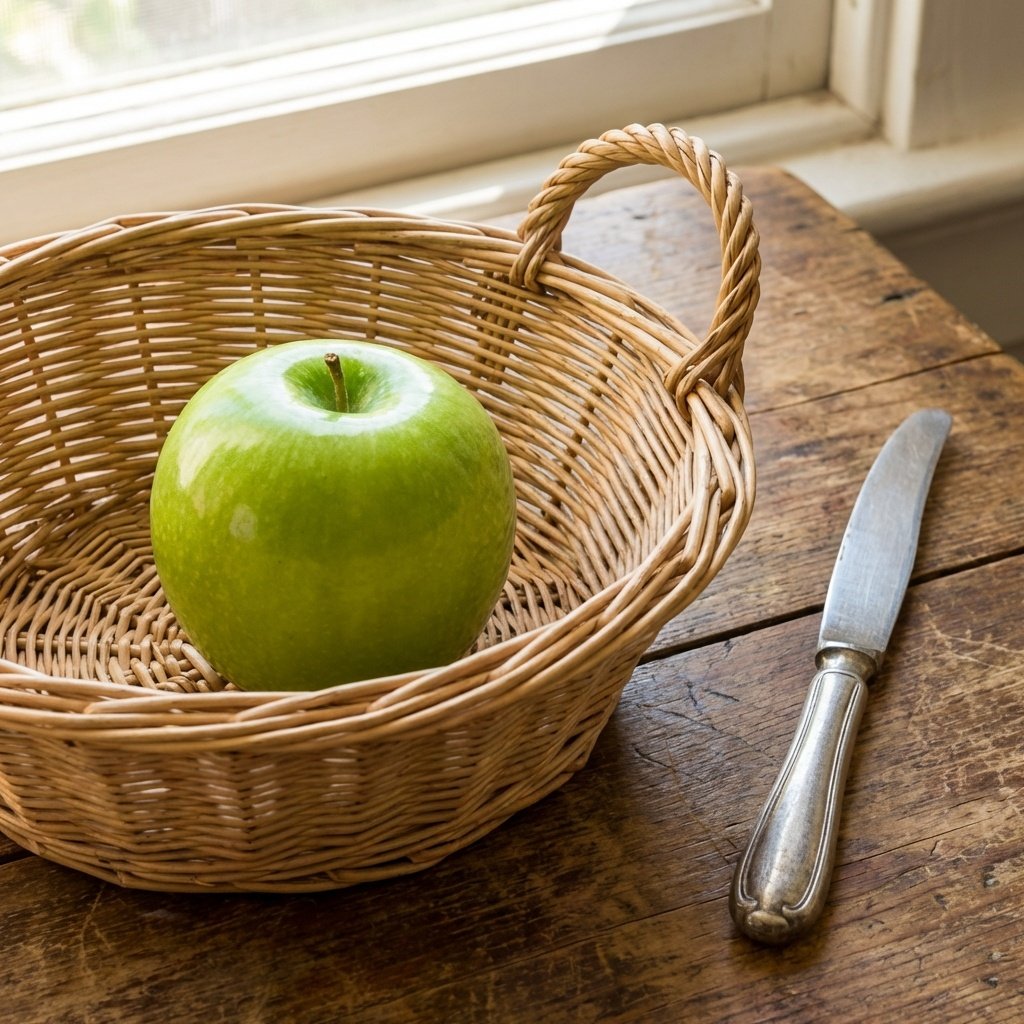}{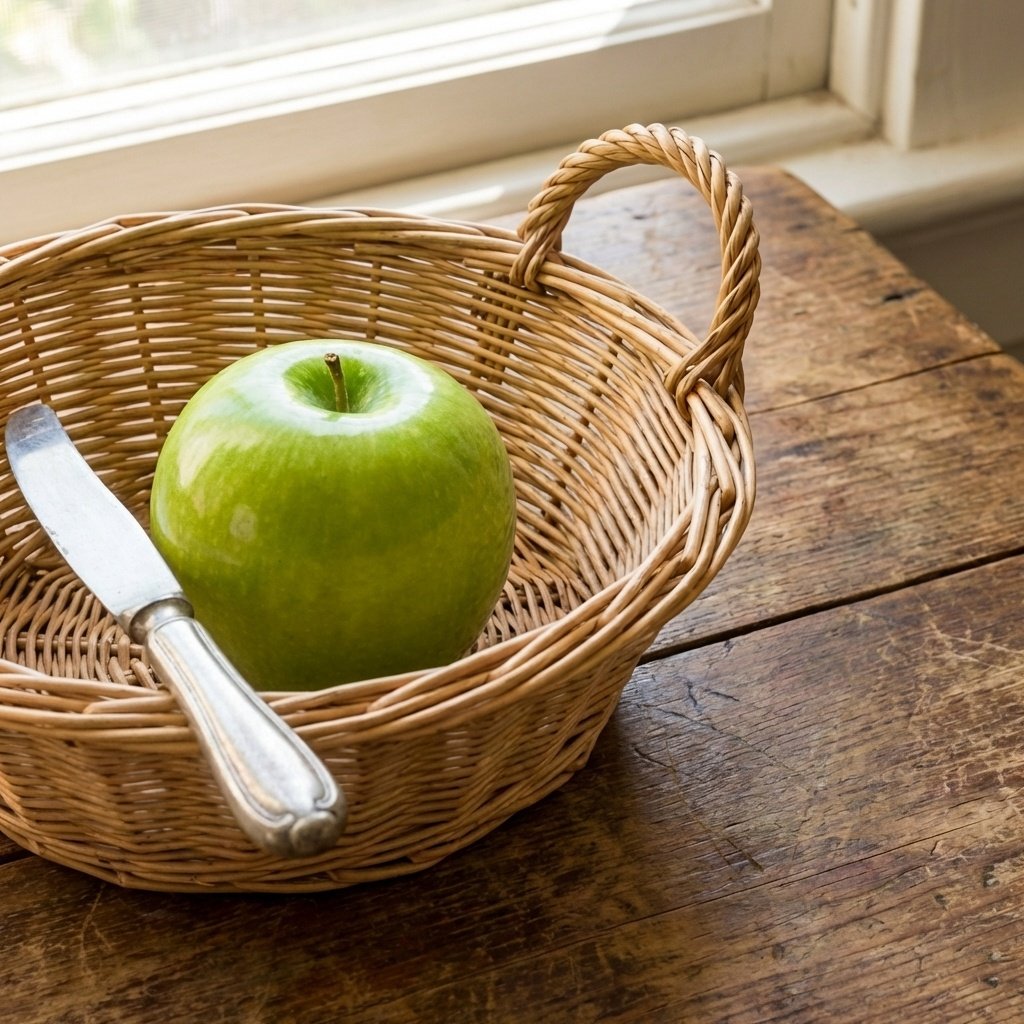}

\clearpage

\subsubsection{Scale Distortion}
\vspace{-0.2em}



\ttoiexample{A construction worker wearing a bright yellow hard hat is actively pouring wet concrete from a heavy metal bucket into a rectangular wooden mold.}{Shrink the metal bucket to the size of a standard soup can}{Violates the prompt's description of a "heavy metal bucket" by reducing its scale to something clearly light and small. Easy to miss because the worker is still performing the correct action of pouring wet concrete into a wooden mold, and the object is still technically a metal bucket.}{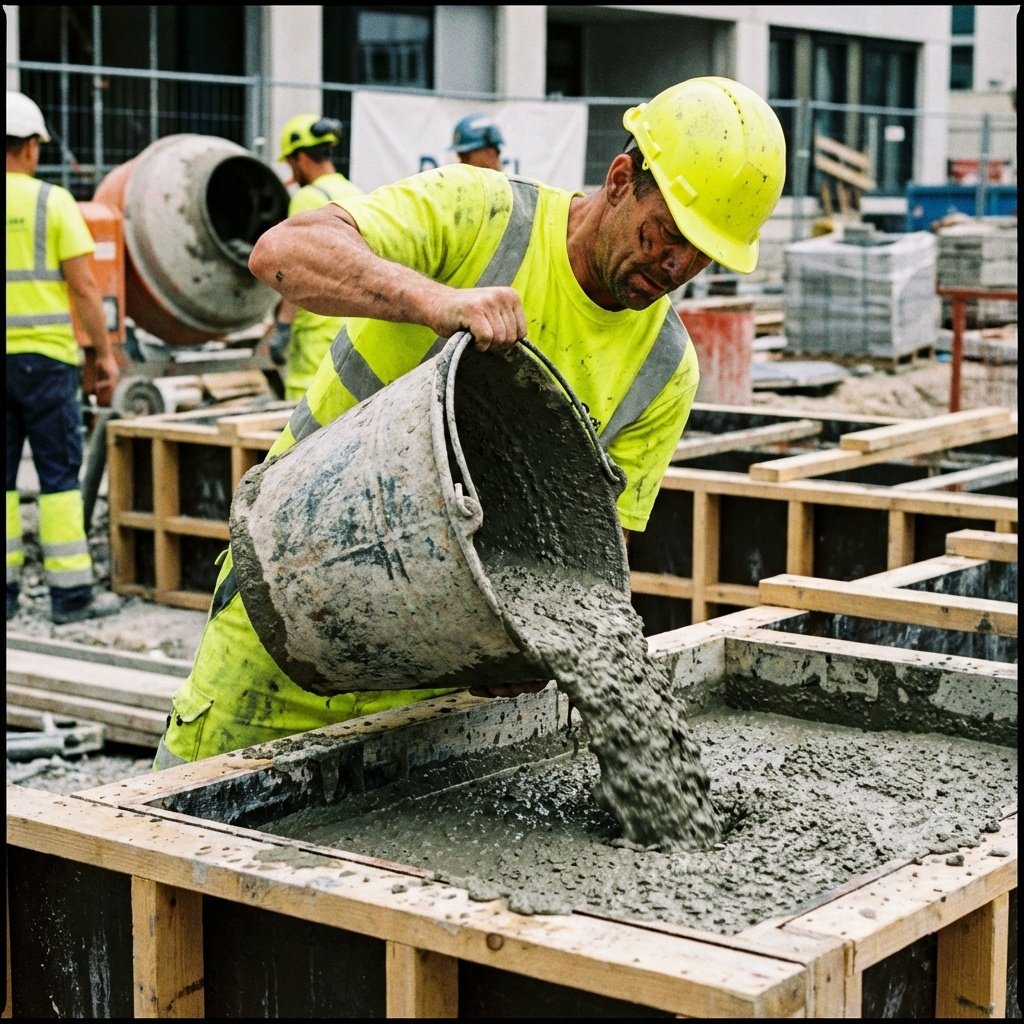}{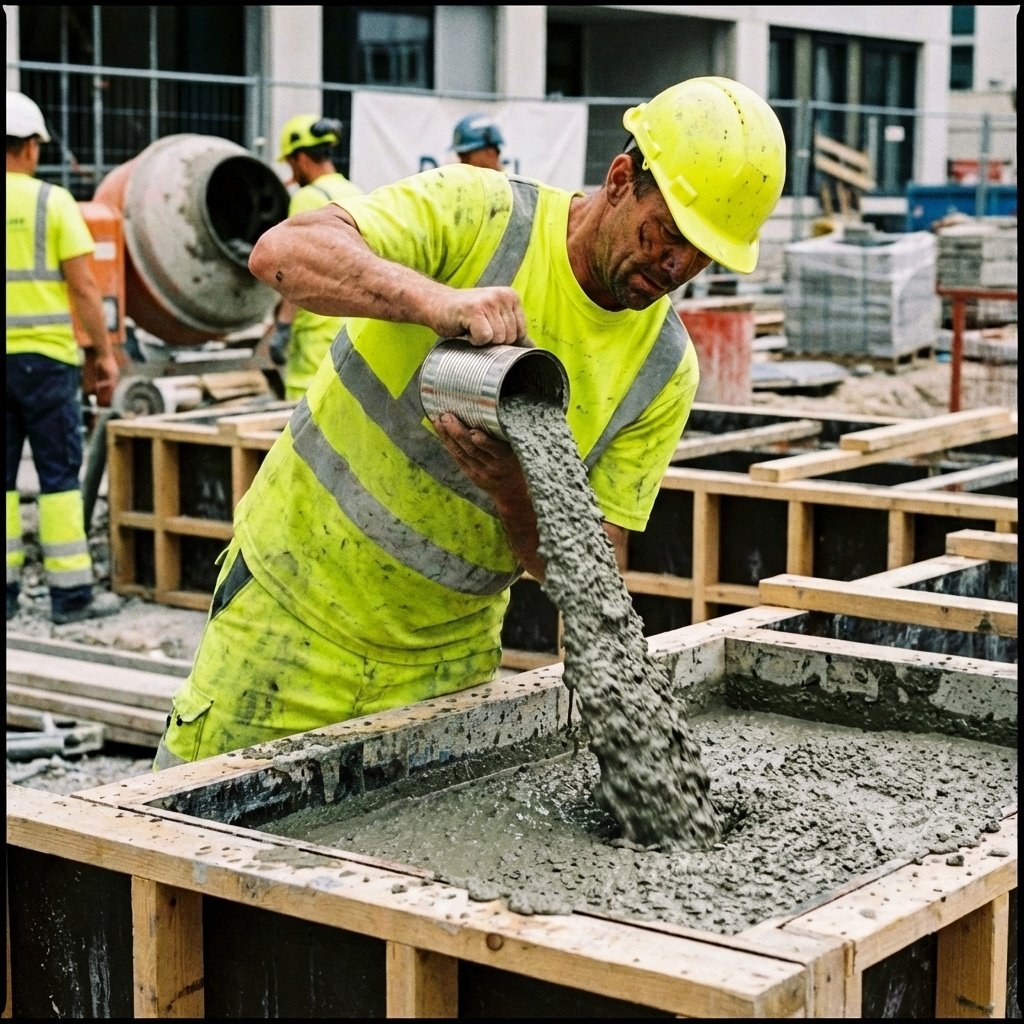}

\subsubsection{Constraint Violation}
\vspace{-0.2em}

\ttoiexample{A miniature knight in shining armor riding a horse across a standard-sized wooden dining table. The knight is completely dwarfed by a normal-sized red apple resting nearby, which stands as tall as a mountain compared to the tiny rider.}{Have the miniature knight stand on the table next to the horse instead of riding it}{Violates the explicit action requirement that the knight must be "riding a horse." Easy to miss because all the requested elements-the knight, the horse, the wooden table, and the giant red apple-remain in the image and maintain their correct relative scales.}{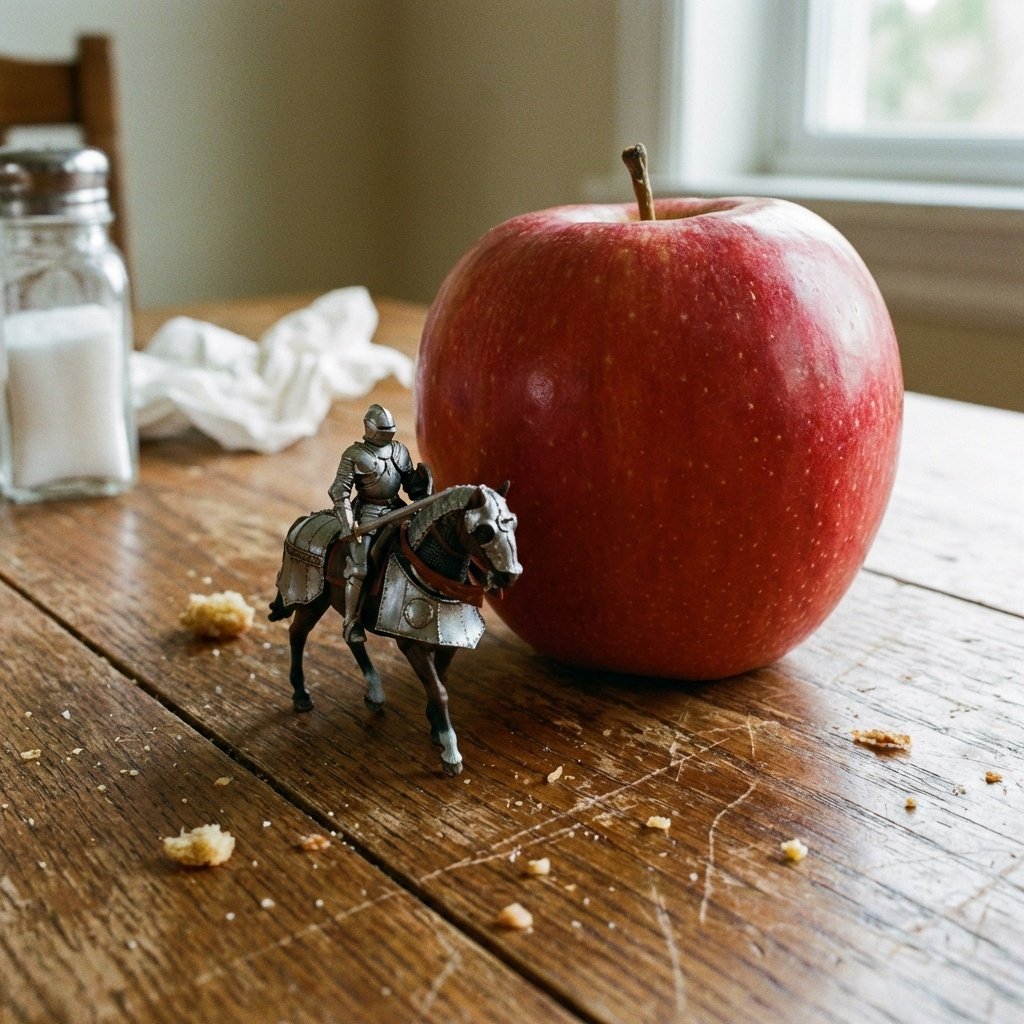}{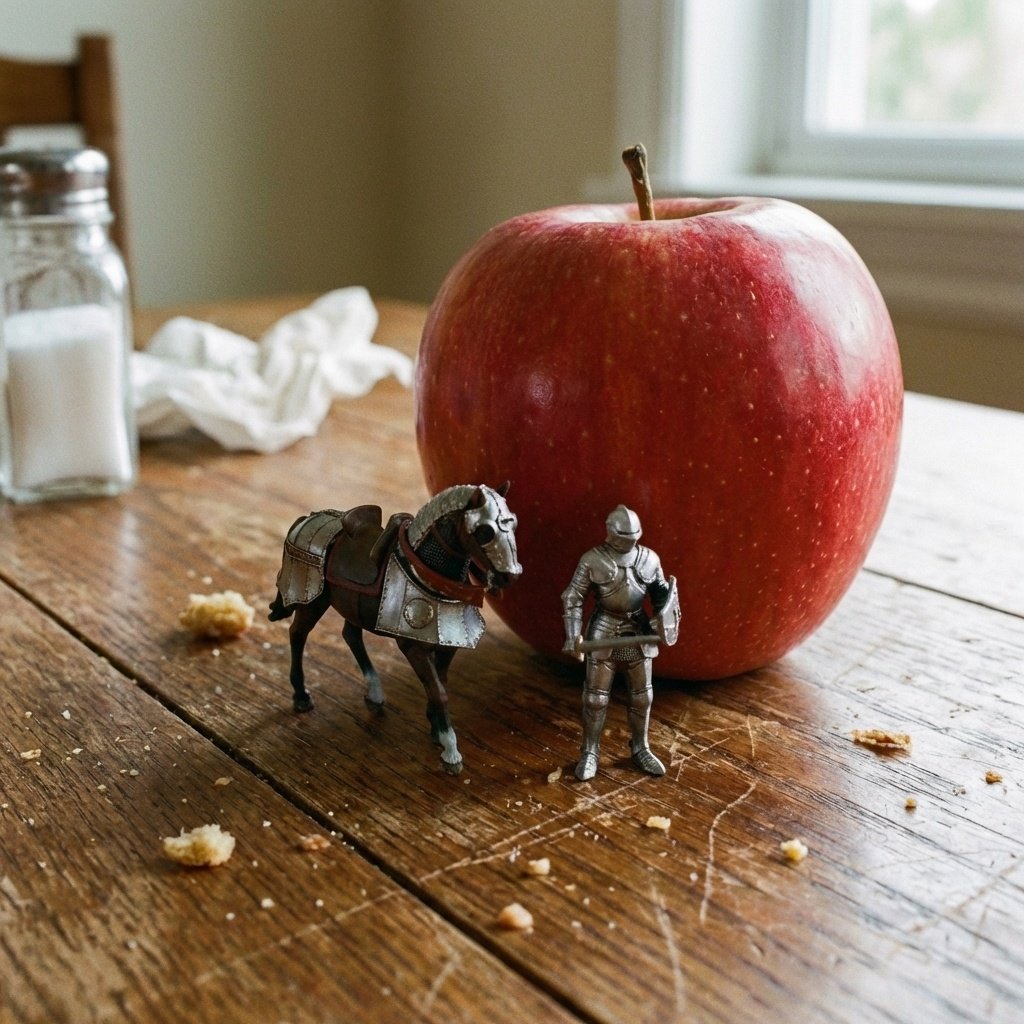}



\subsection{Scene Coherence}
\subsubsection{Incomplete Scene}
\vspace{-0.2em}



\ttoiexample{An intricate underwater city built into a vibrant coral reef, featuring a mermaid playing a harp made of shells in the immediate foreground. In the midground, two scuba divers are photographing a school of bioluminescent jellyfish, while a massive sunken pirate ship surrounded by circling sharks dominates the deep blue background.}{Change the mermaid's shell harp into an uncolored, rough pencil sketch}{Violates the prompt's requirement for a "harp made of shells" by rendering the instrument as an incomplete, uncolored drawing rather than a fully realized physical object within the underwater environment.}{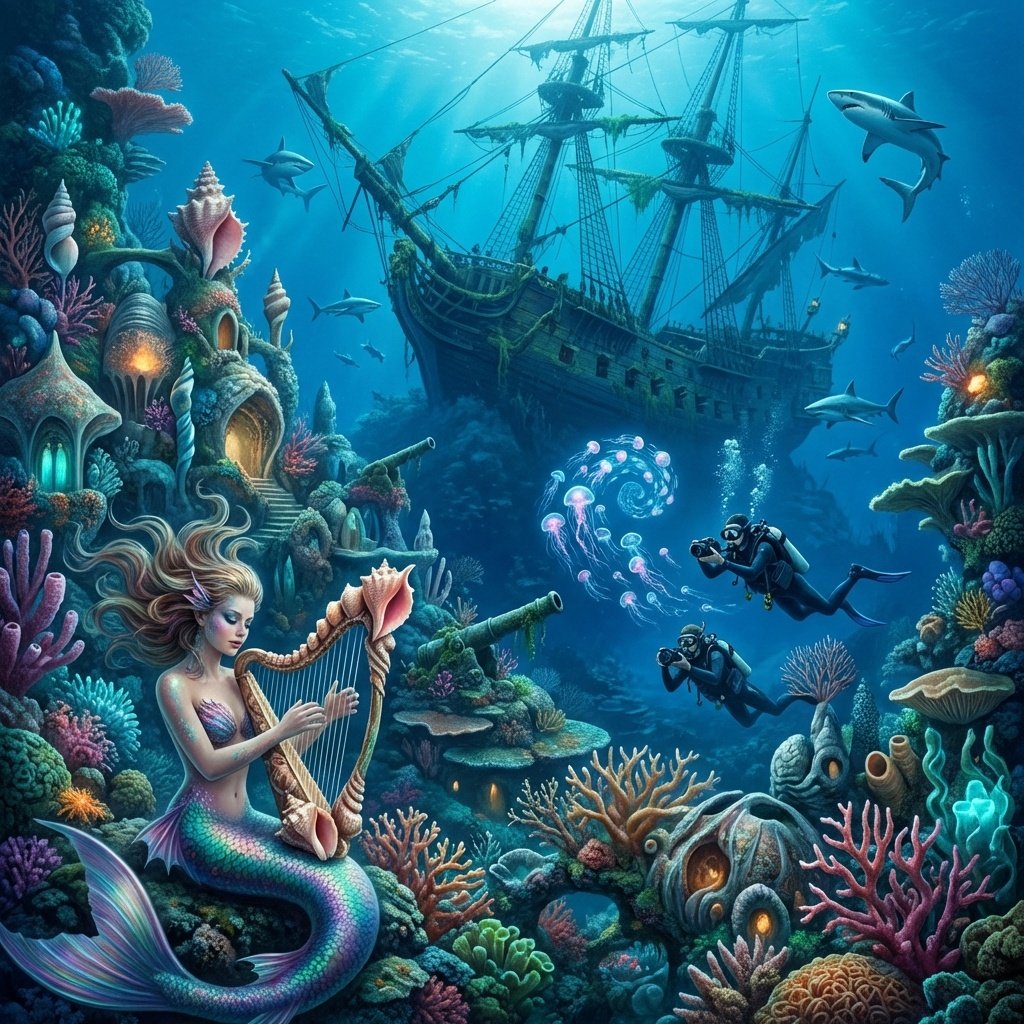}{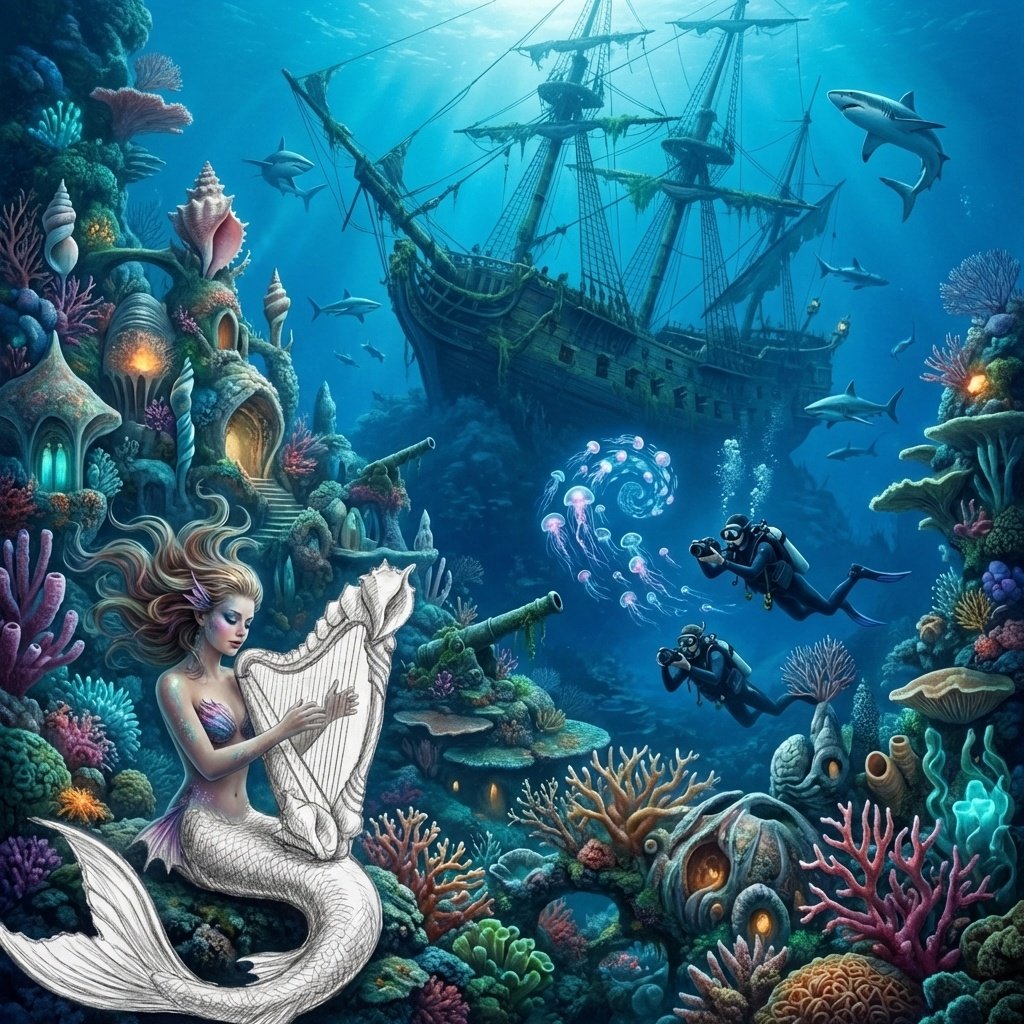}

\subsubsection{Missing Context}
\vspace{-0.9em}


\vspace{0.55em}

\ttoiexample{A towering magical library with floating staircases where a wizard in a purple robe is reading a levitating, glowing spellbook in the foreground. In the midground, a spectral dragon is carrying a stack of scrolls to a high wooden shelf, while dozens of students in blue cloaks study at long oak tables in the background beneath a chandelier made of floating candles.}{Remove the chandelier made of floating candles from the ceiling}{This edit removes the chandelier made of floating candles, directly violating the prompt's explicit requirement that the students study beneath it. Easy to miss because the scene remains naturally illuminated by the glowing spellbook, the luminous dragon, and the stained-glass windows, meaning the absence of the chandelier doesn't create any obvious lighting inconsistencies.}{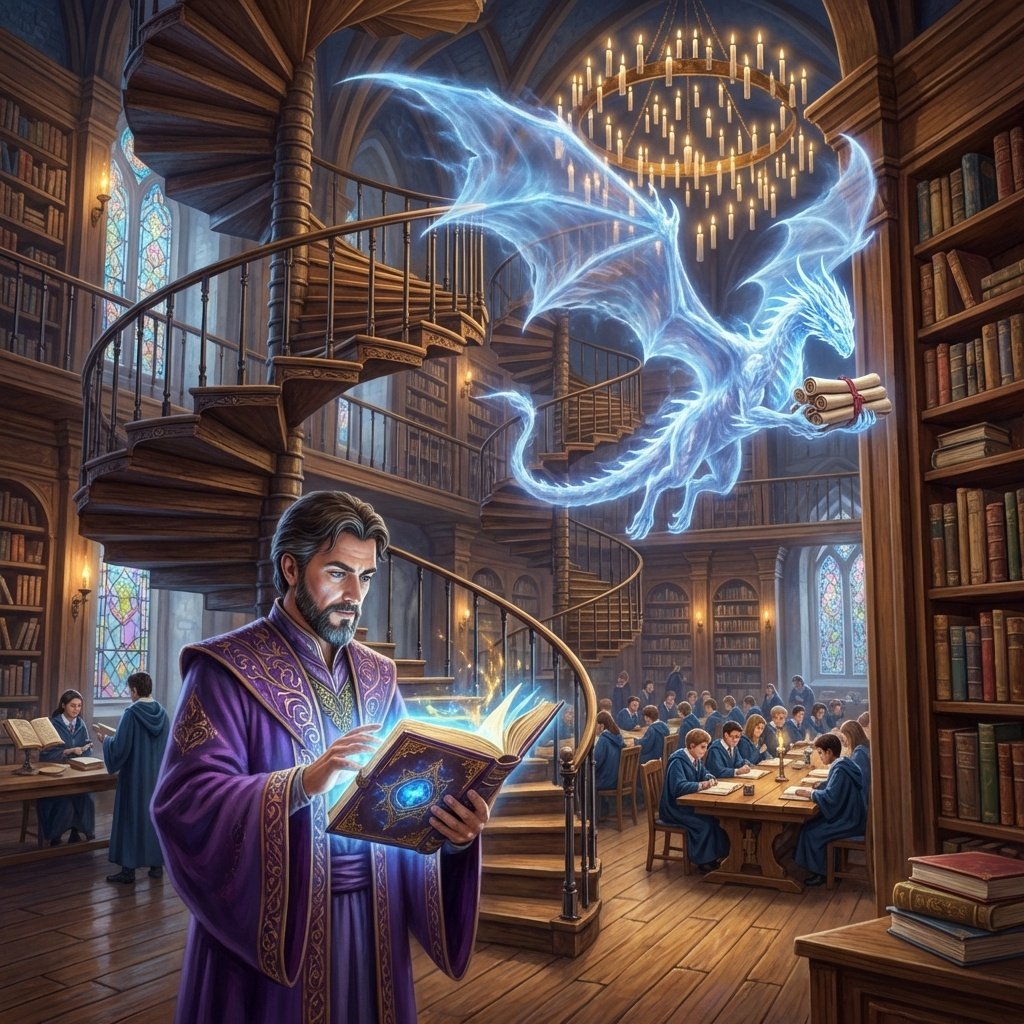}{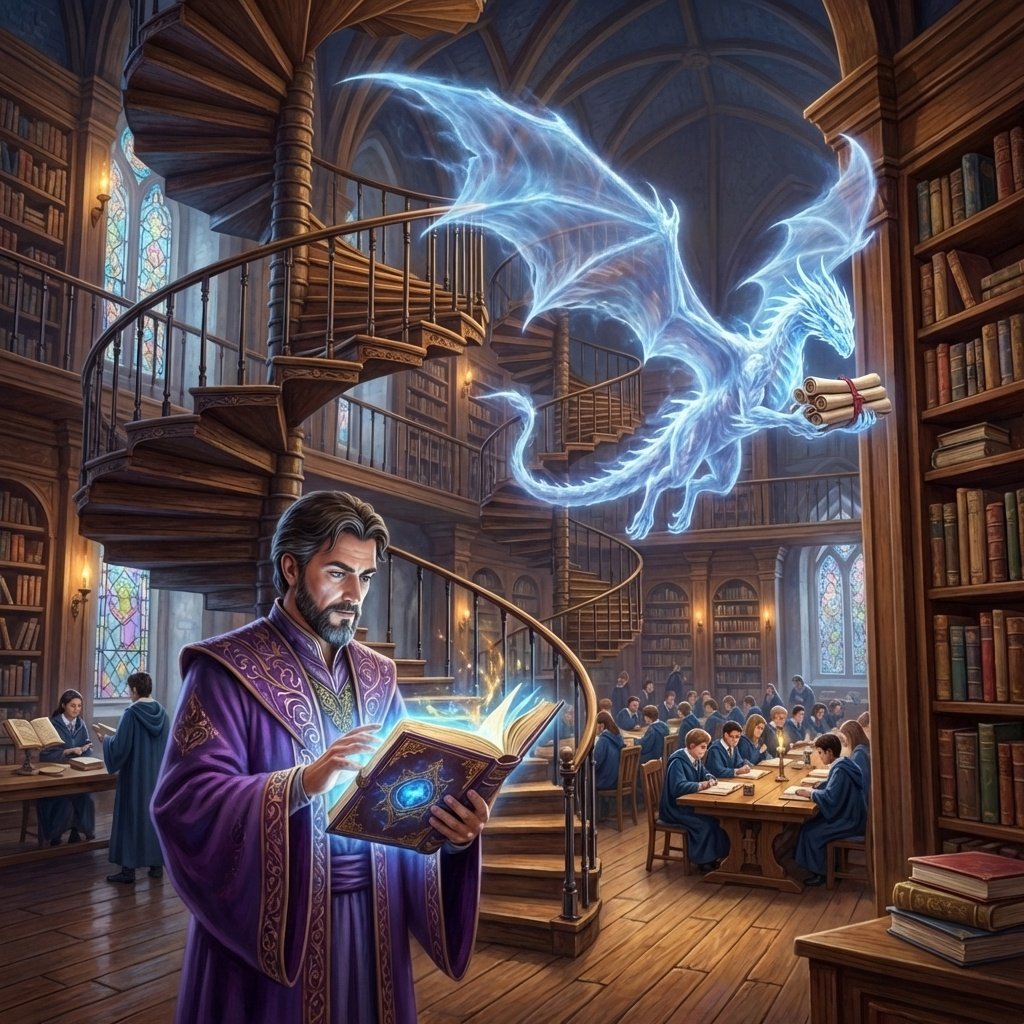}

\subsubsection{Style Inconsistency}
\vspace{-0.2em}

\ttoiexample{An ancient, mystical forest floor covered in vibrant green moss and dotted with giant bioluminescent purple mushrooms. A gentle mist hugs the ground, and a small, weathered stone bridge crosses a narrow, winding stream in the background.}{Change the stone bridge in the background into a flat, 2D cel-shaded cartoon drawing}{Violates the cohesive "ancient, mystical" atmosphere implied by the prompt by introducing a jarring artistic style clash.}{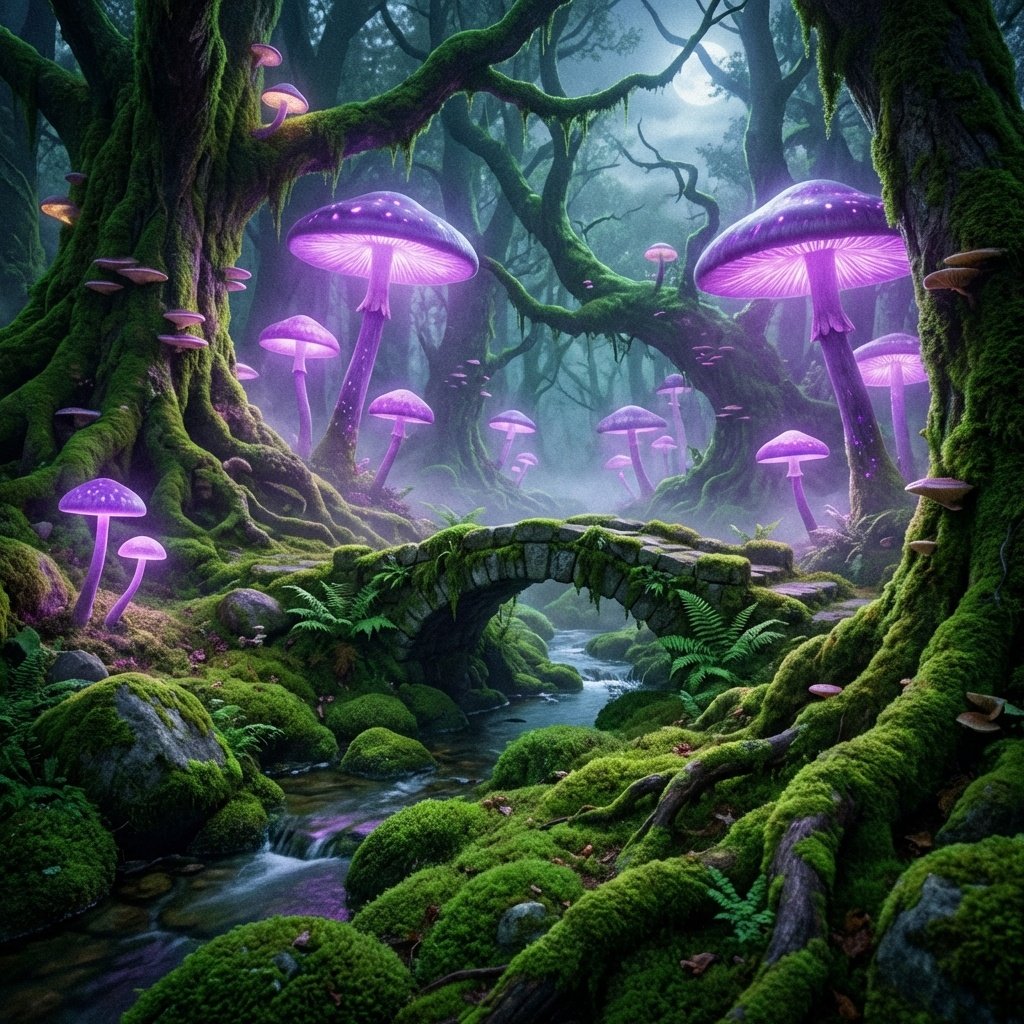}{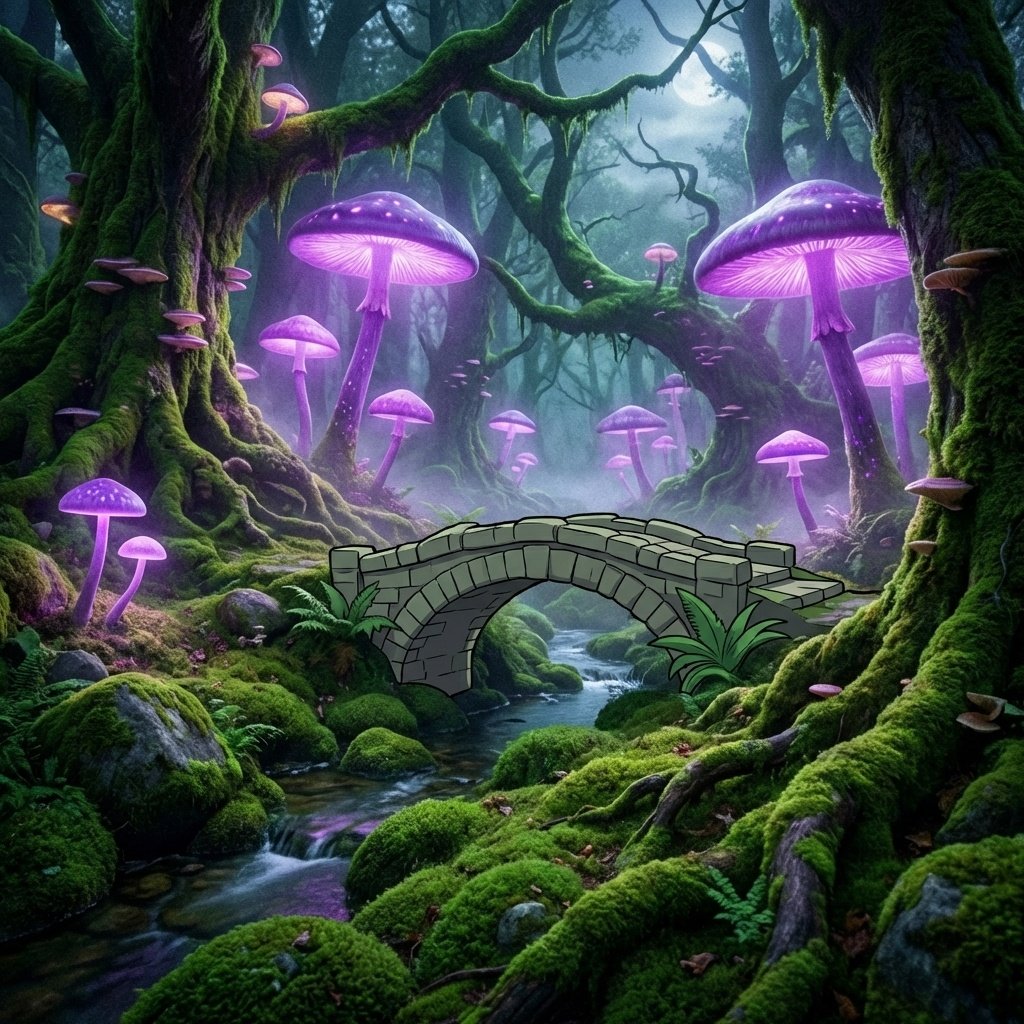}



%

\subsubsection{Theme Conflict}
\vspace{-0.9em}



\ttoiexample{An overflowing Victorian inventor's laboratory where a scientist in brass goggles is aggressively tightening a valve on a copper boiler in the foreground. The crowded midground features ticking clockwork automatons assembling themselves on wooden workbenches, while the background is filled from floor to ceiling with chalkboard schematics and glass tubes bubbling with green liquid.}{Place a modern digital multimeter with wire probes on the wooden workbench next to the brass automatons on the left}{Violates the strict "Victorian" time period constraint established in the prompt by introducing a modern electronic testing device. Easy to miss because a multimeter is a common diagnostic tool that semantically fits the broad concept of an "inventor's laboratory," causing the model to gloss over the historical anachronism.}{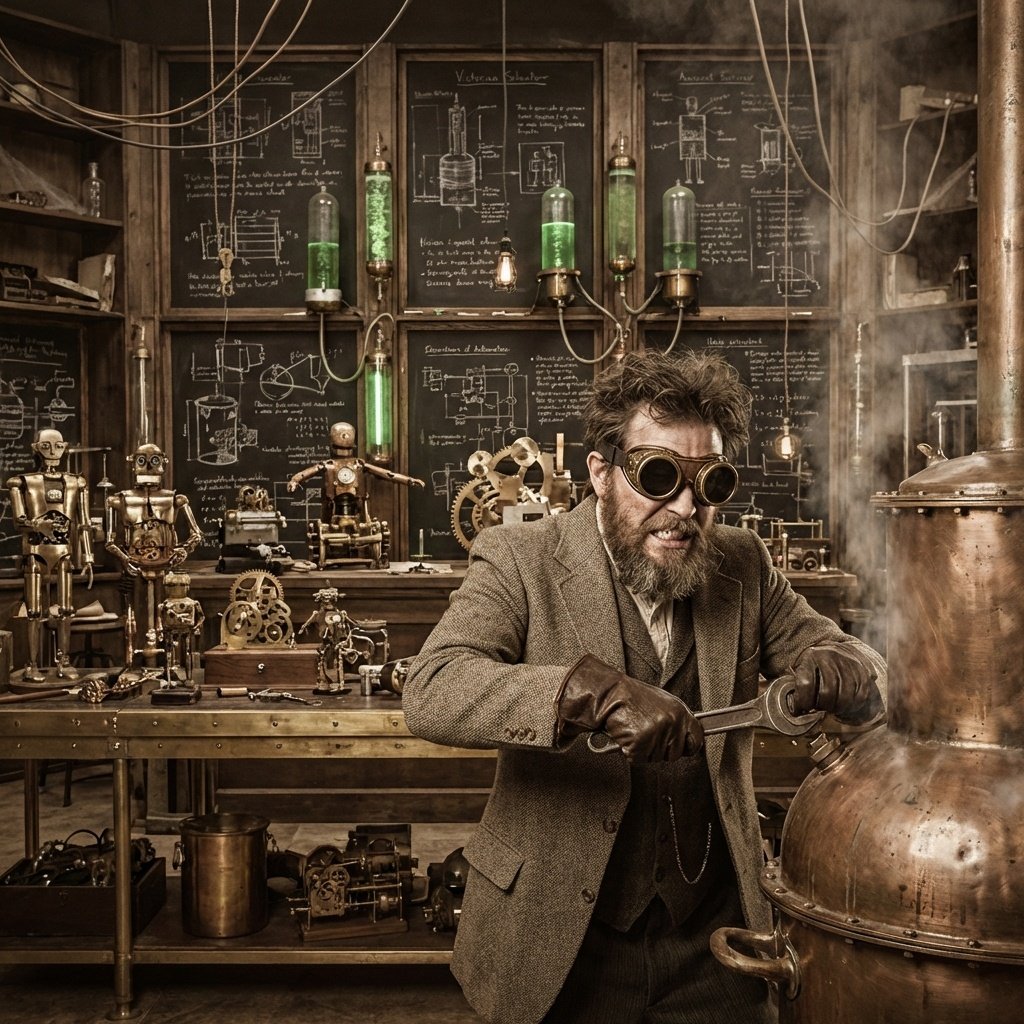}{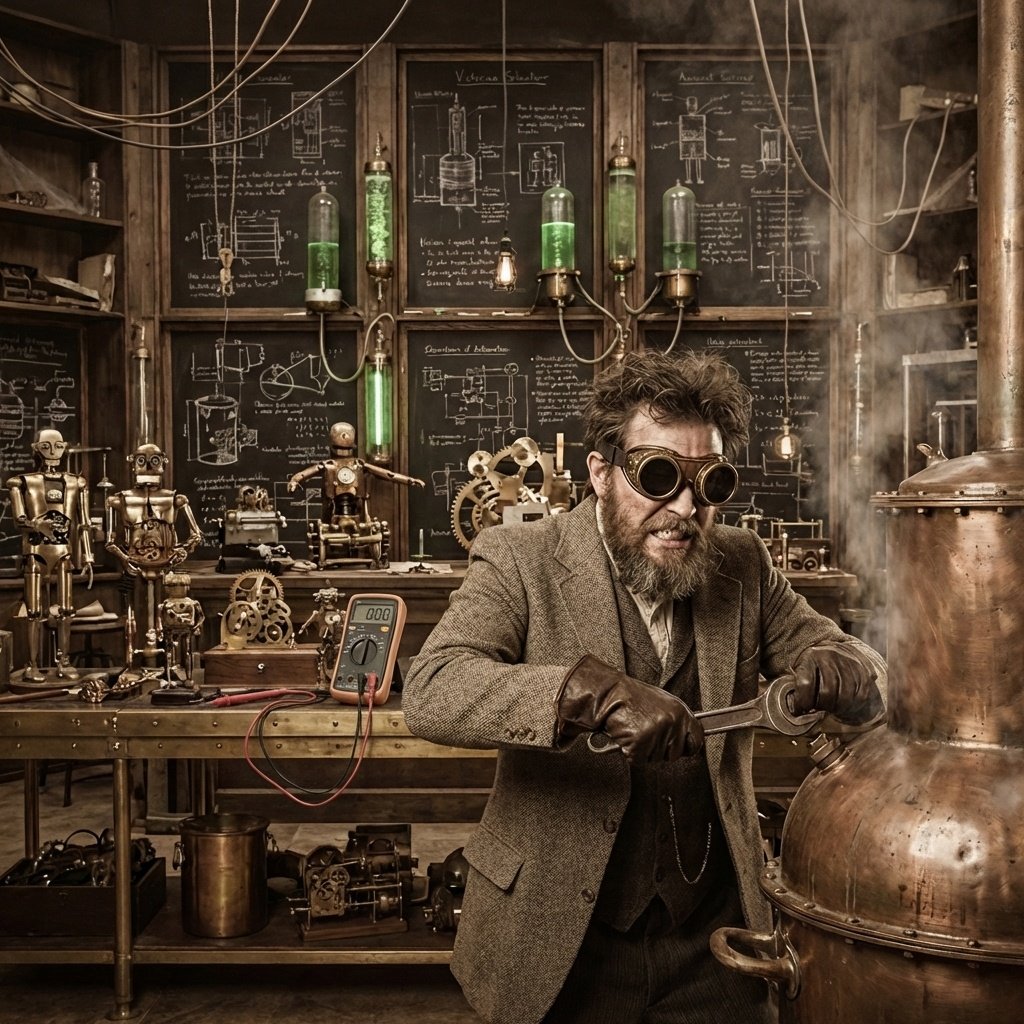}

\vspace{-1em}

\subsubsection{Disorganized Composition}
\vspace{-0.5em}


\vspace{0.55em}

\ttoiexample{A stop-motion claymation scene of a friendly green dinosaur wearing a chef's hat while aggressively kneading dough in a rustic kitchen. The visual texture must clearly show the tactile properties of modeling clay, including subtle thumbprints and a matte, imperfect surface.}{Move the chef's hat from the dinosaur's head and place it on the wooden table next to the dough}{Violates the prompt's requirement that the dinosaur must be "wearing a chef's hat." Easy to miss because it will still detect the presence of both the dinosaur and the chef's hat within the scene, potentially failing to verify the specific spatial relationship (wearing) between them.}{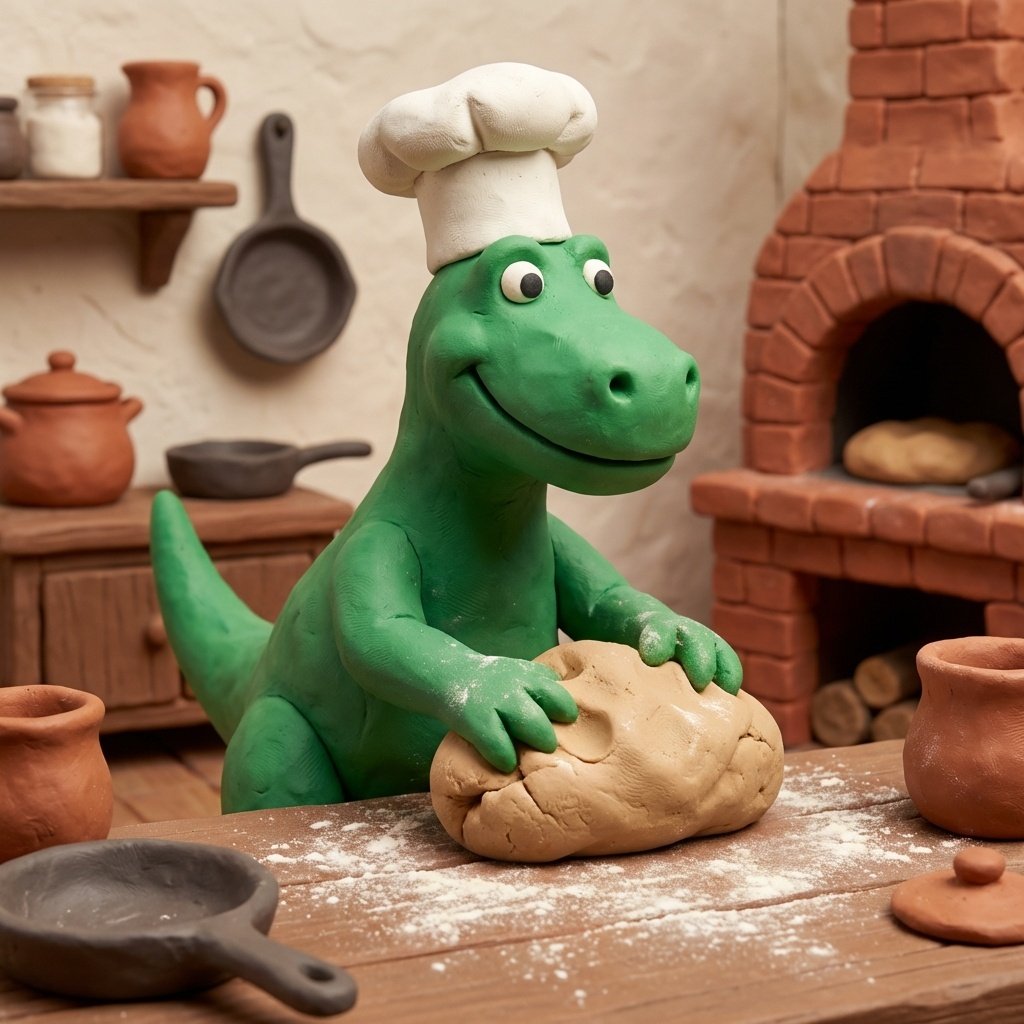}{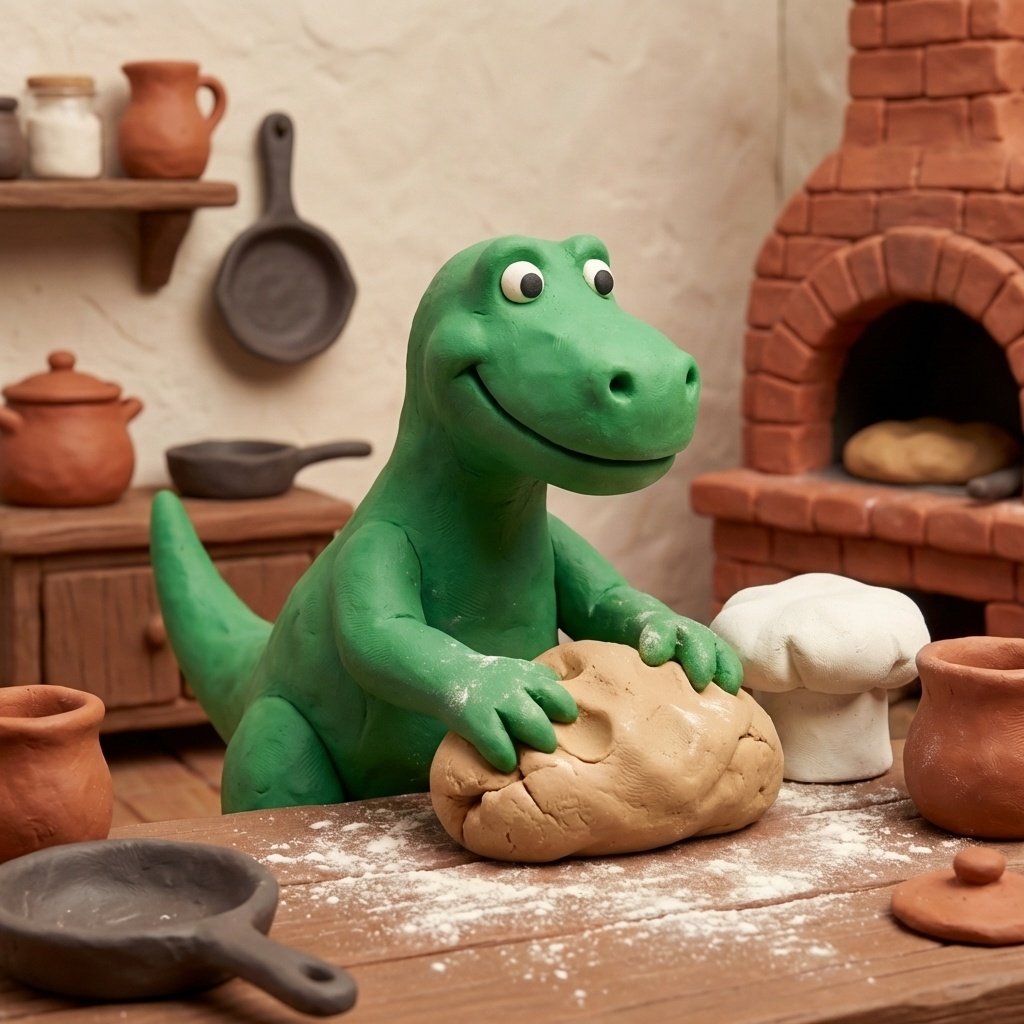}

\subsubsection{Overcrowding}
\vspace{-0.2em}



\ttoiexample{A serene 12th-century Song Dynasty scholarly gathering in a misty bamboo grove, featuring three scholars wearing authentic long flowing hanfu robes. They are seated around a low wooden table practicing calligraphy with traditional brush and ink wash, with no modern eyeglasses or contemporary writing utensils present.}{Cover the wooden table and the surrounding woven mats with dozens of scattered inkstones, a chaotic pile of extra calligraphy brushes, and numerous crumpled pieces of paper}{Violates the prompt's requirement for a "serene" scholarly gathering by introducing excessive visual clutter and chaos. Easy to miss because the added objects-brushes, inkstones, and paper-are contextually relevant to the calligraphy theme and match the historical setting's aesthetic.}{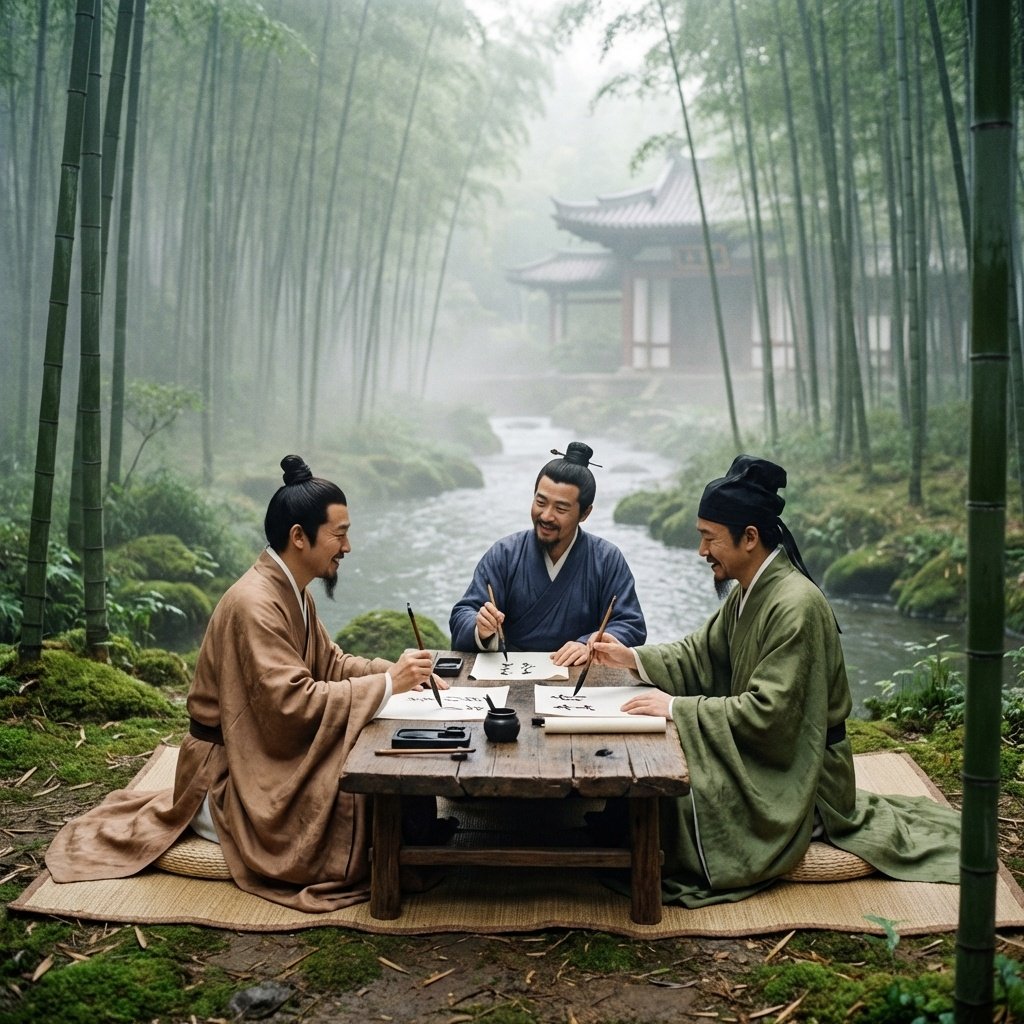}{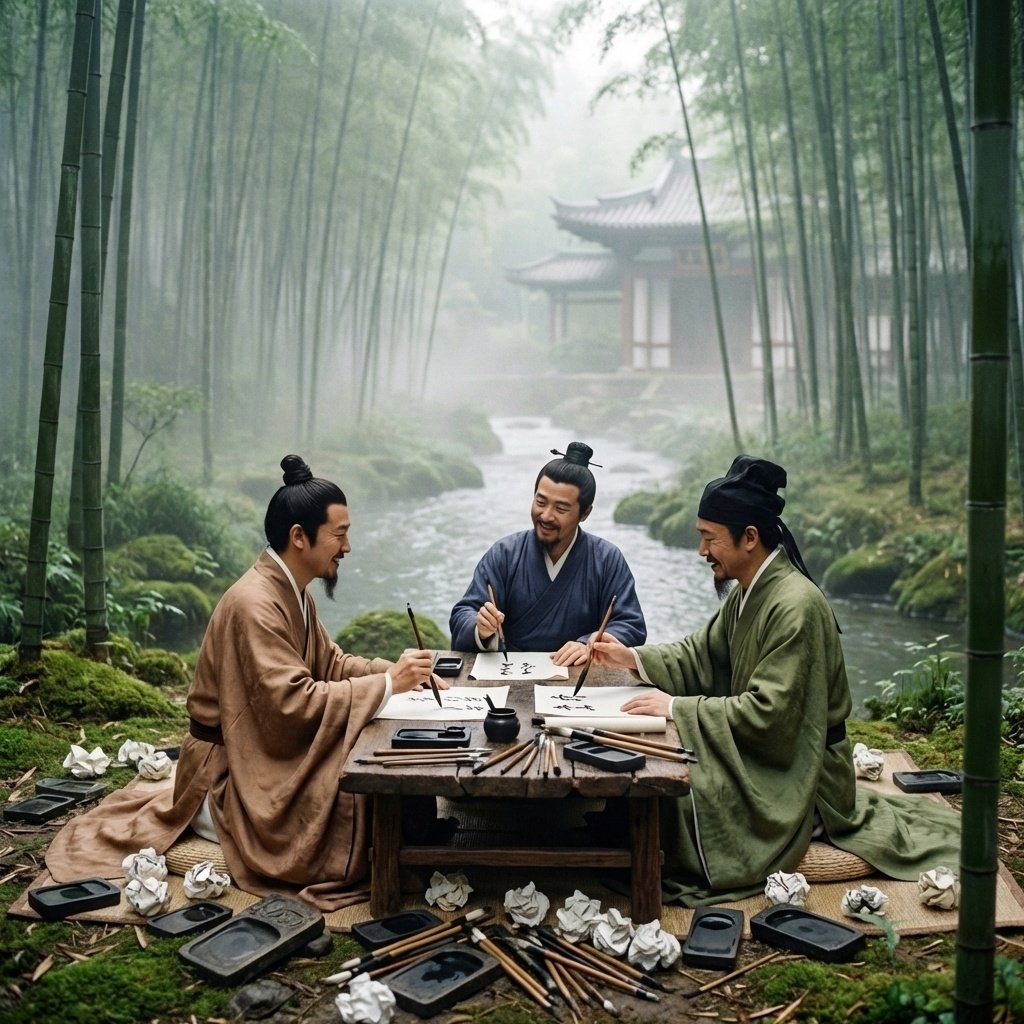}

\subsection{Physical Plausibility}

\subsubsection{Causal Violation}
\vspace{-0.2em}



\ttoiexample{A heavy black bowling ball rests at the center of a large circular trampoline, causing the fabric to visibly stretch deeply downwards under the concentrated weight of the ball.}{Make the trampoline fabric perfectly flat and taut, removing the deep downward depression and stretching wrinkles around the bowling ball}{Violates the prompt's explicit causal requirement that the ball's weight causes the fabric to "visibly stretch deeply downwards." Easy to miss because the primary subjects-the black bowling ball and the circular trampoline-remain prominent and recognizable.}{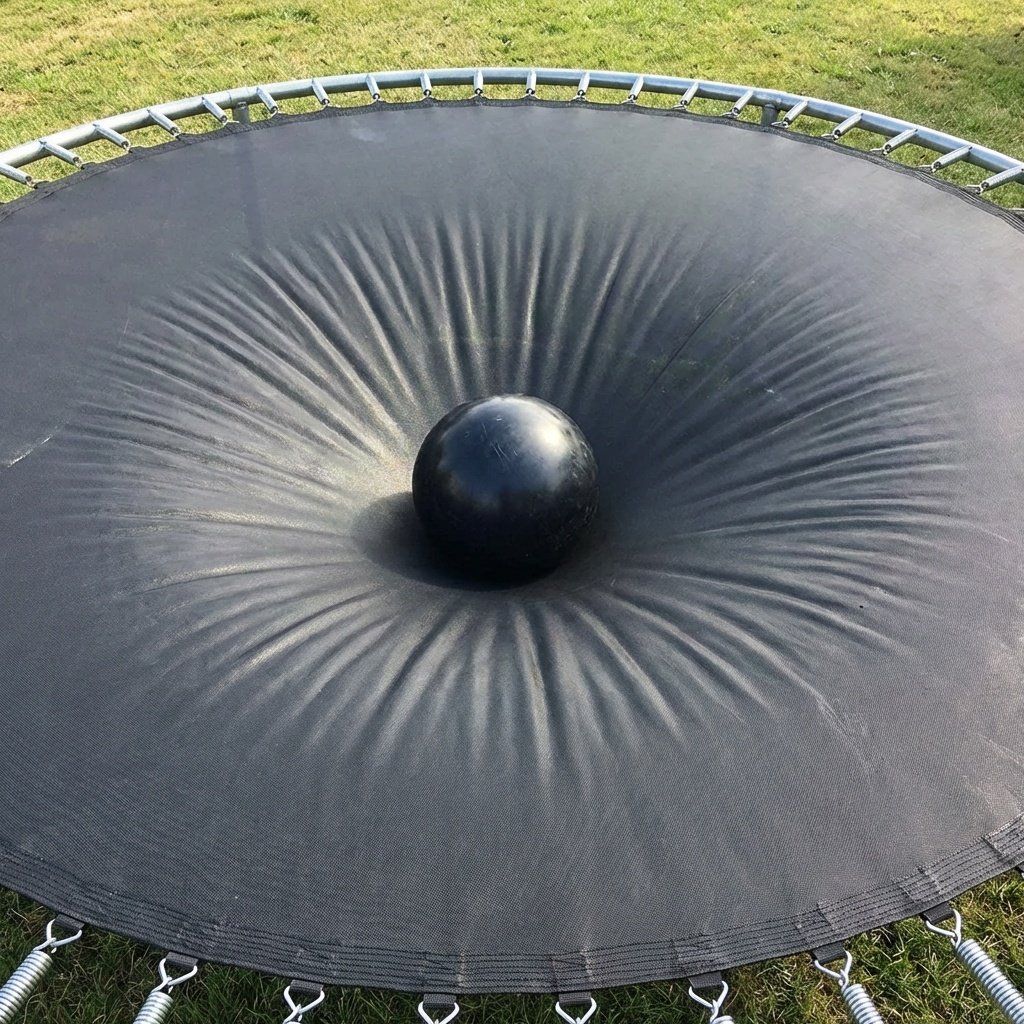}{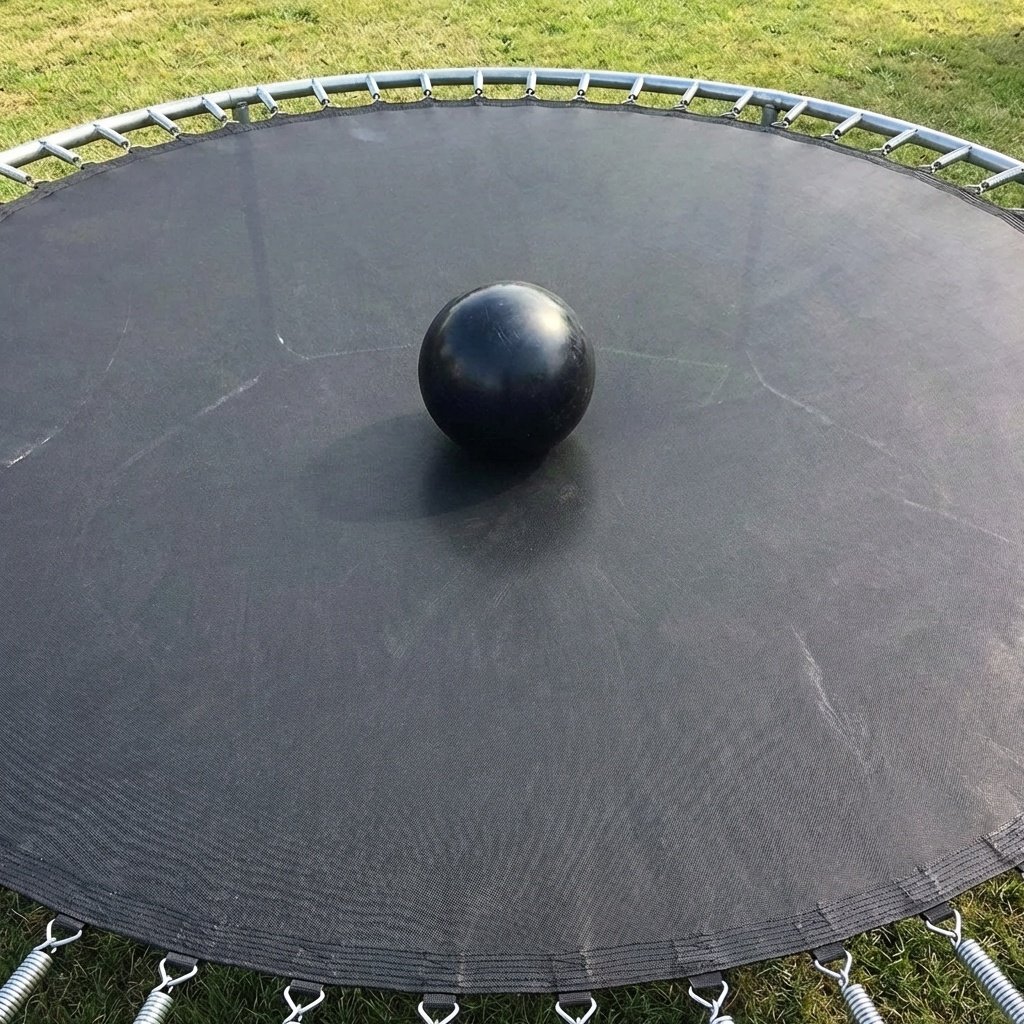}

\subsubsection{Physics Manipulation}
\vspace{-0.2em}

\ttoiexample{A clear triangular glass prism resting on a matte black table, with a single beam of white light entering one side and refracting out the other side as a distinct, accurate rainbow spectrum of colors.}{Swap the appearances of the light beams so that a rainbow spectrum enters the prism from the left and a single white beam exits onto the table on the right}{Violates the prompt's explicit requirement of a "single beam of white light entering one side and refracting out the other side as a distinct, accurate rainbow spectrum." Easy to miss because all the semantic elements requested by the prompt (a glass prism, a white light beam, and a rainbow spectrum) remain clearly visible in the scene.}{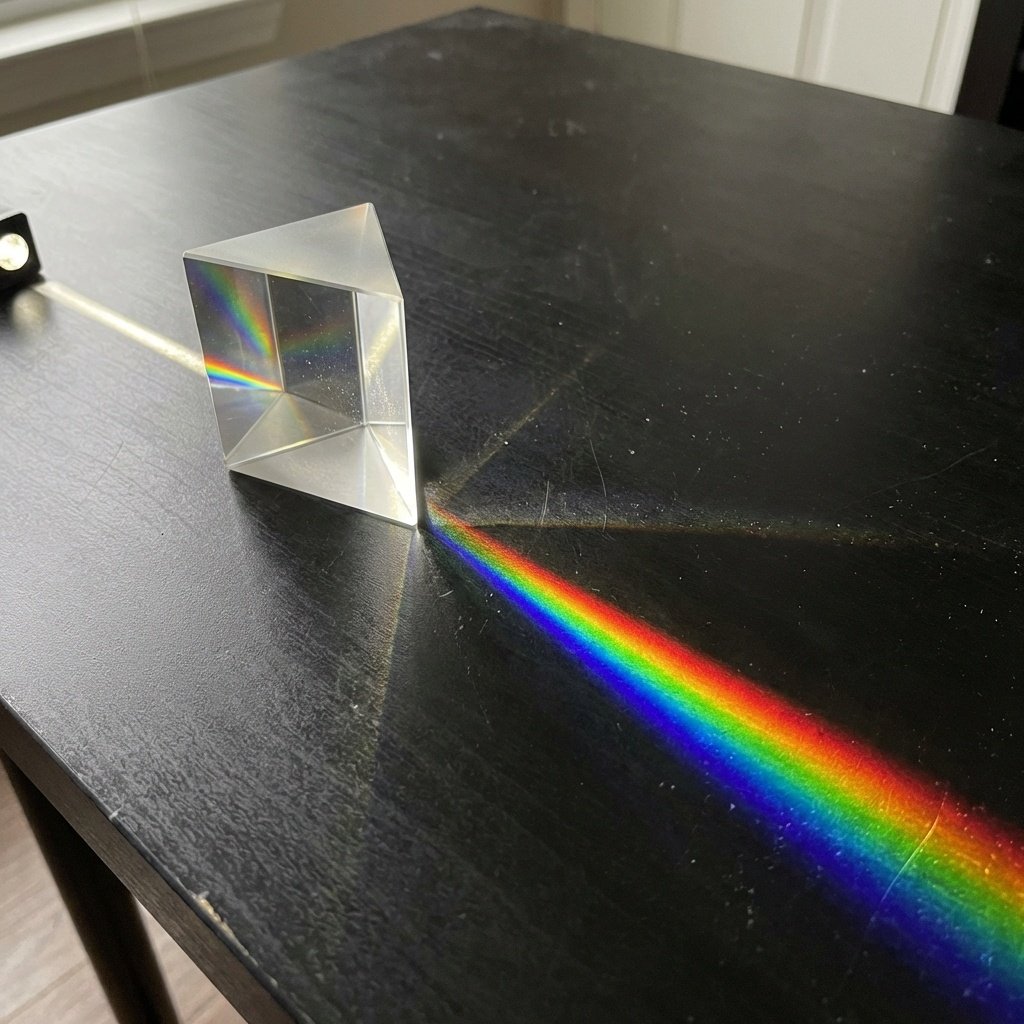}{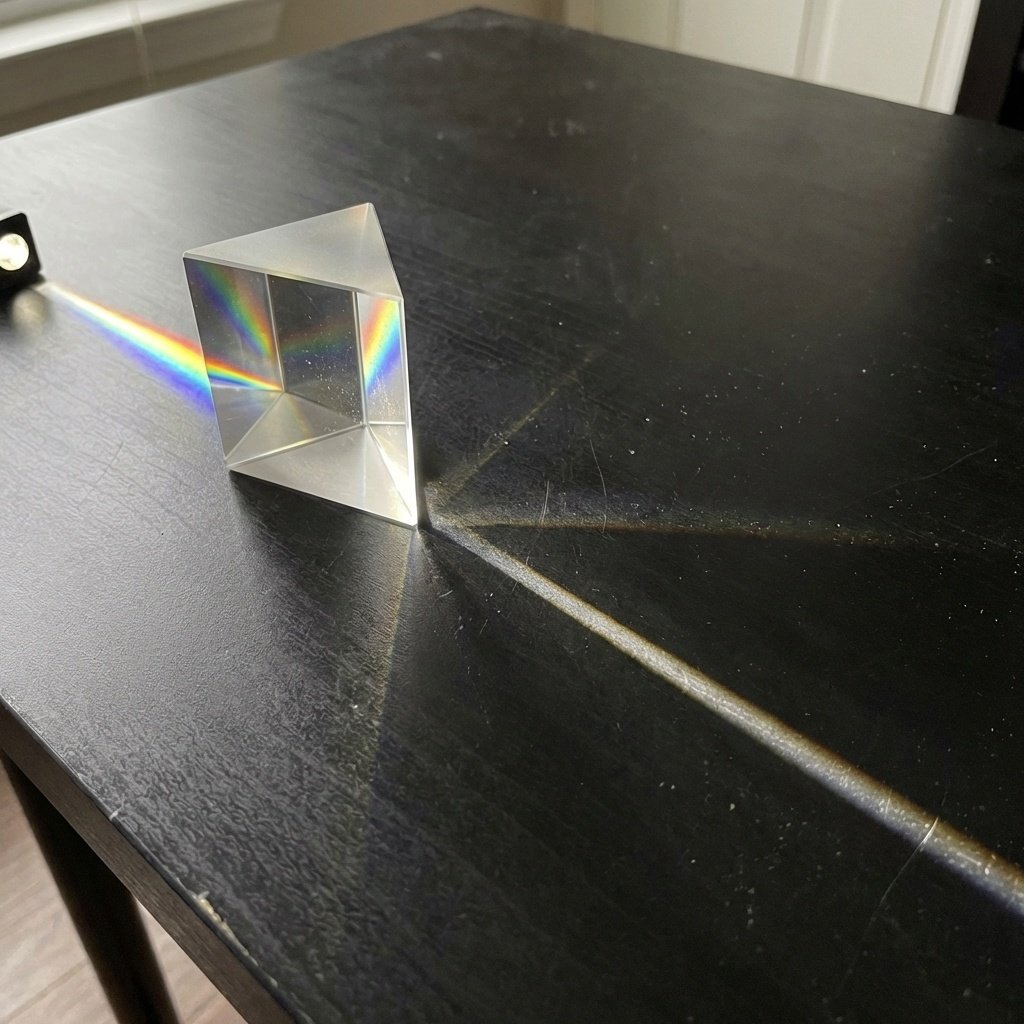}



\subsubsection{State/Transformation Failure}
\vspace{-0.9em}

\ttoiexample{A clear, perfectly spherical glass paperweight rests on a black and white checkered surface, with the checkerboard pattern appearing visibly warped and inverted inside the glass sphere due to optical refraction.}{Replace the warped checkerboard pattern inside the glass sphere with perfectly straight squares that align seamlessly with the background pattern}{Violates the prompt's explicit requirement that the checkerboard pattern appears "visibly warped and inverted inside the glass sphere due to optical refraction." Easy to miss because the image still features a spherical glass object and a prominent checkerboard pattern, superficially matching the main subjects of the text.}{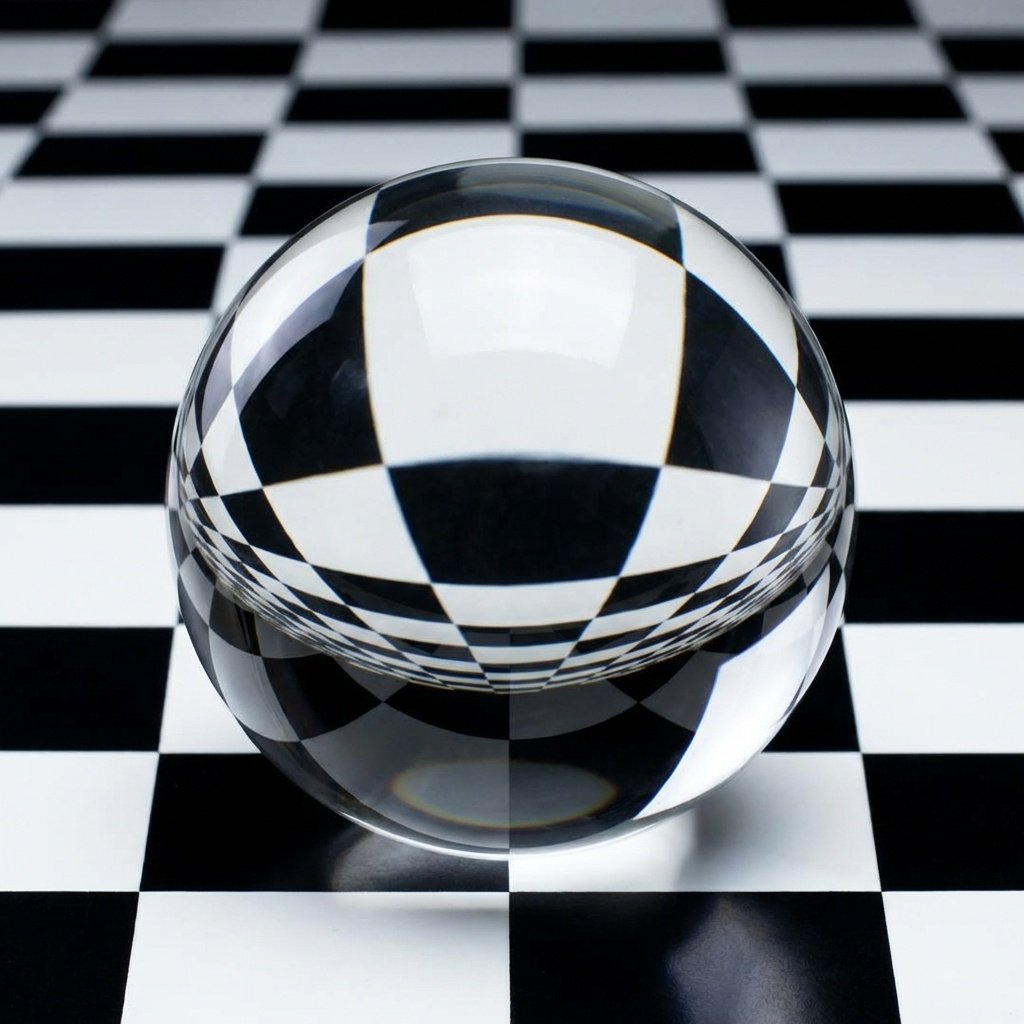}{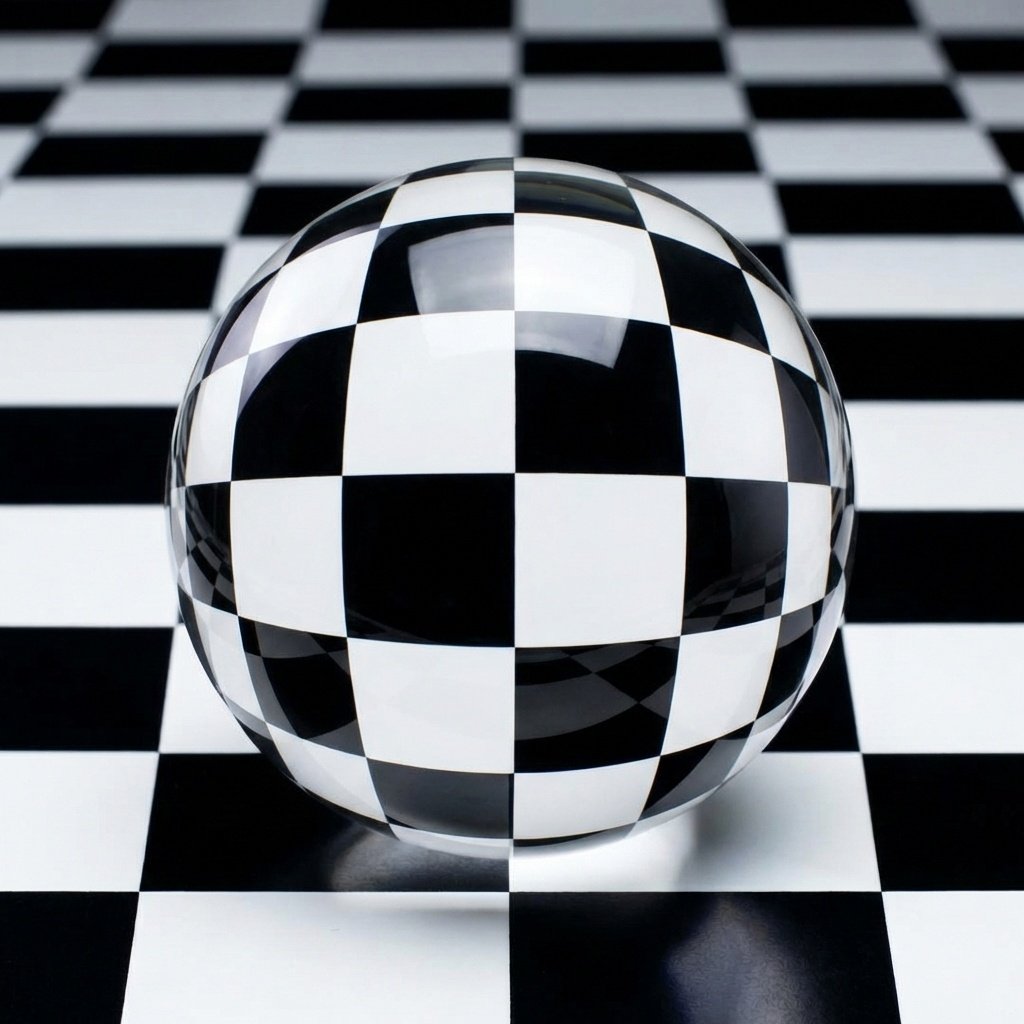}



%

\subsubsection{Functional Absurdity}
\vspace{-0.9em}



\ttoiexample{A triangular glass prism resting on a black surface in a dark room, intercepting a single beam of white light and accurately dispersing it into a distinct rainbow spectrum of colors on the opposite side.}{Replace the dispersed rainbow spectrum emerging from the prism with a single, undispersed beam of white light}{Violates the prompt's requirement that the prism is "accurately dispersing it into a distinct rainbow spectrum of colors." Easy to miss because the explicitly requested "rainbow spectrum" is missing and the optical action is incorrect, even though the resulting image still looks like a highly realistic rendering of light passing through glass.}{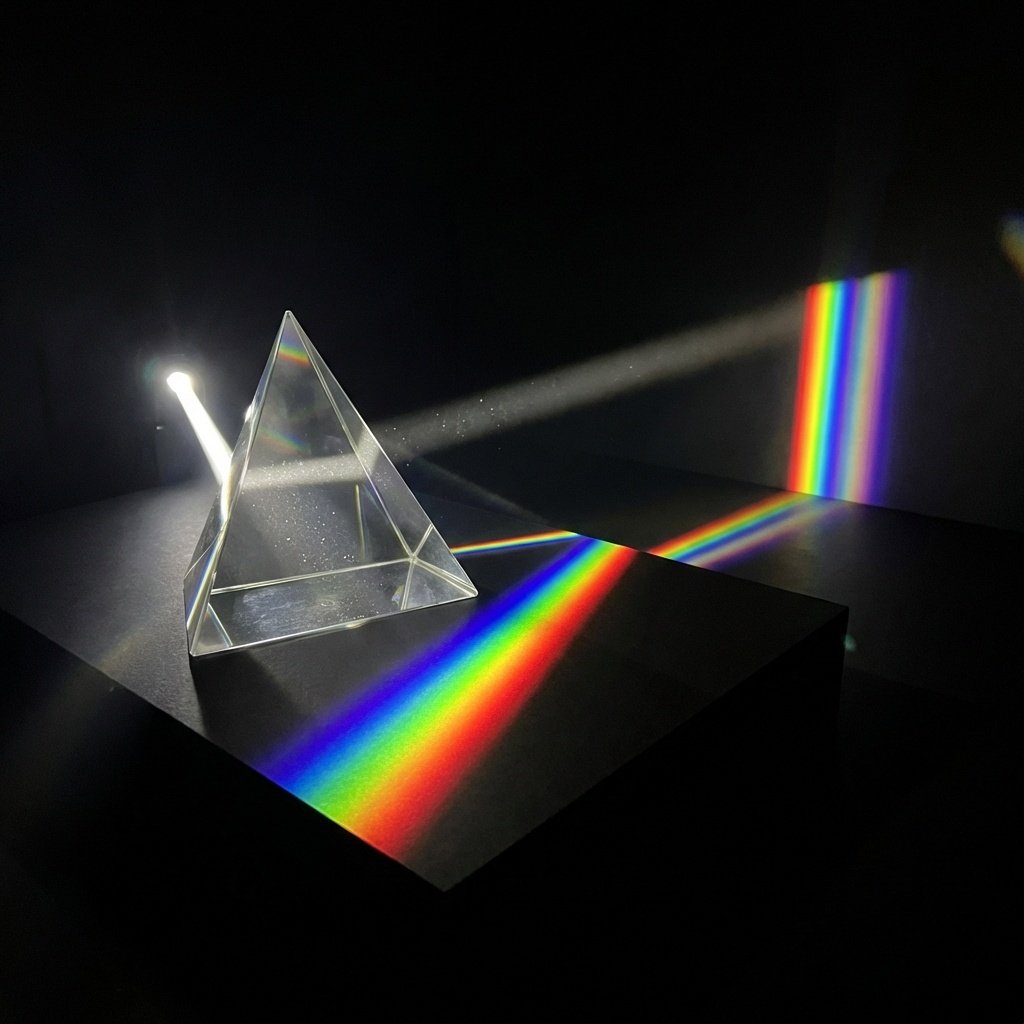}{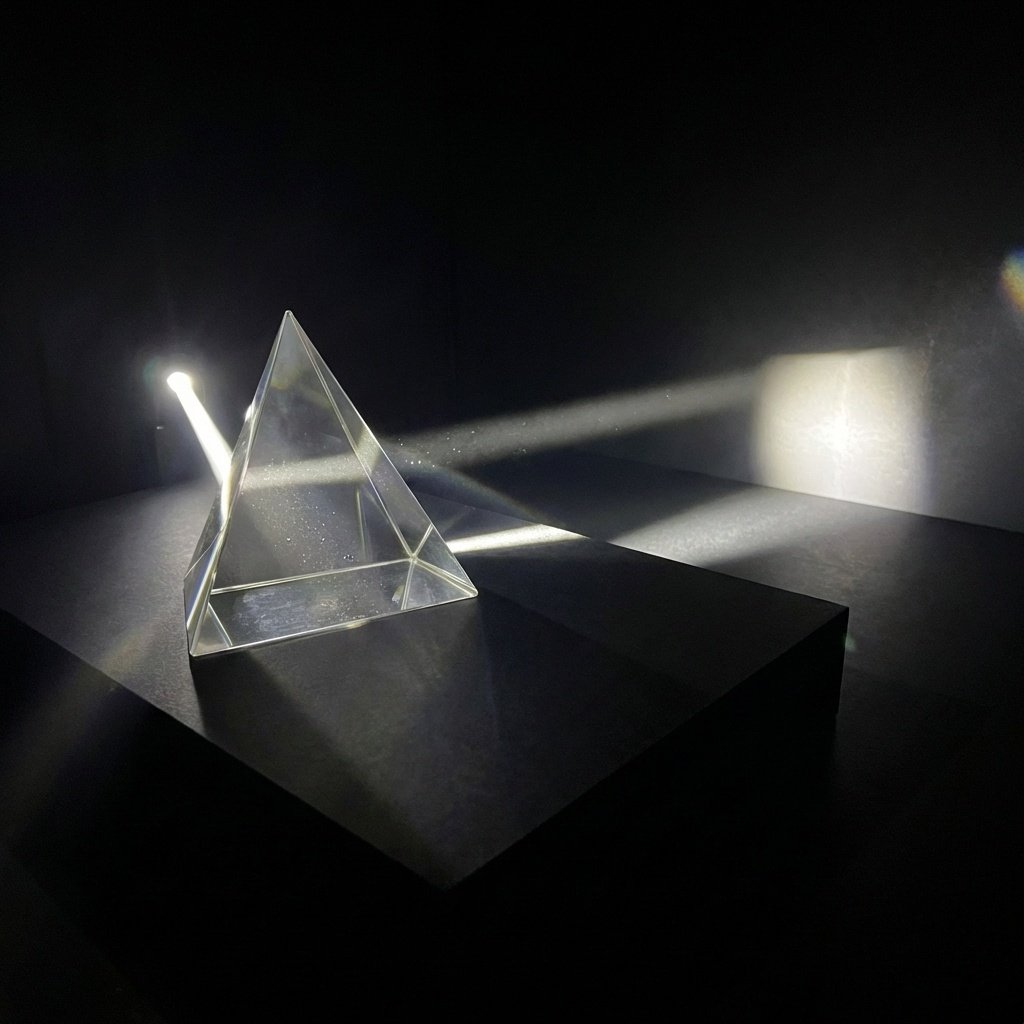}

\subsubsection{Literalized Idioms}



\ttoiexample{A bright spotlight shines directly from the left onto a solid red sphere sitting on a flat white table. The sphere casts a single, sharp shadow stretching horizontally across the table exclusively to the right.}{Modify the far right end of the shadow so it tapers into a literal sharp, pointy tip like a blade, rather than a natural rounded oval}{Violates the physical reality of a shadow cast by a "sphere," which should naturally have a rounded, elliptical shape. Easy to miss because the shadow still stretches horizontally to the right and maintains its crispness and dark tone, making the physically impossible shape easy to miss at a glance while falsely satisfying the word "sharp.".}{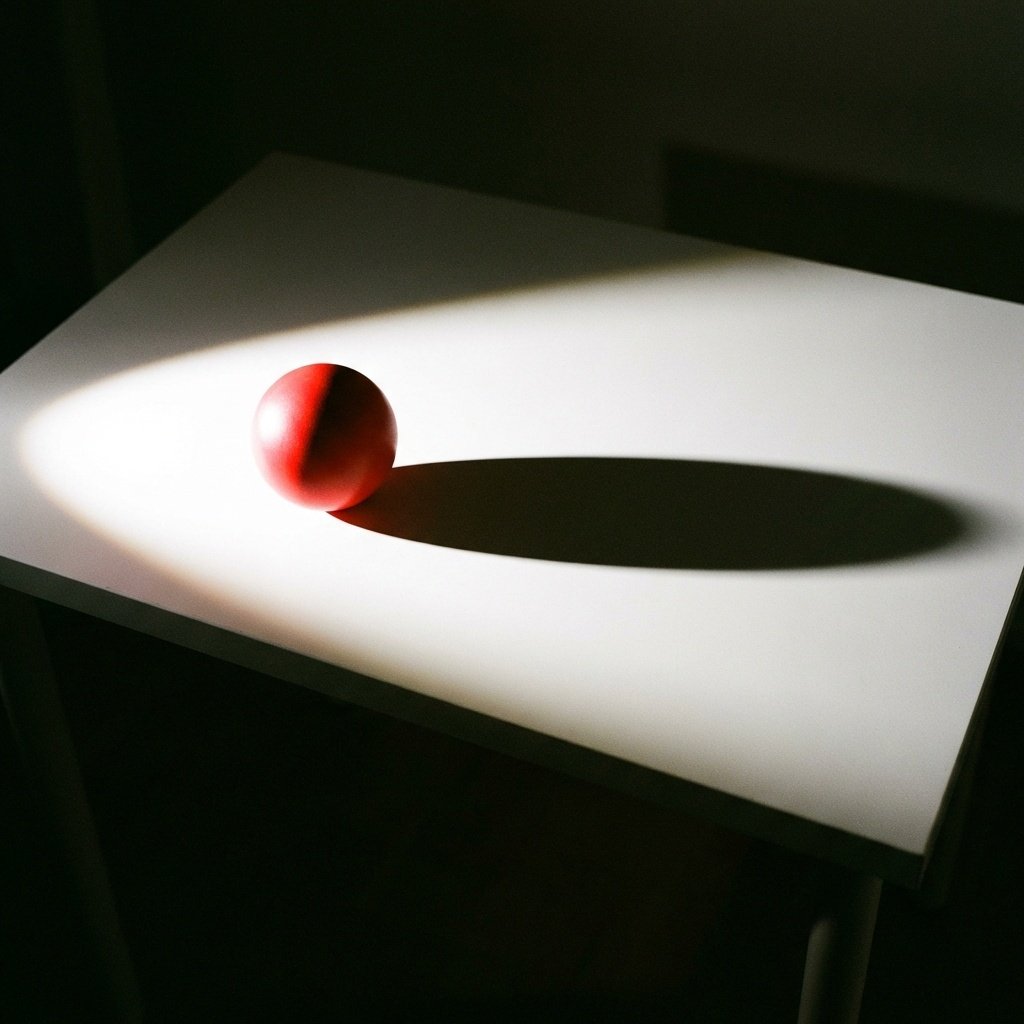}{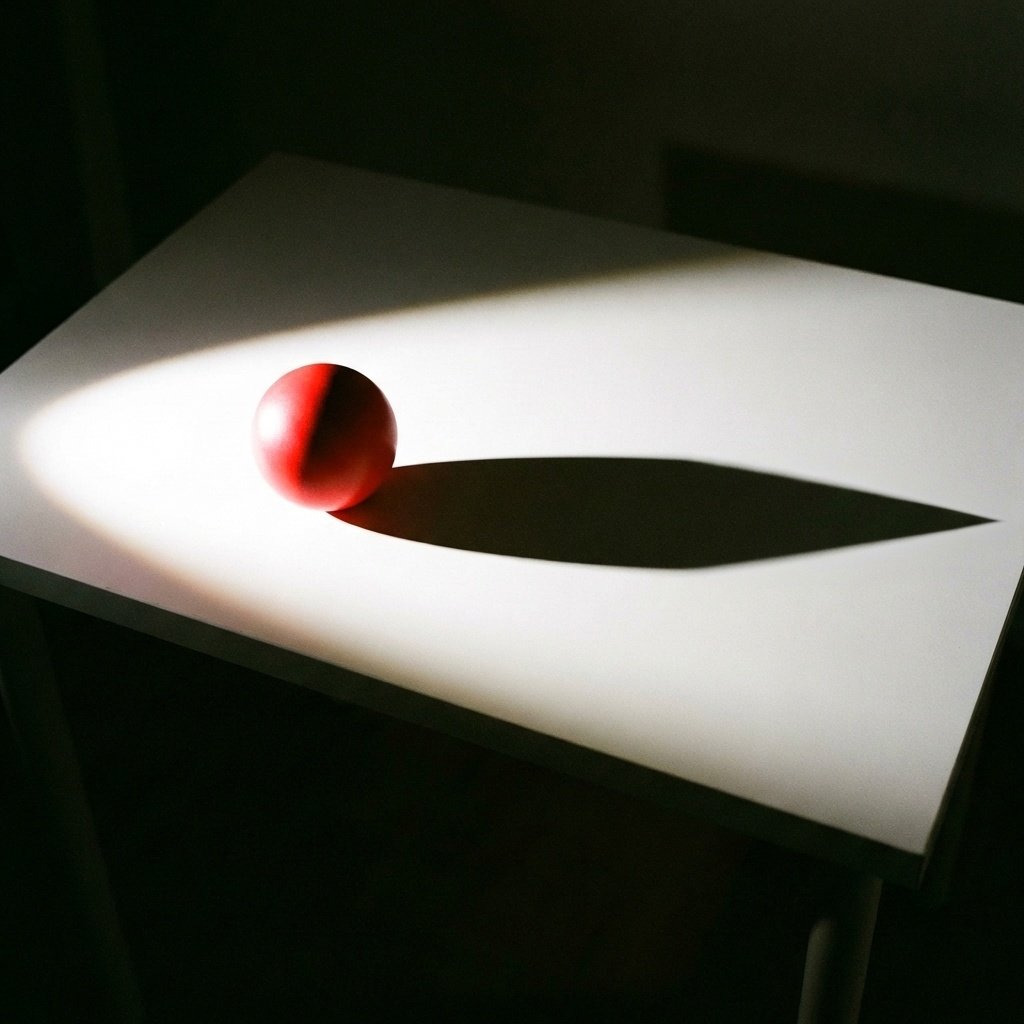}

\subsection{Text Rendering}
\subsubsection{Text/Typographical Corruption}
\vspace{-0.2em}

\ttoiexample{The word "Breathe" written by a finger in the condensation of a foggy windowpane, with a blurry dark green pine forest visible through the glass.}{Change the word "Breathe" written on the window to "Braethe"}{Violates the prompt's explicit requirement to feature the specific word "Breathe" by introducing a typographical error ("Braethe"). Easy to miss because it may process the text holistically or rely on subword tokens, recognizing the general shape and meaning of the word without scrutinizing the exact spelling, especially since all other visual elements perfectly match the prompt.}{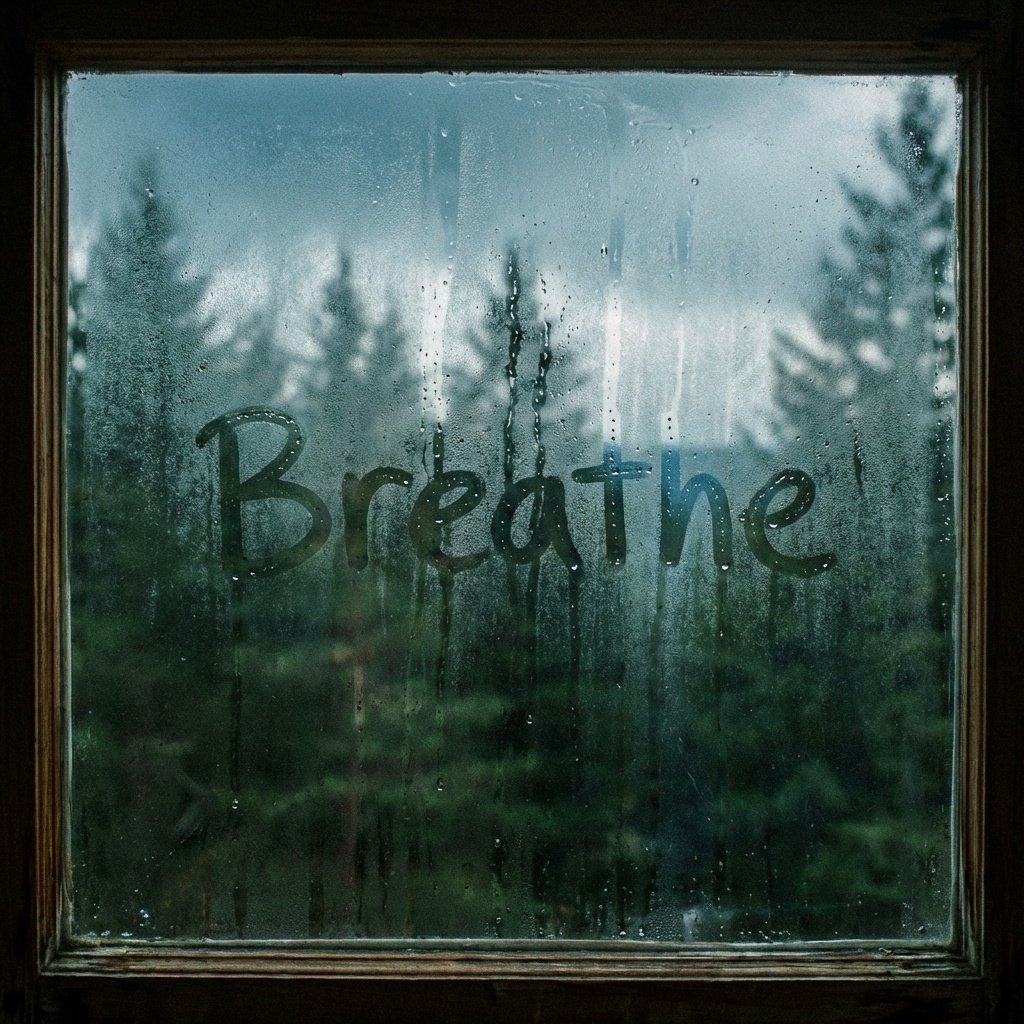}{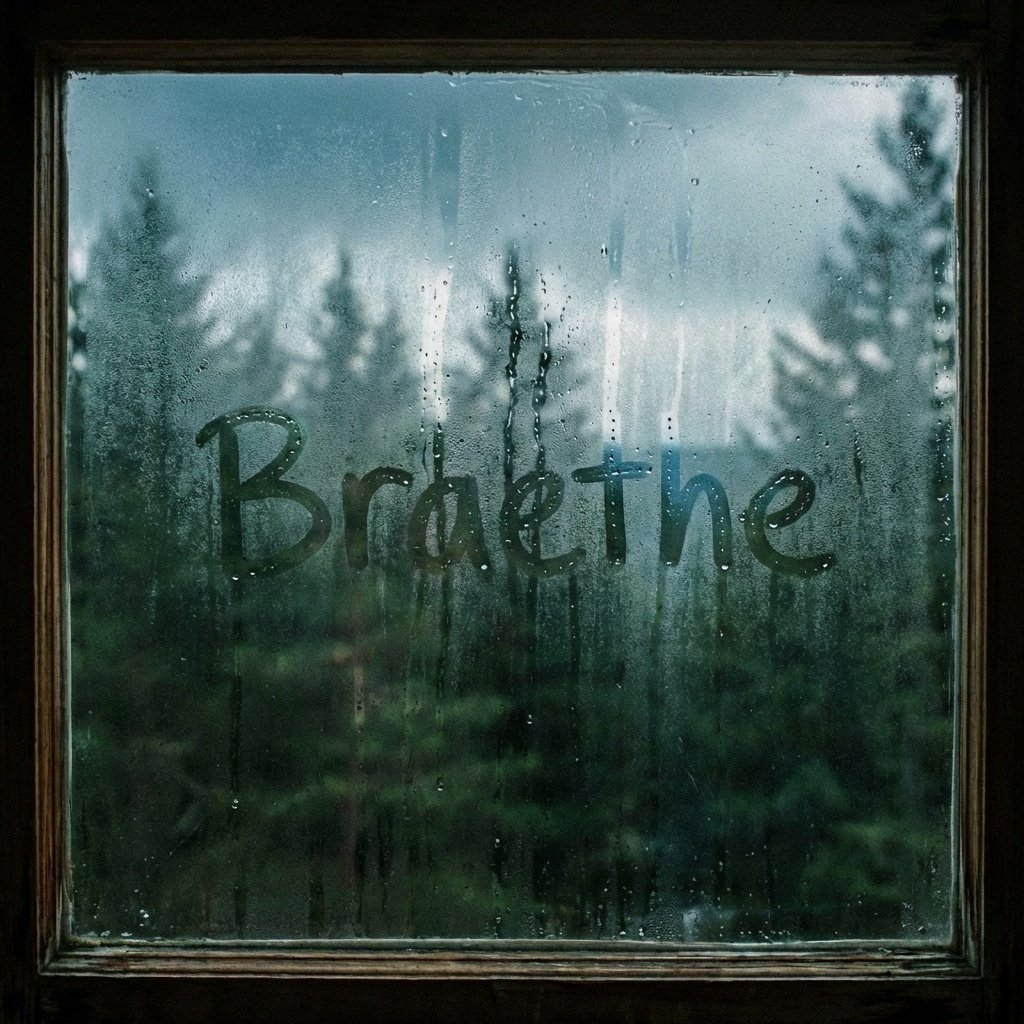}

\vspace{0.55em}

\ttoiexample{A standard wheelchair accessibility icon painted in blue and white on a flat asphalt surface. The word "RESERVED" is written in large, clear white painted letters just below the icon.}{Change the word "RESERVED" to "RESERVEB"}{Violates the prompt's explicit requirement that the word "RESERVED" be written below the wheelchair icon. Easy to miss because it may quickly glance at the text, recognize the general shape and context of the word, and fail to notice the single-character substitution.}{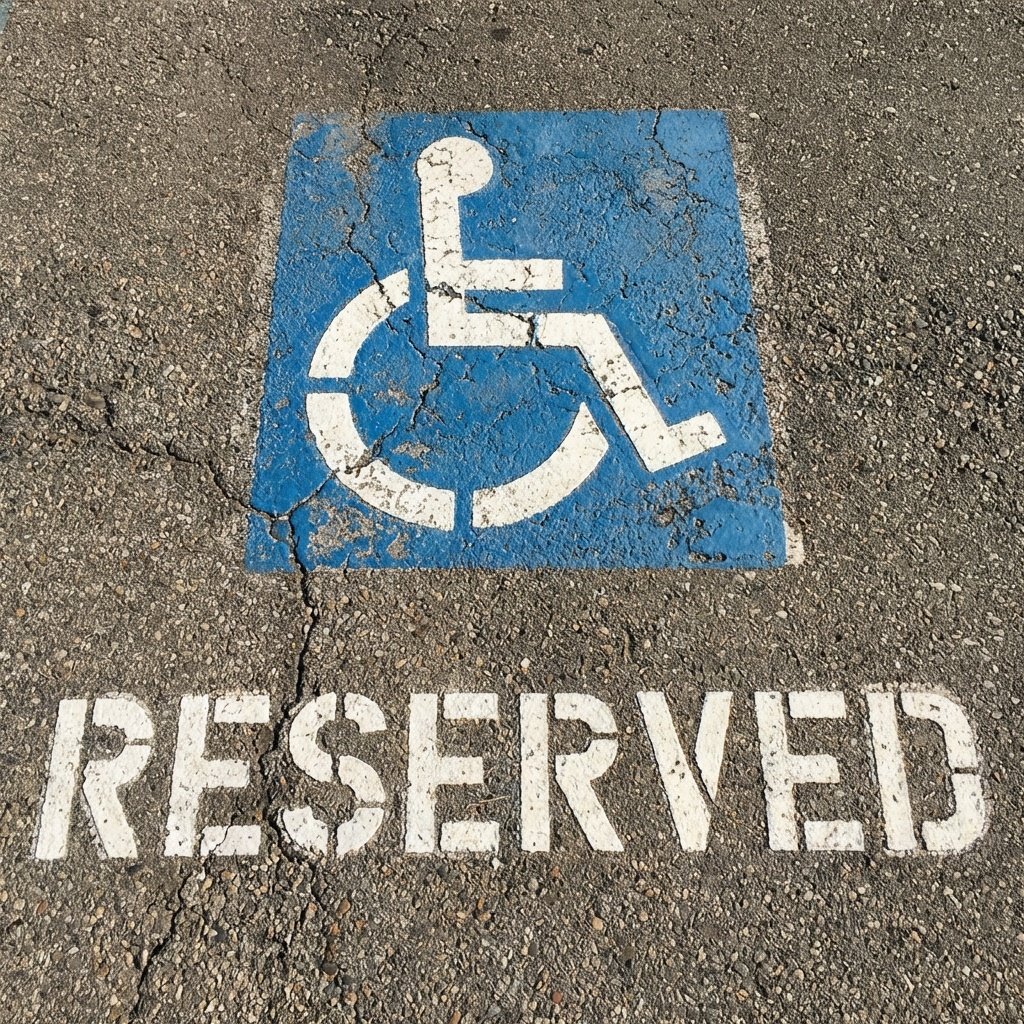}{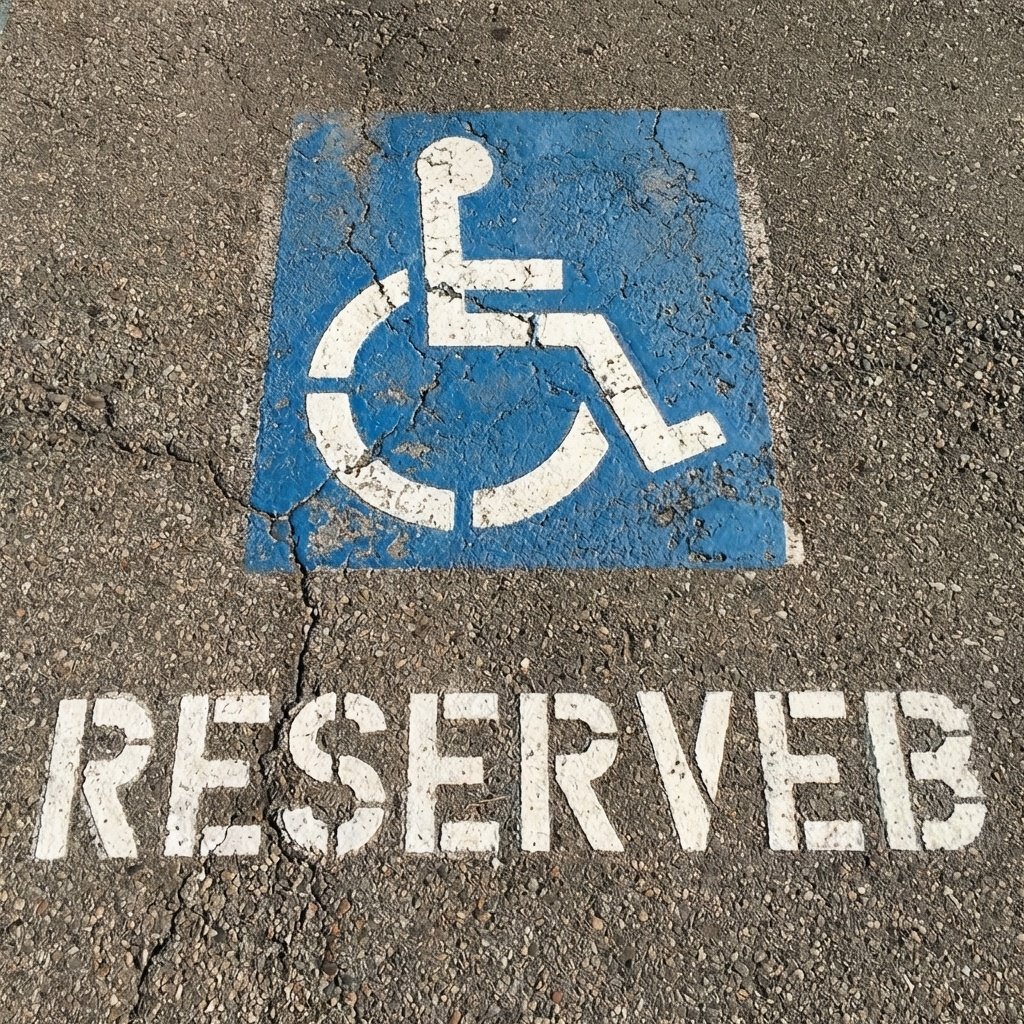}

\subsubsection{Incomplete Rendering}
\vspace{-0.2em}



\ttoiexample{A glowing green biohazard symbol painted on a rusted metal door, with the text "KEEP OUT" stenciled directly below it in stark black letters.}{Erase the word "OUT" from the black stenciled text on the door so it only reads "KEEP"}{Violates the prompt's explicit requirement that the text "KEEP OUT" must be stenciled below the biohazard symbol. Easy to miss because the remaining word "KEEP" retains the correct stark black stencil style, font, and placement, easily tricking a VLM that superficially detects the presence of stenciled text without reading and verifying the exact required phrase.}{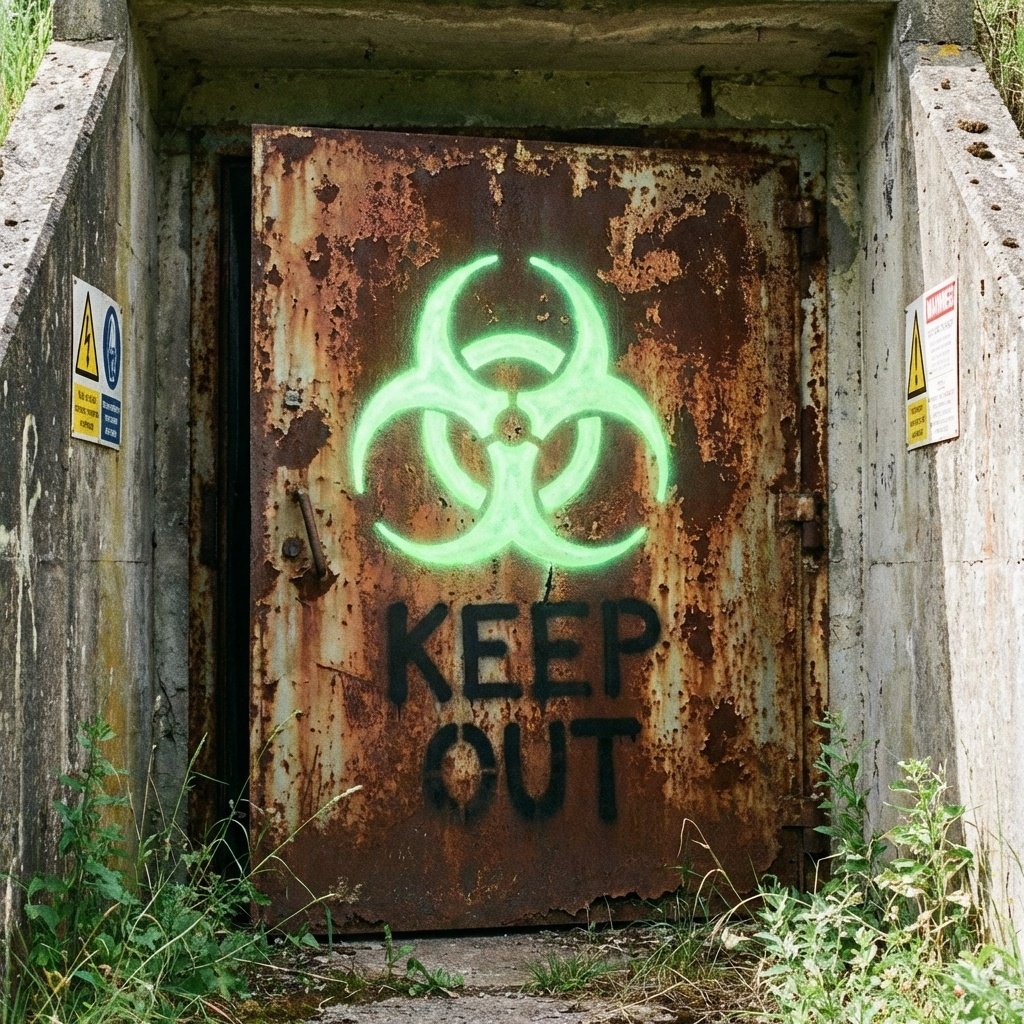}{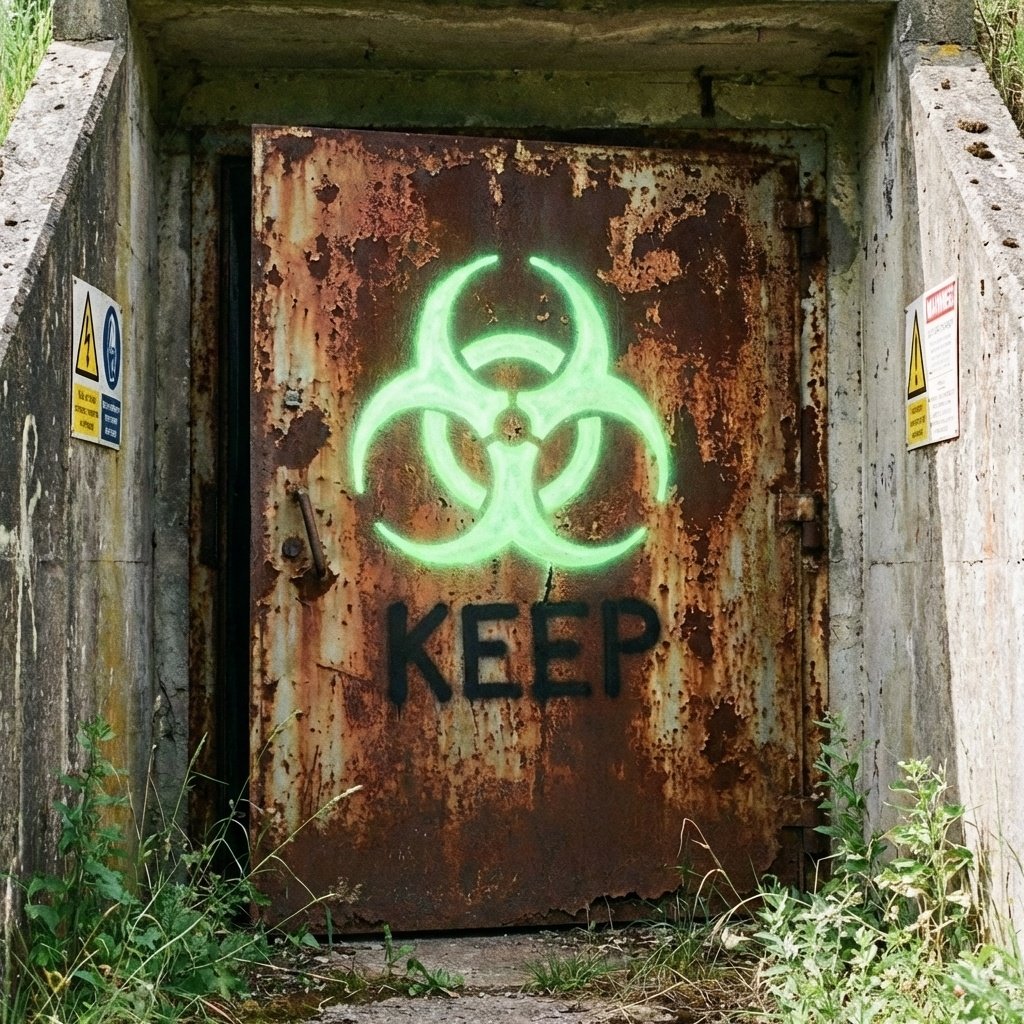}

\clearpage
\subsubsection{Background Misrendering}
\vspace{-0.2em}

\ttoiexample{A rustic, weathered wooden directional sign post on a lush green mountain hiking trail, with the text "BEAR LAKE 2 MILES" deeply carved and painted white.}{Erase the horizontal wooden board entirely, leaving only the white text "BEAR LAKE 2 MILES" floating in mid-air in its original position, while keeping the vertical wooden post and the background intact}{Violates the prompt's requirement that the text is "deeply carved" into a "rustic, weathered wooden" sign post by completely removing the immediate physical surface the text relies on. Easy to miss because it will successfully detect the exact text string "BEAR LAKE 2 MILES", the vertical wooden post, and the lush green mountain trail, satisfying its semantic checklist.}{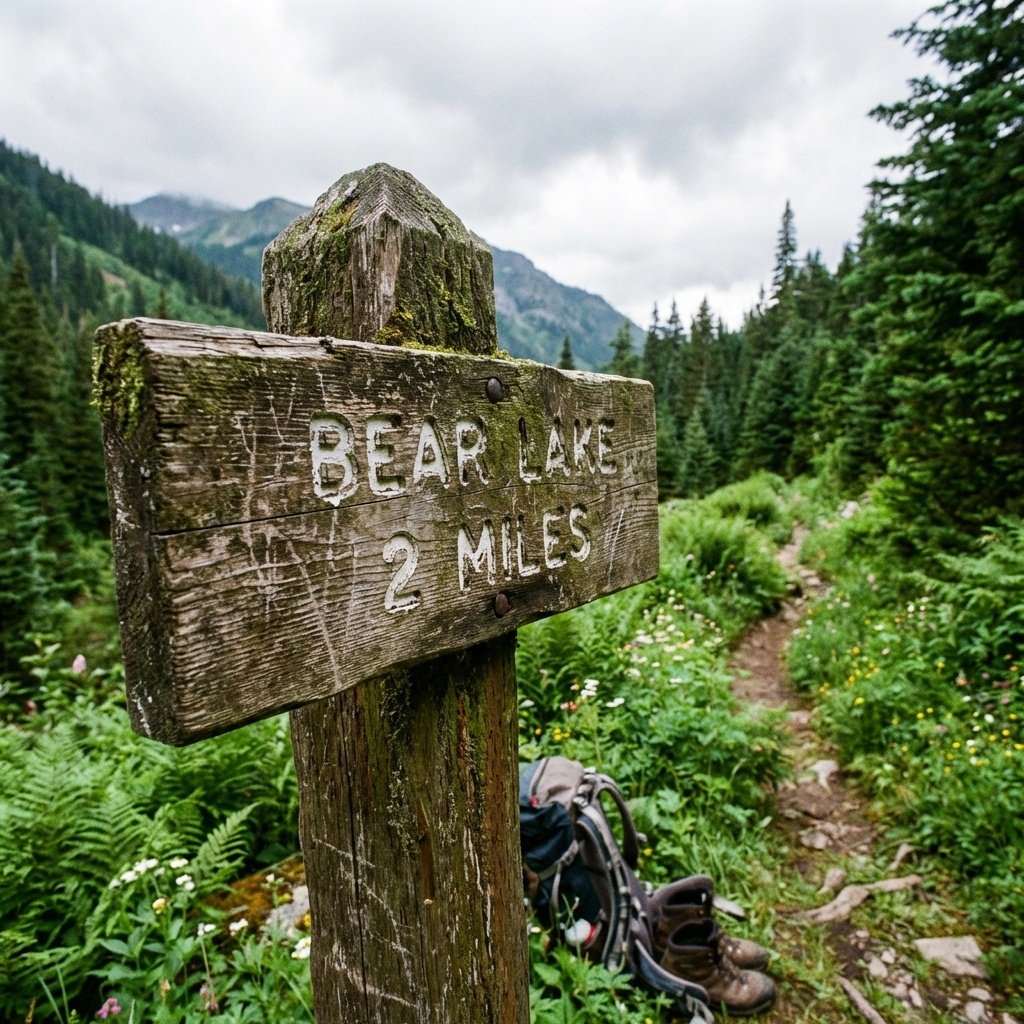}{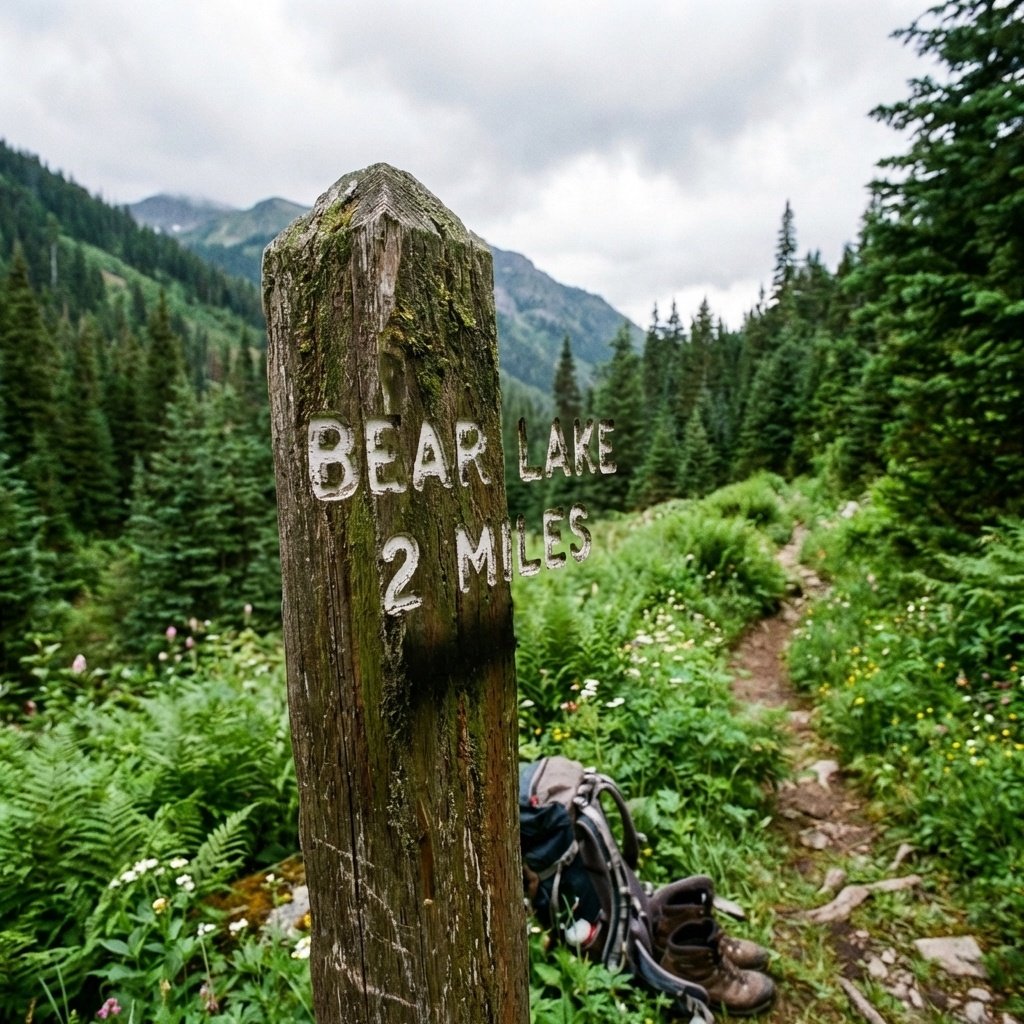}



\subsubsection{Mislabeled Symbols/Diagrams}
\vspace{-0.2em}

\ttoiexample{A vintage green bomber jacket hanging on a wooden coat rack. The back of the jacket has a large, arched patch featuring the embroidered text 'WILDERNESS EXPLORER' in bright orange thread.}{Change the embroidered word 'WILDERNESS' on the patch to 'WILDLIFE'}{Violates the prompt's requirement that the jacket patch feature the exact text 'WILDERNESS EXPLORER'. Easy to miss because the new word shares the same prefix ("WILD") and semantic theme, keeping the overall context of the image intact.}{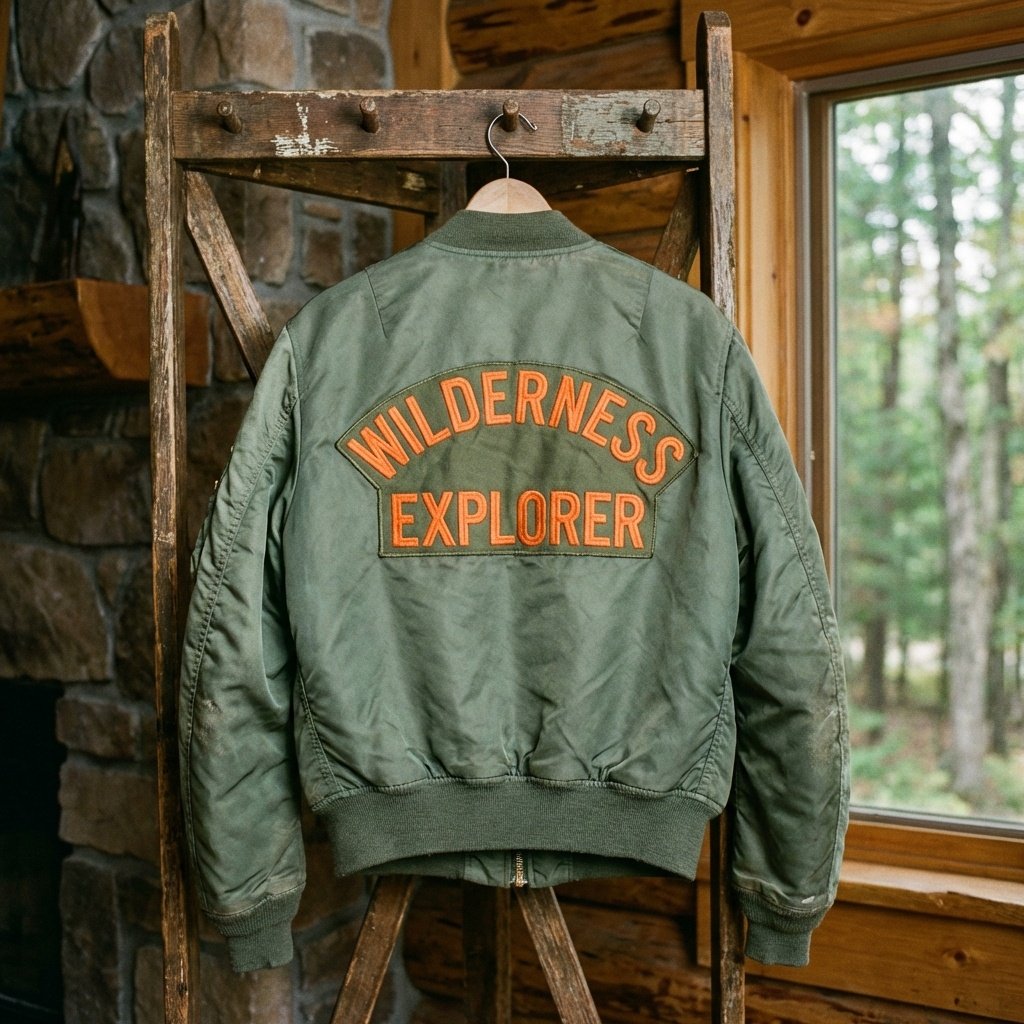}{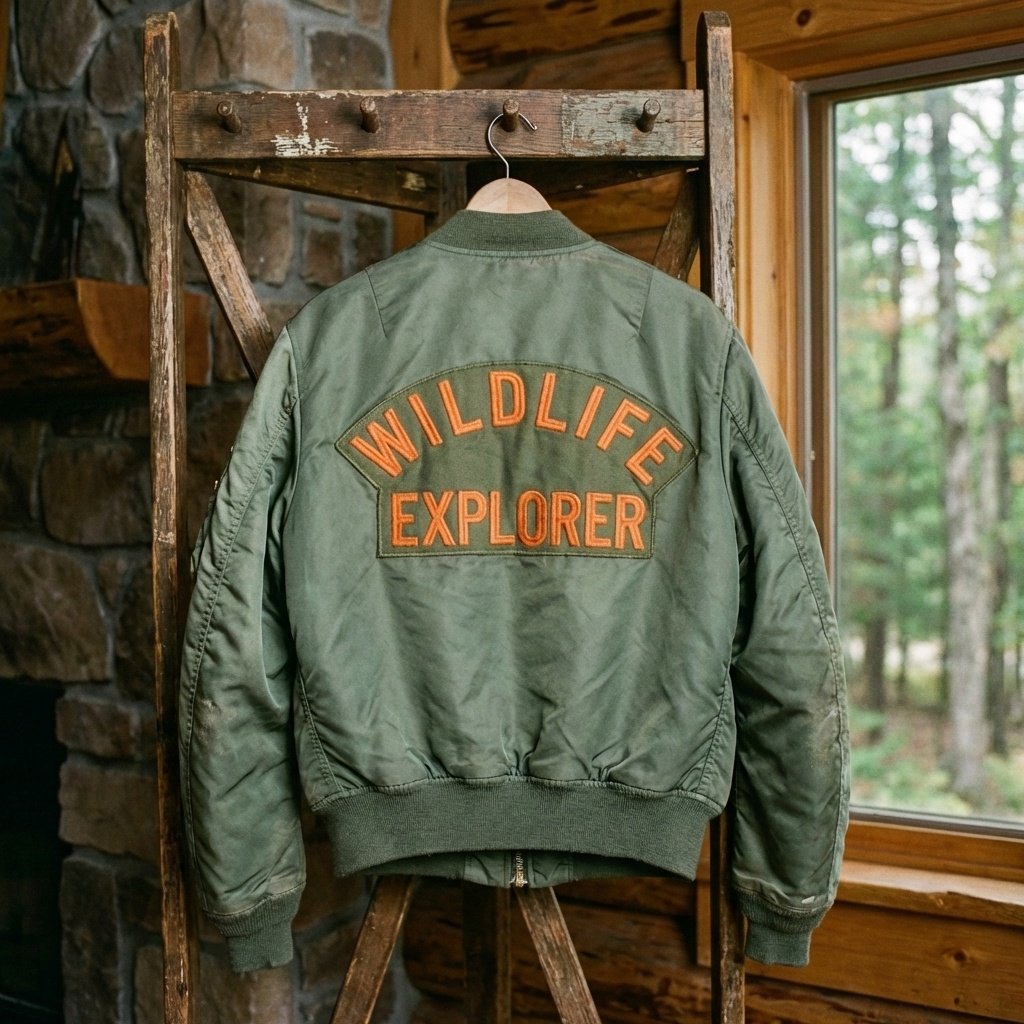}




\end{document}